\documentclass[12pt]{article}
\usepackage{enumerate}
\usepackage{url} 

\newcommand{\blind}{1}

\RequirePackage{amsthm,amsmath,amsfonts,amssymb}
\usepackage[hidelinks]{hyperref}
\hypersetup{
    colorlinks,
    linkcolor={red!50!black},
    citecolor={blue!50!black},
    urlcolor={blue!80!black}
}

\usepackage[show]{simple_notes}
\usepackage[authoryear]{natbib}
\usepackage{natbib}
\usepackage{url}
\usepackage{bm}
\usepackage{graphicx}
\usepackage{color,xcolor}
\usepackage[ruled]{algorithm2e}
\usepackage{longtable}
\usepackage{supertabular}
\usepackage{booktabs}
\usepackage{multirow}
\usepackage{listings} 
\usepackage{xcolor} 
\usepackage{booktabs,dcolumn,caption}
\usepackage{lineno}
\usepackage{url}

\usepackage{graphicx}
\usepackage{color,xcolor}
\usepackage{longtable}
\usepackage{supertabular}
\usepackage{enumitem}
\usepackage{booktabs}
\usepackage{multirow}
\usepackage[normalem]{ulem}
\usepackage{listings} 
\usepackage{xcolor} 
\usepackage{booktabs,dcolumn,caption}
\usepackage{lmodern}

\newtheorem{assumption}{Assumption}
\usepackage{setspace}
\newcommand{\EE}{\mathbb{E}}

\usepackage{placeins}
\usepackage{float}
\def\be{\begin{equation}}
\def\ee{\end{equation}}
\def\bea{\begin{eqnarray}}
\def\eea{\end{eqnarray}}
\def\nn{\nonumber}

\def\nin{\noindent}
\newcommand{\PP}{\mathrm{P}}
\newcommand{\one}{1_{k_1k_2}}
\usepackage{lastpage}

\usepackage[ruled]{algorithm2e}
\def\be{\begin{equation}}
\def\ee{\end{equation}}
\def\bea{\begin{eqnarray}}
\def\eea{\end{eqnarray}}
\def\nn{\nonumber}

\def\nin{\noindent}
\newcommand{\vertiii}[1]{{\left\vert\kern-0.25ex\left\vert\kern-0.25ex\left\vert #1 
    \right\vert\kern-0.25ex\right\vert\kern-0.25ex\right\vert}}

\newtheorem{theorem}{Theorem}
\newtheorem{lemma}{Lemma}

\theoremstyle{remark}

\usepackage{setspace}
\doublespace 

\begin{document}          

\if1\blind
{
  \title{\bf Identifying Heterogeneity in Distributed Learning}
\author{
  Zelin Xiao\thanks{Email: xiaozelin@stu.pku.edu.cn} \\
  Center for Statistical Science, Peking University
  \and
  Jia Gu\thanks{Email: gujia@zju.edu.cn} \\
  Center for Data Science, Zhejiang University
  \and
  Song Xi Chen\thanks{Email: sxchen@tsinghua.edu.cn} \\
  Department of Statistics and Data Science, Tsinghua University
}
  \maketitle
} \fi

\if0\blind
{
  \bigskip
  \bigskip
  \bigskip
  \begin{center}
    {\LARGE\bf Identifying Heterogeneity in Distributed Learning}
\end{center}
  \medskip
} \fi

\bigskip

\begin{abstract}
We study methods for identifying heterogeneous parameter components in distributed M-estimation with minimal data transmission. One is based on a re-normalized Wald test, which is shown to be consistent as long as the number of distributed data blocks $K$ is of a smaller order of the minimum block sample size {and the level of heterogeneity is dense}.   
The second one is an extreme contrast test (ECT) based on the difference between the largest and smallest component-wise estimated parameters among data blocks.  
By introducing a sample splitting procedure, the ECT can  avoid the bias accumulation arising from the M-estimation procedures, and exhibits consistency for $K$ being much larger than the sample size while the heterogeneity is sparse. The ECT procedure is easy to operate and communication-efficient. A  combination of the Wald and the extreme contrast tests is formulated to attain more robust power under varying levels of sparsity of the heterogeneity. We also conduct intensive numerical experiments to compare the family-wise error rate (FWER) and the power of the proposed methods.  Additionally, we conduct a case study to present the implementation and validity of the proposed methods.
\end{abstract}

\noindent%

\bigskip

\section{Introduction}\label{section:Introduction}
Distributed learning has gained considerable attention in modern machine learning and statistics. A primary advantage of distributed learning is its ability to efficiently process massive datasets that cannot be centralized due to storage or computational limitations or privacy concerns. By distributed computation across multiple data blocks, it mitigates the need for expensive data transmission, especially when the data is geographically dispersed. Additionally, distributed learning frameworks, such as federated learning, facilitate collaborative model training without sharing raw data, thereby preserve the privacy. The ``split-and-conquer'' (SaC) estimation  \citep{jmlr2013} is a standard method in the distributed learning. There have been extended studies on it, including \cite{Chen-Xie-2014} and \cite{Battey-2018} 
by incorporating regularization techniques for high-dimensional signal detection.  \cite{chenpeng2021} refined the SaC method by examining the estimation efficiency and asymptotic properties of general asymptotically symmetric statistics.  Despite these advances, a major challenge in distributed learning remains the trade-off between global aggregation and local adaptability in the presence of heterogeneous data.

The aforementioned studies in distributed inference typically assumed homogeneity across data blocks, an assumption that may be held too strong in real-world scenarios. Recent research has increasingly focused on balancing global consistency with local specificity in the context of heterogeneous data, where identifying parameter heterogeneity is essential for several reasons. One is that it serves as a prerequisite for the weighted aggregation of the common parameters. \cite{Duan-2021}, who extended the surrogate likelihood framework  \citep{Jordan-2019} to heterogeneous settings, facilitating communication-efficient aggregation of parameters from heterogeneous models. \cite{jiagu2023} demonstrated that properly weighted SaC estimator could outperform full-sample methods under  heterogeneity with results. However, both of these approaches needed prior knowledge of the common and heterogeneous parameters. More recently, \cite{Guo_ROBUST_JASA_2025} introduced a federated meta-learning framework for statistical inference, aiming to identify and inference on the majority model's parameters.

In situations with severe heterogeneity, directly aggregating model parameters without accounting for these differences often leads to suboptimal outcomes. \cite{Li_Sai_Transfer_Learning} demonstrated the importance of selecting a candidate set before transferring learning to a target source. \cite{Heterogeneous_Quantization_in_Federated_Learning}  considered the assignment of different aggregation weights to clients based on a given  heterogeneity. \cite{weijiesu_threshold} showed that when the data heterogeneity is too large, pure local training becomes minimax optimal.  These studies highlight the critical role of identifying and quantifying heterogeneity as a key step improving federated learning and transfer learning.

{This paper is aimed at 
developing test procedures to identify heterogeneity in distributed learning with special attention for situations where the number of distributed data blocks \( K \) far exceeds the minimal sample size \( n_{\min} \), so as to facilitate the needs of the explosive growth in the number of devices in modern distributed learning. We consider three distributed heterogeneity tests which are both communicational and computational efficient. } 
{A natural choice is the Wald test, which builds on a statistic that conducts pairwise comparison among the distributed parameters, but incurs bias accumulation when \( K \) is large.  The second one is the extreme contrast test (ECT), which involves splitting each distributed data block to two parts. The first split is used to identify the maximum and minimum parameter estimates among data blocks and obtain block labels to which the two extreme parameter estimates belong. The second splits are used to formulate a t-statistic base on  parameter estimates of the  two identified ``extreme label pairs", which is showed to be asymptotically normally distributed due to the sample splitting and avoids the bias accumulation while being computational efficient. 
The ECT is showed to be operational without imposing much restriction between the number of 
the data blocks $K$ and the minimal sample size $n_{\min}$. 
A power analysis reveals that the Wald test achieves higher (lower) power than the ECT when the heterogeneity is  dense (sparse)} among components of the parameters, namely when there are many (only a few) dimensions of the parameter taking different values among the data blocks. 
In order to adapt to unknown level of heterogeneity sparsity,  
 we propose a combined test whose statistic is a properly weighted sum of the Wald and ECT statistics to combine their strengths against dense or sparse alternatives, with the combined statistic shown to be asymptotically normally distributed due to the asymptotic independence of the two member statistics and is demonstrated to have robust power performance for both dense and sparse heterogeneity.

The remainder of this paper is organized as follows. Section \ref{section:Preliminaries} introduces the setting of the distributed learning and the problem of testing homogeneity among parameters, together with basic assumptions. In Section \ref{Section:Threemethods} we introduce the two distributed testing procedures. Section \ref{section:combination_test} considers combining the two statistics. Section \ref{section:simulation} and Section \ref{section:case_study} report the numerical simulation and case study, 
respectively. Section \ref{Section:conclusions} summarizes our results.  Additional technical details and results are reported in the Appendix.

\section{Preliminaries and Notations}\label{section:Preliminaries}
For a vector $\boldsymbol{\theta}=(\theta^{(1)},\ldots,\theta^{(p)})^{\top}$, denote $\|\boldsymbol{\theta}\|_2 = \sqrt{\sum_{l = 1}^{p}(\theta^{(l)})^2}$.  For a matrix $\boldsymbol{A}\in\mathbb{R}^{d_1\times d_2}$, define $\|\boldsymbol{A}\|=\sup_{\|\boldsymbol{v}\|_2 = 1}\|\boldsymbol{A}\boldsymbol{v}\|_2$ as the matrix norms. For a sequence of random variables $\{X_n\}_{n = 1}^{\infty}$, we denote $X_n = \mathcal{O}(a_n)$ if there holds $\lim\limits_{C\rightarrow\infty}\limsup\limits_{n\rightarrow\infty}\mathbb{P}(|X_n|>Ca_n)=0$, and denote $X_n = o(a_n)$ if there holds that $\lim\limits_{C\rightarrow 0^+}\limsup\limits_{n\rightarrow\infty}\mathbb{P}(|X_n|>Ca_n)=0$.  For two sequences $a_n,b_n$, we say $a_n\preceq b_n$ when $a_n=\mathcal{O}(b_n)$, $a_n\succeq b_n$ when $b_n=\mathcal{O}(a_n)$, and $a_n\asymp b_n$ when $a_n = \mathcal{O}(b_n)$ and $b_n = \mathcal{O}(a_n)$ hold at the same time. We say $a_n\approx b_n$ if $\lim_{n\rightarrow\infty}a_n/b_n = 1$. 

Suppose there is a large dataset of size \(N\), which is stored in $K$ separated 
nodes (or blocks). Each node has a local dataset $\mathcal{D}_k=\{X_{k,i}\}_{i=1}^{n_k}$, which contains \(n_k\) independent and identically distributed (i.i.d.) random vectors  drawn from  distribution \(F_k\), defined on \((\mathbb{R}^d, \mathcal{R}^d)\). Denote  \( n_{\min} = \min_k n_k \) as the minimum sample size among the data blocks.
 For the $k$-th data block, there exists a loss function \(M_k(X; \theta_k)\) with unknown parameter $\theta_k$. The distribution \(F_k\) which depends on a $p-$dimensional parameter \(\theta_k\), given by  \(\theta_k = (\theta_k^{(1)}, \theta_k^{(2)}, \dots, \theta_k^{(p)})^\top\).
 
 We allow the parameter $\theta_k$ to vary across data blocks to reflect potential heterogeneity among the data blocks. The true parameter \(\theta_k^*\)  is defined as the unique minimizer of the expected loss function, namely $\theta_k^* = \arg\min_{\theta_k \in \Theta_k} E_{F_k} \left( M_k(X_{k,1}; \theta_k) \right).$  Here, \(\Theta_k \subseteq \mathbb{R}^p\) is the parameter space and the expectation is taken with respect to the distribution \(F_k\) in the \(k\)-th block. The goal of this study is to identify the common parameters across $K$ data blocks for better federated learning. {Moreover, the local loss
 functions $M_k(X;\theta_k)$ are allowed to be different across data blocks to reflect the reality that practitioners may not want to test the heterogeneity for the whole parameter given some prior knowledge \citep{Duan-2021,jiagu2023}. Under that scenario, the local loss function $M_k$ can be 
the profiled loss function such that 
\be
M_k(X;\theta_k) = \min_{\phi_k \in \Psi_k}L(X;\theta_k,\phi_k),\nn
\ee
where $L(\cdot \ ;\ \cdot)$ is the original loss function, and $\phi_k \in \Psi_k \subset{R^{p_1}}$ is the nuisance parameter.

We aim at testing  heterogeneity for each  \(j = 1, 2, \dots, p\), namely \begin{align}
&\mathcal{H}_0^{(j)} : \theta_1^{(j),*} =  \theta_2^{(j),*}=\cdots = \theta_K^{(j),*}  \text{  versus}\\ 
& \mathcal{H}_1^{(j)} : \text{there exists } 1\leq k \neq k'\leq K\quad  \text {such that}\quad\theta_k^{(j),*} \neq \theta_{k'}^{(j),*}.
\label{the_hypothesis}
\end{align}
 That is, we want to test whether the \(j\)-th component of the parameter vectors \(\{\theta_k^*\}_{1\leq k\leq K}\) are the same across all data blocks versus if there exists difference (heterogeneity) 
 among the parameters of the blocks. {Testing this  hypothesis has a direct implication in transfer learning  \citep{Li_Sai_Transfer_Learning}, personalized federated learning  \citep{Liu_JASA_Personalized_FL_2024}, and  multitask learning  \citep{Duan-AoS-2023,Huang_multitask_bandits}, which helps answer the question of whether data pooling is necessary.} 

 Let $\bm S = \{j\in[p]: \mathcal{H}_1^{(j)} \text{ is true}\}$ denote the set of indices corresponding to the true alternative (heterogeneous) hypotheses, where $[p]=\{1,\ldots,p\}$, and  $\widehat{\bm S}$ denote the set of indexes of rejected hypotheses by a test procedure. 
 
With data $\{X_{k,i}\}$ in the $k$-th block or node, we can estimate the true parameter $\theta_k^*$ by minimizing the empirical loss  to obtain the M-estimator  \[\widehat{\theta}_k=\arg\min_{\theta_k \in\Theta_k}\sum\limits_{i=1}^{n_k} M_k(X_{k,i},{\theta}_k).\] 
 If the loss function $M_k$ is smooth, the corresponding $Z-$estimator is attained by solving
 
\begin{equation}\label{eq: estimating_equation}
\sum\limits_{i=1}^{n_k} \nabla_{{\theta}_k}M_k(X_{k,i},\widehat{\theta}_k) = 0.
\end{equation} Let $ \mathbf{V}_k(\theta_k^*){=}[\mathbb{E}\nabla^2_{\theta_k}M_k(X_k,\theta^{*}_{k})]^{-1} [\mathbb{E}\nabla_{\theta_k}M_k(X_k,\theta_k^{*})\nabla_{\theta_k}M_k(X_k,\theta_k^{*})^\top][\mathbb{E}\nabla^2_{\theta_k}M_k(X_k,\theta^{*}_{k})]^{-1}$. 
According to the standard results in the M-estimation  \citep{Vaart_1998}, 
$$\sqrt{n_k}(\widehat{\theta}_k-\theta_k^{*})\xrightarrow[]{d}   \mathcal{N}(0,\mathbf{V}_k(\theta_k^*)) \quad \hbox{as } n_k\to\infty. $$  
Denote $Z_k(X_{k,i},\widehat{\theta}_{k})=M_k(X_{k,i},\widehat{\theta}_{k})M_k(X_{k,i},\widehat{\theta}_{k})^\top$ and the $j$-th diagonal element of $\mathbf{V}_k(\theta_k^{*})$  as $(\sigma_k^{(j)})^2$, we can use \begin{equation}\label{X^-1YX}
\widehat{\mathbf{V}}_k(\widehat{\theta}_{k})=\{\dfrac{1}{n_k}\sum\limits_{i=1}^{n_k}\nabla^2_{\theta_k}M_k(X_{k,i},\widehat{\theta}_{k})\}^{-1}\{\dfrac{1}{n_k}\sum\limits_{i=1}^{n_k}Z_k(X_{k,i},\widehat{\theta}_{k})\}\{\dfrac{1}{n_k}\sum\limits_{i=1}^{n_k}\nabla^2_{\theta_k}M_k(X_{k,i},\widehat{\theta}_{k})^{\top}\}^{-1}
\end{equation}to estimate $\mathbf{V}_k(\theta_k^*)$ and denote   $\widehat{\mathbf{V}}_k(\widehat{\theta}_{k})$'s $j$-th diagonal element's square root as $\widehat{\sigma}_k^{(j)}$ which estimates ${\sigma}_k^{(j)}$. 

We make the following assumptions for our analysis.

\begin{assumption}[Identifiability and compactness]\label{assumption:identifiability_compactness}
The parameter $\theta_k^*$ is the unique minimizer of $M_k(\theta_k) = E(M_k(X_{k,1};\theta_k)) $ with respect to $\theta_k \in \Theta_k$. The true parameter $\theta_k^*$ is an interior point of the parameter space $\Theta_k$, which is a compact and convex set in $\mathbb{R}^p$. It holds that $\sup_{\theta_k \in \Theta_k} \|\theta_k-\theta_k^{*}\|_2 \leq r$ for  $1\leq k \leq K$ and some $r > 0$.
\end{assumption}

\begin{assumption}[Smoothness for Hessian matrix and outer product]\label{assumption:Smoothness for Hessian}
The loss function on the $k$-th data block is twice differentiable with respect to $\theta_k$ and there are positive constants $R$, $L$, $v_1$,  $v_2$ and $v_3$ such that for $1\leq k \leq K$, $\EE\left( \|\nabla^2 M_k(X_{k,1}; \theta_k^*)-\mathbb{E}\nabla^2 M_k(X_{k,1}; \theta_k^*)\|_2^{2v_1} \right) \leq L^{2v_1}$
 and $\EE\left(\left\|\nabla_{\theta_k}M_k(X_{k,1}; \theta_k^*) \right\|_2^{2v_2}\right) \leq R^{2v_2}$. Denote \(Z_k(x, \theta_k) = \nabla M_k(x; \theta_k) \nabla M_k(x; \theta_k)^\top\). Moreover, there are positive constants $\rho,B$ and $G$ such that
 \[
\left\|Z_k(x, \theta_k)-Z_k(x, \theta_k')\right\|_2 \leq B(x) \|\theta_k-\theta_k'\|_2,
\]
\[
\|\nabla^2 M_k(x; \theta_k)-\nabla^2 M_k(x; \theta_k')\|_2 \leq G(x)\|\theta_k-\theta_k'\|_2
\]
for all $\theta_k, \theta_k'$ in $U_k = \{\theta_k \mid \|\theta_k-\theta_k^*\|_2 \leq \rho\}$, $x \in \mathbb{R}^d$ and $E(G(X_{k,1})^{2v_3}) \leq G^{2v_3},E(B(X_{k,1})^{2v_4}) \leq B^{2v_4}$. Assume $v=\text{min}\{v_1,v_2,v_3,v_4\}\geq 2.$

\end{assumption}

\begin{assumption}[Local strong convexity]\label{assumption: Local Strong Convexity}
The population loss function on the $k$-th data block, $M_k(\theta_k^*) = E(M_k(X_{k,1};\theta_k^*))$, is twice differentiable. Moreover, denote $\mathbb{E} Z_k(X_{k,1};\theta_k^*)=Z_k(\theta_k^*)$, there exists  constants $\sigma^{+} > \sigma^{-} > 0$ such that for $1\leq k \leq K,$
\[ 
(\sigma^{+})^2 I_{p \times p} \succeq \nabla_{\theta_k}^2 M_k(\theta_k^*),  Z_k(\theta_k^*)\succeq (\sigma^{-})^2 I_{p \times p}.
\] 

\end{assumption}Here $A \succeq B$ means $A-B$ is a non-negative semi-definite matrix.

Assumptions \ref{assumption:identifiability_compactness}-\ref{assumption: Local Strong Convexity} are standard in  distributed estimation literature \citep{jmlr2013,Duan-2021,jiagu2023}. Besides,  while existing works require balanced sample size across the data blocks, only a much relaxed condition, namely $\max
\limits_{k_1,k_2}\ \frac{n_{k_1}}{n_{k_2}}=o(K^{1-1/v})$, will be imposed to ensure the validity of the combined test in Section \ref{section:combination_test}.

\section{Heterogeneity Tests of Wald-Type and Extreme Contrast} 
\label{Section:Threemethods}
This section first puts forward two test procedures suitable for distributed settings, a Wald-type test and an extreme contrast test (ECT). We compare their power in local alternative and show that each test performs better under different levels of heterogeneity sparsity. 

\subsection{A Re-normalized Wald-type Test }\label{section:The Wald-type Test}

Since the local estimators $\{\widehat{\theta}_k\}$ are asymptotically normal under certain conditions, we can construct a Wald-type test for heterogeneity. 
We first consider the fixed$-K$ scenario when considering testing $\mathcal{H}_{0}^{(j)}\ v.s. \ \mathcal{H}_{1}^{(j)}$ base on the  $j$-th dimension estimates:
\begin{equation}\label{equation_aggregation_each_dimension_Wald}\widehat{\theta}^{(j)}=(\widehat{\theta}^{(j)}_1,\widehat{\theta}^{(j)}_2,\cdots,\widehat{\theta}^{(j)}_K)^\top, \end{equation}
{which involves sharing only the local estimates $\widehat{\theta}^{(j)}$ to the central server.} 

The homogeneity hypothesis  $\mathcal{H}_{0}^{(j),*}$ is equivalent to $R_K{\theta}^{(j)}=0$ where 
\begin{equation}
R_K=\begin{pmatrix}
  1  &  -1 &  0  & \cdots & 0\\
  0  &  1  &  -1 & \cdots & 0\\
  \cdots & \cdots & \cdots & \cdots & \cdots \\
  \cdots & \cdots &  0  &  1  & -1 
\end{pmatrix}_{(K-1)\times K\quad.}
\label{1_-1_0_0_matrix_R_k}
\end{equation}

Let $\Lambda_j=\mathrm{diag}\{n^{-1}_{1}(\sigma_1^{(j)})^2,$ 
 $\cdots,n^{-1}_{K}(\sigma_K^{(j)})^2\}$ and  
$\widehat{\Lambda}_j=\mathrm{diag}\left\{n^{-1}_{1}(\widehat{\sigma}_1^{(j)})^2,\cdots,n^{-1}_{K}(\widehat{\sigma}_K^{(j)})^2\right\}$ which can be obtained by  \eqref{X^-1YX}. 
A  Wald-type test statistic is $$W_{n}^{(j)}=\widehat{\theta}^{(j)^\top}R_K^\top(R_K\widehat{\Lambda}_jR_K^\top)^{-1} R_K\widehat{\theta}^{(j)},$$ 
which only needs transmission of the local $\{\widehat{\theta}_k\}_{1\leq k\leq K}$ and $\{\widehat{\sigma}_k^{(j)}\}_{1\leq j\leq p,1\leq k\leq K}$, and can be achieved with one-round communication. 

To gain insight into the statistic, suppose that $\Lambda_j=n^{-1}I_{K}$. Then, for the vector $\widehat{\theta}^{(j)}$, the statistic becomes $n\widehat{\theta}^{(j)^\top} R_K^\top (R_K R_K^\top)^{-1} R_K\widehat{\theta}^{(j)}$ which can be written as  
$ nK^{-1}\sum\limits_{1 \leq k_1 < k_2 \leq K} (\widehat{\theta}^{(j)}_{k_1}-\widehat{\theta}^{(j)}_{k_2})^2$. Generally speaking, $W_{n}^{(j)}$ is a weighted average of $\{(\widehat{\theta}^{(j)}_{k_1}-\widehat{\theta}^{(j)}_{k_2})^2\}_{k_1\ne k_2}$. 
Hence, the Wald-type statistic performs pairwise comparisons of the estimates $(\widehat{\theta}_{k_1}, \widehat{\theta}_{k_2})$. Under the $\mathcal{H}_0^{(j)}$, due to the independence of the data blocks,  it can be shown that for the fixed$-K$ scenario 
$W_{n}^{(j)}\overset{d}{\to} \chi^2(K-1) \text{ as }n_{min}\to\infty $, where $n_{min}=\arg\min_k n_k.$

 In order to facilitate the Wald test in distributed setting with diverging $K$, we consider a re-normalized Wald statistic 
\begin{equation}\label{renorm_Wald}
W^{(j)}_{N,K}=\dfrac{\widehat{\theta}^{(j)^\top}R_K^\top(R_K\widehat{\Lambda}_jR_K^\top)^{-1} R_K\widehat{\theta}^{(j)}-(K-1)}{\sqrt{2K-2}}.
\end{equation}
It can be shown in Theorem \ref{Wald consistency} that $W^{(j)}_{N,K}$ converges to a standard normal distribution under mild conditions. 
\begin{algorithm}
\caption{Wald-type test For Testing Heterogeneity }
\LinesNumbered
\KwIn{$\{X_{k,i},k = 1,...,K; i = 1,...,n_k\}$, level $\alpha,$ $K=o(n_{\text{min}}).$}
\KwOut{$\bm{\hat{S}}=\{j:\mathcal{H}_{0}^{(j)}\text{ is rejected}, 1\leq j \leq p.\}$}
Each data block $k$ transmits $\widehat{\theta}_k$ and 
$\{(\widehat{\sigma}_k^{(j)})^2/n_k\}_{1\leq j \leq p}$ to the central server.

Calculate re-normalized Wald-type statistic $W^{(j)}_{N,K}$ via equation  \eqref{renorm_Wald} for each dimension $1\leq j \leq p.$
 in the central server. The $p-$value for dimension $j$ is ${P}_{Wald}^{(j)}=1-\Phi(W^{(j)}_{N,K}).$
 
Reject the null hypothesis $\mathcal{H}_0^{(j)}$ if ${P}_{Wald}^{(j)}<\alpha/p.$
\end{algorithm}
Let $\Phi(x)$ be the distribution function of the standard normal distribution and $\Phi^{-1}(x)$ be its quantile function. We can design a distributed test for $\mathcal{H}_0^{(j)}$. We calculate the statistic $W_{N,K}^{(j)}$ when the central server receives the estimates needed for $W_{N,K}^{(j)}$ from each node. The asymptotic normality of $W_{N,K}^{(j)}$ facilitate us to  obtain the  $p$-value $p_{Wald}^{(j)}=1-\Phi(W_{N,K}^{(j)})$ for each $j\in[p]$. Denote the set of hypotheses rejected by using the re-normalized Wald-type test as $\bm{\widehat{S}}_{\text{Wald}} \subset [p]$ and the family-wise error rate (FWER) is defined as $\PP(|\bm S^c\cap \hat{\bm S}_{\text{Wald}}|\geq1).$ Thus we can reject the null $\mathcal{H}_0^{(j)}$ if ${p}_{Wald}^{(j)}<\alpha/p$ to control the FWER at level  $\alpha$ via Bonferroni correction.

Denote the non-central part of Wald statistic as \begin{equation} \label{equation: nu_j}\nu_{N,K}^{(j)}=\dfrac{\theta^{(j),*^\top}R_K^\top (R_K\Lambda_j R_K^{\top})^{-1} R_K\theta^{(j),*}}{\sqrt{2K-2}},
    \end{equation} which is nonzero unless  $\mathcal{H}_{0}^{(j)}$ holds.  Then we have the following theorem.
\begin{theorem}[Wald-type test's Consistency]
\label{Wald consistency}
  For $K=o(n_{\text{min}})$ and each dimension $1\leq j \leq p$, under the null hypothesis $\mathcal{H}_{0}^{(j)}$ and Assumptions \ref{assumption:identifiability_compactness}-\ref{assumption: Local Strong Convexity},  we have $W^{(j)}_{N,K}\overset{d}{\to} \mathcal{N}(0,1)$ as $K \text{ and }n_{\text{min}}\xrightarrow[]{}\infty $. Under the alternative hypothesis $\mathcal{H}_1^{(j)},$ if the non central part $\nu_{N,K}^{(j)}\xrightarrow[]{} \infty$ as $K \text{ and }n_{\text{min}}\xrightarrow[]{}\infty $ holds, then   $\mathrm{P}(j\in \bm{\widehat{S}}_{\text{Wald}})\xrightarrow[]{}  1$ as $K \text{ and }n_{\text{min}}\xrightarrow[]{}\infty $.
\end{theorem} 

There are two reasons for constraining the growth rate of \(K=o(n_{\min})\)  to ensure the asymptotic normality of $W^{(j)}_{N,K}$ under the null. The first one is the remainder term of the M-estimator. By conducting a Taylor expansion of equation  \eqref{eq: estimating_equation} around \(\theta_k^*\), we obtain 
\begin{equation}\label{eq:Def_for_R_k}
\widehat{\theta}_k-\theta_k^*=-(\mathbb{E}\nabla^2_{\theta_k}M_k(X_{k,i},\theta_k^*))^{-1}\cdot\frac{1}{n_k}\sum_{i = 1}^{n_k}\nabla_{\theta_k}M_k(X_{k,i},\theta_k^*)+R_k,\text{ where}\end{equation}
 \[R_k=-(\mathbb{E}\nabla^2_{\theta_k}M_k(X_{k,i},\theta_k^*))^{-1}\cdot\left(\mathbb{E}\nabla^2_{\theta_k}M_k(X_{k,i},\theta_k^*)-\frac{1}{n_k}\sum_{i = 1}^{n_k}\nabla_{\theta_k}M_k(X_{k,i},\widehat{\theta}_k)\right)(\widehat{\theta}_k-\theta_k^*).\]
 Under mild conditions,  \(\mathbb{E}|R_k| = \mathcal{O}(n_k^{-1})\) as  shown by \cite{jmlr2013}.  Thus, restricting \(K = o(n_{\text{min}})\) is essential to manage the cumulative impact of \(R_K\). There are several methods available for reducing bias in M-estimation  \citep{biometrika_debiased_1993,jmlr2013, kim_bias_correction, Chernozhukov_debias,jiagu2023} which may alleviate the bias accumulation.  However, even if $R_k=0$, the bias in $\{\widehat{\sigma}_k^{(j)}\}_{1\leq k \leq K}$ in the formulation of \((R_K\widehat{\Lambda}_jR_K^\top)^{-1}\)  is another reason that limits the growth of $K$. 
For most variance estimators,  \(\widehat{\sigma}_k^{-1}\) is consistent  to $\sigma_k^{-1}$ but biased in finite samples, and this bias can also accumulate in the  Wald-type statistic $W_{N,K}^{(j)}$. 
 But it can be proved that the bias mentioned above is also negligible as \(n_{\text{min}}\rightarrow\infty\) when \(K = o(n_{\text{min}})\). Given the simplicity of the proposed extreme contrast test in Section \ref{section_ECT}, which can avoid the bias accumulation by sample splitting,  we do not get into  the debiasing the re-normalized Wald statistics. 

\subsection{An Extreme Contrast Test}\label{section_ECT}
We introduce an extreme contrast test (ECT) which is not only computational and communicational efficient, but 
   also allow a much larger number of devices $K$ than the minimal sample size. The ECT, which is based on sample splitting, is shown as follows.  First, each sample \( \mathcal{D}_k \) of size \( n_k \) is split to two mutually exclusive parts, \( \mathcal{D}_k^{[1]} \) and \( \mathcal{D}_k^{[2]} \). 
   The first part \( \mathcal{D}_k^{[1]} \) contains approximately a \(1-\gamma\) proportion of the local sample \( \mathcal{D}_k \) for a \( 0 < \gamma < 1 \), namely \( \mathcal{D}_k^{[1]} = \{ X_{k,i} \}_{1\leq i \leq \lfloor(1-\gamma)n_k\rfloor} \), which is used to obtain the M-estimates \( \{\widehat{\theta}_k^{(j),[1]}\}_{1\leq j\leq p } \) for $1\leq k \leq K$. The second part \( \mathcal{D}_k^{[2]} \) is used to compute another set of estimators \(\{ \widehat{\theta}_k^{(j),[2]}\}_{1\leq j\leq p} \) and their standard errors \(\{ \widehat{SE}_{k}^{(j),[2]}\}_{1\leq j\leq p} =\{ \frac{\widehat{\sigma}_{k}^{(j),[2]}}{\sqrt{\lceil\gamma n_{k}\rceil}} \}_{1\leq j\leq p}\) for $1\leq k \leq K$. We suggest  choosing  $\gamma=2/3$ to improve the power of the ECT which is explained in Section \ref{section:discussion Recommended_gamma} in the Appendix.

Subsequently, the central server identifies the data blocks whose M-estimates from the first split \( \{D_k^{[1]}\}_{k=1}^K \) attain the maximum and minimum values, respectively:
\begin{equation}\label{k_j^max_origin}
    \hat{k}_{\text{max}}^{(j)} = \underset{k}{\arg\max}\, \widehat{\theta}_k^{(j),[1]},  \quad \hat{k}_{\text{min}}^{(j)} = \underset{k}{\arg\min}\, \widehat{\theta}_k^{(j),[1]}.
\end{equation} 

{If there are multiple splits with the same extreme values, we choose one of them. With the extreme indices $ \hat{k}_{\text{max}}^{(j)} $ and $\hat{k}_{\text{min}}^{(j)}$ identified from the first subsample splits, an extreme contrast  statistic is formulated based on the two local M-estimates from the second subsample splits of the  $\hat{k}_{\text{max}}^{(j)} $ and $\hat{k}_{\text{min}}^{(j)}$-th local samples:} 
\begin{equation}\label{equation_ECT_Definition}
T_{N,K}^{(j)} = \frac{\widehat{\theta}_{\hat{k}_{\text{max}}^{(j)}}^{(j),[2]}-\widehat{\theta}_{\hat{k}_{\text{min}}^{(j)}}^{(j),[2]}}{\sqrt{\left(\widehat{SE}^{(j),[2]}_{\hat{k}^{(j)}_{\max}}\right)^2 + \left(\widehat{SE}^{(j),[2]}_{\hat{k}^{(j)}_{\min}}\right)^2}}.
\end{equation}
It can be shown that under certain conditions, $T_{N,K}^{(j)} \overset{d}{\to}\mathcal{N}(0,1)$ without   restriction on the relationship between $K$ and $n_{\min}$. We use the following lemma to illustrate this phenomenon.   

\begin{lemma}\label{lem: ECT explanation}
Let $(\Omega, \mathcal{F}, P)$ be  a probability space on which a collection of random variables $\{t_{k,n}\}_{k\geq 1, n\geq1}$ and indicator random variables $\{\mathbb{I}_k\}_{k\geq 1}$ are defined, where $\{\mathbb{I}_k\}_{k\geq 1}$ are mutually independent of $\{t_{k,n}\}_{k\geq 1, n\geq1}$ and satisfy $\textstyle\sum\limits_{k\geq 1} \mathbb{I}_{k} = 1$. Assume that there exist constants $C>0,\alpha_1>0$, which are free of $K$, such that for each $k \geq 1$, 
\begin{equation}\label{lemma——assumption}
    \sup_{x \in \mathbb{R}} \left|\mathrm{P}(t_{k,n} \leq x)-\Phi(x) \right| \leq Cn^{-\alpha_1},
\end{equation}
then we have 
$\sup_{x \in \mathbb{R}} \left| \mathrm{P}\left(\sum\limits^{K}_{k=1} \mathbb{I}_k t_{k,n} \leq x\right)-\Phi(x) \right| \leq Cn^{-\alpha_1}.
$
\end{lemma}
\begin{proof}
Since $\sum\limits_{k=1}^{K} \mathbb{I}_k = 1$ ensures  exactly one indicator \(\mathbb{I}_k\)  to be 1,  for any $\omega \in \Omega,$ one can define a  random variable $k^*(w)$ such that $I_{k^*}(w)=1$. Then,   
\begin{align*}
\left| \mathrm{P}\left(\sum\limits^{K}_{k=1} \mathbb{I}_k t_{k,n} \leq x\right)-\Phi(x) \right| = &\left| \mathbb{E}_{k^*} \left[ \mathrm{P}\left( \sum\limits^{K}_{k=1} \mathbb{I}_k t_{k,n} \leq x \,\big|\, k^* \right)-\Phi(x) \right] \right|\\ = &\left| \mathbb{E}_{k^*} \left[ \mathrm{P}\left(  t_{k^*,n} \leq x \,\big|\, k^* \right)-\Phi(x) \right] \right| \notag \\
\leq &\mathbb{E}_{k^*} \left| \mathrm{P}\left( t_{k^*,n} \leq x \right)-\Phi(x) \right| \\ \leq  &\sup_k \sup_x \left|\mathrm{P}(t_{k,n} \leq x)-\Phi(x) \right| \leq Cn^{-\alpha_1}.\end{align*}The first equality holds due to the law of total expectation. For the second and third lines, given \(k^*(w)\), we have \(t_{k^*,n}=\sum\limits^{K}_{k=1}\mathbb{I}_k t_{k,n}\).  Thus, the proof is completed.\end{proof}

There are two noticeable aspects in Lemma \ref{lem: ECT explanation}. One is that the result is valid without any constraint on $K$. The other is that the asymptotic normality of $\sum\limits_{k\geq 1}\mathbb{I}_k t_{k,n}$ can be guaranteed by the independence of $\{\mathbb{I}_k\}$ and $\{t_{k,n}\}$ and the asymptotic normality of $\{t_{k,n}\}$. Thus it can help us establish the asymptotic normality of ECT statistic. 
{To better illustrate the  connection between Lemma \ref{lem: ECT explanation} and the ECT statistic \(T_{N,K}^{(j)}\), we denote 
\[
\mathbb{I}_{k_1k_2}^{(j),[1]} = \mathbb{I}\left\{k_1=\underset{k}{\arg\max} \ \widehat{\theta}_{k}^{(j),[1]},\quad k_2=\underset{k}{\arg\min} \ \widehat{\theta}_{k}^{(j),[1]}\right\}, \] \[ t_{k_1k_2}^{(j),[2]} = \frac{\widehat{\theta}_{k_1}^{(j),[2]}-\widehat{\theta}_{k_2}^{(j),[2]}}{\sqrt{(\widehat{SE}_{k_1}^{(j),[2]})^2 + (\widehat{SE}_{k_2}^{(j),[2]})^2}}.
\]
We also denote the centralized version of $t_{k_1k_2}^{(j),[2]}$ as 
\[ \widetilde{t}_{k_1k_2}^{(j),[2]} = \frac{(\widehat{\theta}_{k_1}^{(j),[2]}-{\theta}_{k_1}^{(j),*})-(\widehat{\theta}_{k_2}^{(j),[2]}-{\theta}_{k_2}^{(j),*})}{\sqrt{(\widehat{SE}_{k_1}^{(j),[2]})^2 + (\widehat{SE}_{k_2}^{(j),[2]})^2}}.
\]

The ECT statistic 
\(T_{N,K}^{(j)} = \sum\limits_{1 \leq k_1,k_2 \leq K} t_{k_1k_2}^{(j),[2]} \cdot \mathbb{I}_{k_1,k_2}^{(j),[1]}\)  and its centralized version 
 is \(\widetilde{T}_{N,K}^{(j)} = \sum\limits_{1 \leq k_1,k_2 \leq K} \widetilde{t}_{k_1k_2}^{(j),[2]} \cdot \mathbb{I}_{k_1,k_2}^{(j),[1]}\) satisfying $T_{N,K}^{(j)}=\widetilde{T}_{N,K}^{(j)}$ under the null hypothesis. Though we choose extreme values in $\mathbb{I}_{k_1,k_2}^{(j),[1]}$, due to the independence of \(\{\widetilde{t}_{k_1k_2}^{(j),[2]}\}\) and \(\{\mathbb{I}_{k_1,k_2}^{(j),[1]}\}\), we can regard \(\{\mathbb{I}_{k_1,k_2}^{(j),[1]}\}\) as \(\{\mathbb{I}_k\}\) and \(\{\widetilde{t}_{k_1k_2}^{(j),[2]}\}\) as \(\{t_{k,n}\}\) in the Lemma \ref{lem: ECT explanation} above, thus the asymptotic normality of 
 $\widetilde{T}_{N,K}^{(j)}$ can still be guaranteed as shown in the next theorem. Recall that  \(v \geq 2\) is a constant  appeared in Assumption \ref{assumption:Smoothness for Hessian}. We also define a constant \(\rho_1 > 0\)  independent to \(K\), \(j\) and \(n_{\min}\).

\begin{theorem}[Asymptotic Normality for ECT]\label{theorem: ECT_asym_normal}
 For each dimension \(1 \leq j \leq p\), under the  Assumptions \ref{assumption:identifiability_compactness}-\ref{assumption: Local Strong Convexity}, for a fixed pair \(k_1\ne k_2\), we have 
\bea && 
\sup_{x \in \mathbb{R}} \left| \mathrm{P}(\widetilde{t}_{k_1k_2}^{(j)} \leq x)-\Phi(x) \right| \leq \rho_1 \cdot n_{\min}^{-\frac{v}{2(v+2)}}, \quad \hbox{and} \nn  \\ 
&& \sup_{x \in \mathbb{R}} \left| \mathrm{P}(\widetilde{T}^{(j)}_{N,K} \leq x)-\Phi(x) \right| \leq \rho_1 \cdot n_{\min}^{-\frac{v}{2(v+2)}}. \nn 
\eea 
\end{theorem}

{ It is noted that despite both $\widetilde{t}_{k_1k_2}^{(j)}$ and  $\widetilde{T}^{(j)}_{N,K}$ are formulated with the extreme values, the sample splitting lead to the asymptotic normality, which is different from the conventional  extreme value distributions. }We also note that the convergence rate  \(n_{\min}^{-\frac{v}{2v+4}}\) in Theorem \ref{theorem: ECT_asym_normal} 
can be improved to \(n_{\min}^{-1/2}\) by using the true variance instead of estimated \(\{\widehat{\sigma}_k^{(j)}\}\).}  

{ As Theorem \ref{theorem: ECT_asym_normal} suggests that 
 ${T}_{N,K}^{(j)}$ converges to $\mathcal{N}(0,1)$,   
{
the ECT rejects \( \mathcal{H}_{0}^{(j)} \)  } if $T_{N,K}^{(j)} > \Phi^{-1}(1-\alpha/p)$,  which controls the ECT's FWER at level \( \alpha \). }

 \begin{algorithm}
 \LinesNumbered
\KwIn{Samples $\{X_{k,i}\}$ from $K$ devices, $\gamma \in (0, 1)$,  level $\alpha$.}
\KwOut{Test statistics $\{T^{(j)}_{N,K}\}_{1 \leq j \leq p}$ for each dimension.}
\textbf{Broadcast} $\gamma$.

\ForPar{$1 \leq k \leq K$}{
   Each device $k$ use the sample split $D_k^{[1]}$ to compute the local estimate $\widehat{\theta}_k^{[1]}$ and transmit to central server.
}

\ForPar{$1 \leq j \leq p$}{
   At the central server, find out $\hat{k}^{(j)}_{\max}$,$\hat{k}^{(j)}_{\min}$ and broadcast them.
   At devices $k = \hat{k}^{(j)}_{\max}$ and $k = \hat{k}^{(j)}_{\min}$, compute the estimates $\widehat{\theta}_{k}^{(j),[2]}$ and the standard errors $ \widehat{SE}_{k}^{(j),[2]}$.
   Transmit them to the central server and compute the t-statistic $T^{(j)}_{N,K}$ via equation  \eqref{equation_ECT_Definition}.
}
 Reject  $\mathcal{H}_{0}^{(j)}$ if $T^{(j)}_{N,K} > \Phi^{-1}(1-\alpha/p)$.
\caption{Extreme Contrast Test Procedure}
\end{algorithm}

We also note that Theorem \ref{theorem: ECT_asym_normal}  imposes no restrictions on the 
relationship between $K$ and $n_{\min}$ while the re-normalized Wald statistic requires $K=o(n_{\min})$. It can be seen that the ECT is both communication and computation efficient. The overall procedure incurs a communication cost of $\mathcal{O}((K + 2)p+K)$, which is independent of the sample sizes. Let $\delta_{n_k}$ be the computation cost of the local estimates with sample size $n_k$, then the ECT requires the computation cost about $\mathcal{O}((K+2)p)+2\sum\limits_{k=1}^K\delta_{n_k/2}$. In contrast, calculating a Wald-type test for each dimension involves matrix multiplication and inversion of size $K$, resulting in a computational complexity of about $\mathcal{O}(K^3)+\sum\limits_{k=1}^K\delta_{n_k}$, which leads to higher computational costs with an increasing number of devices $K$. 

In order to evaluate the power of the ECT under the alternative, we denote $\Delta_{j}=\sup_{k_1\ne k_2}|\theta_{k_1}^{(j),*}-\theta_{k_2}^{(j),*}|$ { as the largest spacing among the $j$-components of parameters}. Define 
\begin{equation}\label{equation: D, w_{N,K,j}}
\Delta_{\max}=\sup\limits_{1\leq j\leq p } \Delta_j,\ \ \ w_{N,K}^{(j)} = T_{N,K}^{(j)}-\widetilde{T}_{N,K}^{(j)} 
= \frac{\theta_{\hat{k}_{\max}^{(j)}}^{(j),*}-\theta_{\hat{k}_{\min}^{(j)}}^{(j),*}}{\sqrt{(\widehat{SE}_{\hat{k}_{\max}^{(j)}}^{(j),[2]})^2 + (\widehat{SE}_{\hat{k}_{\min}^{(j)}}^{(j),[2]})^2}}.
\end{equation}
Since we have established the asymptotic normality of $\widetilde{T}_{N,K}^{(j)}$, $w_{N,K}^{(j)}$ is the leading term under the alternative hypothesis. The power of ECT hinges on selection consistency, that is, whether the selected random variables $\hat{k}_{\max}^{(j)}$ and $\hat{k}_{\min}^{(j)}$ can ensure that the corresponding true parameter $\theta_{\hat{k}_{\max}^{(j)}}^{(j),*}-\theta_{\hat{k}_{\min}^{(j)}}^{(j),*}$ is sufficiently large. To elucidate the power of ECT, we define the event ``approximately correct'' for the $j$-th dimension as
\[
\text{Approx}_{j} = 
\left\{ \theta^{(j),*}_{\hat{k}_{\text{min}}^{(j)}} \leq \min_k \theta_k^{(j),*} + \frac{\Delta_j}{3} \right\} \cap \left\{ \theta^{(j),*}_{\hat{k}_{\text{max}}^{(j)}} \geq \max_k \theta_k^{(j),*}-\frac{\Delta_j}{3} \right\}.
\] 
Recall the definition of $\sigma^-$ in Assumption \ref{assumption: Local Strong Convexity} that $\sigma_k^{(j)}\geq \sigma^-.$ It can be shown that under event $\text{Approx}_j$,  $w_{N,K}^{(j)}\geq\dfrac{\sqrt{n_{\min}}\Delta_j}{6\sqrt{2}\sigma^{-}}$ and the ECT can reject the null hypothesis with high probability. Denote $S=\{j|\mathcal{H}_1^{(j)}\text{ holds,} 1\leq j\leq p \}$  as the heterogeneous dimensions and $\bm{\widehat{S}}_{\text{ECT}}=\{j|1-\Phi(T_{N,K}^{(j)})<\alpha/p,1\leq j\leq p\}$ as the dimensions rejected by the ECT.  The following theorem analyze the family-wise error rate (FWER) and selection consistency of the ECT in finite sample, where  $C_1$ is a  constant free of $K,p,j,n$ and $\rho_0$ is a constant defined in Theorem \ref{theorem: ECT_asym_normal}.

\begin{theorem}[Finite Sample Analysis For ECT]\label{theorem: new_method_consistent}
Under Assumptions \ref{assumption:identifiability_compactness}–\ref{assumption: Local Strong Convexity} and the alternative $\mathcal{H}_{1}^{(j)},$  the event $\text{Approx}_{j}$ holds with probability at least $1-C_1 K n_{\min}^{-v} \Delta_j^{-2v}$, and: \[ \mathrm{P}(|\bm{S}\cap(\bm{\widehat{S}}_{\text{ECT}})^c|\geq 1) \leq C_1\cdot pK\Delta_{\max}^{-2v}n_{\min}^{-v}.\]  Moreover, 
 the ECT's FWER with nominal level $\alpha$ can be bounded by:
\[ \mathrm{P}(|\bm{\widehat{S}}_{\text{ECT}}\cap\bm{S}^c|\geq 1) \leq \alpha+\rho_0\cdot pn_{\min}^{-\frac{v}{2(v+2)}}.\]
 In conclusion, we have: 
 \[ \mathrm{P}(\bm{\widehat{S}}_{\text{ECT}} \ne \bm{S}) \leq \alpha+\rho_0\cdot pn_{\min}^{-\frac{v}{2(v+2)}}+ C_1\cdot pK\Delta_{\max}^{-2v}n_{\min}^{-v}.\]
\end{theorem}

Theorem \ref{theorem: new_method_consistent} informs that the ECT is consistent under the fix alternative when $n_{\min}\to\infty$ and $K=o(n_{\min}^v)$ where $v\geq 2$, which is a far weaker restrictions on the relationship between \( K \) and \( n_{\min} \) than the Wald-type test, which needs $K=o(n_{\min})$, and the maximum test with bootstrap calibration  \citep{zhenhualin2021} which assumes equal sample size $n$ and needs $K\preceq e^{\sqrt{\log n}}=n^{1/\sqrt{\log n}} \ll  n$ to control the size. Also, the ECT procedure shows communication and computation efficiency as it does not require bootstrap calibration and additional data transmission to determine the critical value, making it suitable for distributed settings.   Additionally, the consistency of ECT does not require the sample sizes are balanced which is  needed in some related works  \citep{jiagu2023,Huang_multitask_bandits}. It demonstrates that the proposed ECT has broad applicability in distributed settings where $K$ is much larger than $n_{\min}$ and the sample sizes are unbalanced.
 
 Theorem \ref{theorem: new_method_consistent} also suggests that the ECT's consistency requires $\Delta_{\max}^{2v}n_{\min}^v/K \to \infty$. Instead, the Wald test's consistency requires the divergence of  $\nu_{N,K}^{(j)}$ which has been defined in \eqref{equation: nu_j}. The difference motivates us comparing their power under the local alternatives next.

\subsection{Local Alternative Power Comparison }\label{section:Power Comparison}
 A question is that under what circumstance does each test exhibit higher power.  Previous works, such as \cite{junli_songxichen_two_sample_tests} and \cite{Cai2014}, suggest that tests based on the  quadratic form statistics, like the Wald-type test, exhibit higher power under dense alternatives but perform poorly under the sparse ones. On the other hand, tests formulated with the maximum form statistics, similar to the ECT, are more powerful against sparse alternatives but less effective for dense ones.  
In this section, we compare the power of the Wald-type test and the ECT under the local alternatives to study in which circumstances each method performs better. For simplicity, we assume  $n_k=n$ for all $k$. Suppose $\theta_0^{(j)}$ be the common value of  $\theta_k^{(j),*}$ under the null hypothesis $\mathcal{H}_0^{(j)}$. Suppose that under $\mathcal{H}_1^{(j)},$ $\theta_k^{(j),*}$ takes the value $\theta_0^{(j)}$ with probability $1-\epsilon_{\beta}$ and takes the value $\theta_0^{(j)}+\mu(K,n)$ with probability $\epsilon_{\beta}=K^{-\beta}>0$ for a sparsity parameter \(\beta\in(0,1)\). This leads to a specific form of the hypotheses in \eqref{the_hypothesis}:
\bea 
 && \mathcal{H}_0^{(j)}: \theta_k^{(j),*}=\theta_0^{(j)} \text{ for } 1\leq k\leq K  \text{ v.s. } \nn \\ 
&&  \mathcal{H}_1^{(j)}: \theta_k^{(j),*}  \overset{\text{i.i.d}}{\underset{1 \leq k \leq K}{\sim}}  (1-\epsilon_{\beta})\nu_{\theta_0^{(j)}}  + \epsilon_{\beta}\cdot \nu_{(\theta_0^{(j)}+\mu(K,n) )},\label{eq:gaussian_hypothesis}
\eea 
where $\nu_{\theta}$ stands for the point mass distribution at $\theta$.   We set \(\mu(K,n)=\sqrt{2c_n\log K/n}\) 
where \(c_n> 0\) is a constant controlling the magnitude of the shift and the constant $\beta$ represents the sparsity level of heterogeneous signals. We assume $\lim\limits_{n\to\infty}c_n=c_{\infty}\in(0,\infty).$

As both the Wald-type and the ECT statistics are asymptotic normal, comparing their power can be made by 
comparing their signal to noise (SNR) ratios. Denoting the standard deviation of the two statistics as $SE(W_{N,K}^{(j)})$ and $SE(T_{N,K}^{(j)})$,  
the SNRs are 
\[\text{SNR}_{Wald}^{(j)}=\dfrac{\EE W_{N,K}^{(j)}}{SE(W_{N,K}^{(j)})} \quad \hbox{and} \ \  \text{SNR}^{(j)}_{ECT}=\dfrac{\EE T_{N,K}^{(j)}}{SE(T_{N,K}^{(j)})}.\]

For easy presentation, we assume 
$\sigma_k^{(j)}=\sigma_j$ for $1\leq k\leq K$. 
 As both test statistics are standardized, their variances are approximately 1 and the SNRs are approximately the mean.   Specifically, it is shown in the Lemma \ref{lem:G_H_k1k2_local_noncentral} of the appendix that 
\bea\label{eq:SNR_wald} && \text{SNR}_{\text{Wald}}^{(j)}\approx \EE W_{N,K}^{(j)}\approx2^{1/2}c_n\sigma_j^{-2}K^{0.5-\beta}\log K 
\eea 
Let $\rho(\beta)=(1-\sqrt{1-\beta})^2$, which is the  detection boundary of the non-distributive maximum test 
for Gaussian data 
 \citep{2004Higher}. 
For the ECT statistic, it can be shown that, if 
$c_n>\sigma_j^{2}\rho(\beta)(1-\gamma)^{-1}$,  we can select the data blocks  with the largest heterogeneity correctly and $$\text{SNR}_{\text{ECT}}^{(j)}\approx \EE T_{N,K}^{(j)}\approx  \sigma_j^{-1}\sqrt{2\gamma c_n\log K},\text{ and 
 }$$ 
\be\label{eq:ratio of Wald and ECT}\dfrac{\text{SNR}_{\text{Wald}}^{(j)}}{\text{SNR}_{\text{ECT}}^{(j)}}\asymp \sigma_j^{-1}K^{(0.5-\beta)/2}\sqrt{2\gamma c_n \log K}.\ee

While the derivation of the exact leading order signal to noise ratio require the variances are the same for expedite derivation, 
this restriction can be relaxed in the power evaluation to allow different variances across data blocks in the next theorem.  Recall that 
$\sigma^-\leq \sigma_k^{(j)}\leq \sigma^+$ were the lower and upper bounds of the variances 
in Assumption \ref{assumption: Local Strong Convexity}. The following theorem reveals the asymptotic local power of the Wald test and the ECT. 
\begin{theorem}\label{Theorem: local_alternative_theorem}
For  testing the hypotheses  \eqref{eq:gaussian_hypothesis}, under Assumption  \ref{assumption:identifiability_compactness}-\ref{assumption: Local Strong Convexity}, if $K=o(\sqrt{n})$ as $K$ and $n\to\infty$, the power of the Wald test goes to 1 for $\beta\leq 0.5 $ and goes to $\alpha/p$ for $\beta>0.5.$ The power of the ECT goes to 1 for $c_{\infty}>\rho(\beta)(\sigma^{+})^2(1-\gamma)^{-1}$ and goes to $\alpha/p$ for $c_{\infty}<\rho(\beta)(\sigma^{-})^2(1-\gamma)^{-1}$. 
\end{theorem}

Theorem \ref{Theorem: local_alternative_theorem} shows that the ECT is more powerful than the Wald under the sparse local alternatives $(\beta>0.5)$ while the Wald is more powerful under the dense ones $(\beta\leq0.5).$  It is quite understandable to see the ECT's power is closely related to the detection boundary of the maximum test for the mean parameter in defining the minimal detectable signal strength, as the ECT has a flavor of the extreme test.
The  discount factor \((1-\gamma)^{-1}\) is due to the data splitting, may be mitigated by employing thresholding method  \citep{chen2019annals}. 
The proposed ECT is computation-efficient and have effective control on the FWERs, only requires $K=o(n_{\min}^v)$ for consistency when the maximum test with bootstrap calibration \citep{zhenhualin2021} 
 requires $K=o(e^{\sqrt{\log n}})<n$. Thus, the ECT is effective in avoiding  bias accumulation, facilitating the test in distributed setting with $K\gg n$. 

In particular, when $\beta>0.5$ and $c_{\infty}<\rho(\beta)(\sigma^{-})^2(1-\gamma)^{-1}$, both tests are inconsistent; when $\beta>0.5$ and $c_{\infty}>\rho(\beta)(\sigma^{+})^2(1-\gamma)^{-1}$, the ECT is consistent and  the Wald test is inconsistent, which means the ECT is more powerful;  when $\beta\leq 0.5$ and $c_{\infty}<\rho(\beta)(\sigma^{-})^2(1-\gamma)^{-1}$, the Wald test is consistent and the ECT is inconsistent, which means the Wald is more powerful; when $\beta\leq 0.5$ and $c_{\infty}>\rho(\beta)(\sigma^{+})^2(1-\gamma)^{-1}$, both the two tests are consistent but the Wald test is more powerful as can be seen by \eqref{eq:ratio of Wald and ECT}.


\section{A Combined Test} 
\label{section:combination_test} 

As the Wald test and ECT are powerful in different range of sparsity, and since the level of the sparsity is unknown 
in practice, we propose a combined test by integrating the re-normalized Wald and the ECT statistics 
to make the tests complimentary to each other. 
The objective is to construct a test that exhibits robust power across all levels of sparsity and can also the size for any value of $K$. \cite{Fisher_Combined_Xuelingzou} proposed a combined test statistic which adds the logarithm of  a quadratic test and a maximum form  test's 
 $p$-values under the  Gaussian distribution setting. In this section, we show that the Wald-type statistic $W_{N,K}^{(j)}$ is asymptotically independent to the ECT statistic $T_{N,K}^{(j)}$ in the context of M-estimation. Since they are both asymptotic normal, we can directly take a properly weighted sum of them which is also asymptotic normal. 

We firstly demonstrate that the two statistics are asymptotically independent.  
To investigate the correlation of \(W_{N,K}^{(j)}\) and \(T_{N,K}^{(j)}\), it is noted that the main source of correlation comes from the pairs \((\widehat{\theta}_{k^{(j)}_{\min}}^{(j)},\widehat{\theta}_{k^{(j)}_{\min}}^{(j),[2]})\) and \((\widehat{\theta}_{k^{(j)}_{\max}}^{(j)},\widehat{\theta}_{k^{(j)}_{\max}}^{(j),[2]})\) for $T_{N,K}^{(j)}$ and $W_{N,K}^{(j)}$ due to the fact that they are calculated by overlapping samples. Consequently, the main correlation is concentrated within \(\widehat{\theta}_{k^{(j)}_{\min}}^{(j)}\) and \(\widehat{\theta}_{k^{(j)}_{\max}}^{(j)}\) of the Wald-type statistics, which is calculated by $\mathcal{D}_{k^{(j)}_{\min}}$ and $\mathcal{D}_{k^{(j)}_{\max}},$ respectively. When the sample sizes are not too unbalanced, i.e.,  \(\max\limits_{1\leq k_1, k_2\leq K}\frac{n_{k_1}}{n_{k_2}}=o(K^{(1-1/v)})\), the correlation is negligible as \(K\to\infty\).

\begin{theorem}[Independence of Wald and ECT]\label{theorem:tilde_W_NK_normal_independence_ECT}
    Under  Assumptions \ref{assumption:identifiability_compactness}-\ref{assumption: Local Strong Convexity} and the null hypothesis \(\mathcal{H}_{0}^{(j)}\), suppose that  \(K = o(n_{\text{min}})\) and  \(\max\limits_{1\leq k_1, k_2\leq K}\frac{n_{k_1}}{n_{k_2}}=o(K^{(1-1/v)})\). 
   Then, for each fixed dimension \(1\leq j \leq p\) and for any \(t, s \in \mathbb{R}\),  \(\mathrm{P}({W}^{(j)}_{N,K}\leq t,T^{(j)}_{N,K}\leq s)\to\Phi(t)\Phi(s)\) as \(K \to\infty\) and \(n_{\text{min}}\to\infty\).  Moreover, $TW_{N,K}^{(j)}\to N(0,1)$ as $K,n_{\min}\to\infty.$
\end{theorem}

The theorem suggests that $T^{(j)}_{N,K}$ and $W^{(j)}_{N,K}$ are asymptotically independent, which lead us to 
consider a weighted statistic of them: \begin{equation}\label{combination_TW_NK_Definition}TW_{N,K}^{(j)} = \dfrac{W^{(j)}_{N,K}\cdot r_{N,K} + T_{N,K}^{(j)}}{\sqrt{(r_{N,K})^2 + 1}},\end{equation} 
where $r_{N,K}$ is a weight. The asymptotic independence between the two involved test statistics ensures that the combined statistic $TW_{N,K}^{(j)}$ is asymptotically standard normally distributed. Since the Wald and the ECT statistics are both standardized, we may set $r_{N,K}= 1$ for  equally  weighted sum of the two statistics. However, to avoid size distortion caused by ${W}^{(j)}_{N,K}$ { in the case of larger $K$},  
{ we assign \(r_{N,K}=\min\left\{\dfrac{n_{\text{min}}}{K \log K}, 1\right\}\) to allow down weight the Wald statistic as its bias is harder to control for larger $K$. 
Thus, { the combined test rejects}  \(\mathcal{H}_{0}^{(j)}\) if \(TW_{N,K}^{(j)} > \Phi^{-1}(1-\alpha/p)\) to control the FWER.

From \eqref{combination_TW_NK_Definition} and Theorem \ref{Theorem: local_alternative_theorem}, we know that the signal of the combined statistic mainly comes from the Wald when $\beta\leq0.5$ and from the ECT when $\beta>0.5$ under the local alternative. Let $\text{SNR}_{\text{Combined}}^{(j)}=\dfrac{\EE \ TW_{N,K}^{(j)}}{SE(TW_{N,K}^{(j)})}$ be the signal to noise ratio of the combined test. It may be shown that  $\text{SNR}^{(j)}_{\text{Combined}}\succeq (r_{N,K}^2+1)^{-1/2}(r_{N,K}\cdot\text{SNR}^{(j)}_{\text{Wald}}+\text{SNR}^{(j)}_{\text{ECT}}).$

The following theorem provides the power performance of the combined test. 

\begin{theorem}[The combined test's local power]\label{theorem:combined_local_power}
   For the local alternative defined in  \eqref{eq:gaussian_hypothesis} where $K=o(\sqrt{n}),\ r_{N,K}=1$ and $K,n\to\infty$, 
 the combined test's power goes to 1 if either $\beta\leq 0.5$ or $c_{\infty}>\rho(\beta)(\sigma^{+})^2(1-\gamma)^{-1}$,  and goes to $\alpha/p$ if $c_{\infty}<\rho(\beta)(\sigma^{-})^2(1-\gamma)^{-1}$ and $\beta> 0.5$ hold simultaneously. 
\end{theorem}

The theorem reveals that the local consistency of the combined test is ensured if $\beta \le 0.5$ where the Wald test is consistent or $c_{\infty}>\rho(\beta)(\sigma^{+})^2(1-\gamma)^{-1}$ for $\beta > 1/2$ where the ECT  is consistent.  Consequently, the combined test has the advantages of both the Wald test and the ECT, exhibiting  robust power under both dense and sparse alternatives (heterogeneity). Denote the set of dimensions rejected by the combined test as $\bm{\widehat{S}}_{\text{Combine}} \subset [p]$.   The following theorem provides the combined test's consistency under the fixed alternative defined in \eqref{the_hypothesis}.
\begin{theorem}[Combined test's consistency]\label{theorem: combine_method_consistent}
Under the conditions of Theorem \ref{theorem:tilde_W_NK_normal_independence_ECT}  and  the fixed alternative hypothesis \(\mathcal{H}_{1}^{(j)}\),  
$\mathrm{P}(j\in \bm{\widehat{S}}_{\text{Combine}})\to 1$ when
$r_{N,K}\cdot\nu_{N,K}^{(j)}\to\infty$ or \(\Delta_{\max} \succeq K^{1/(2v)}n_{\min}^{-1/2}\), where $\nu_{N,K}^{(j)}$ and $\Delta_{\max}$ are defined in \eqref{equation: nu_j}  and \eqref{equation: D, w_{N,K,j}}.
\end{theorem} 

Theorem 7 suggests that when $K=o(n_{\min}),$ the combined test is consistent when the ECT or the Wald test is consistent, i.e., \(\Delta_{\max} \succeq K^{1/(2v)}n_{\min}^{-1/2}\) or $\nu_{N,K}^{(j)}\to\infty$. As shown by Lemma \ref{lem:min_square_diff} in the Appendix, these two conditions are more relaxed than the respective conditions required by either the Wald test or the ECT, indicating wider applicability of the combined test. 

\section{Simulation Study}\label{section:simulation}

{ In this section, we conducted simulations to compare the family-wise error rate 
(FWER) and power of the three proposed methods: the  Wald-type test introduced in Section \ref{section:The Wald-type Test}, the ECT introduced in Section \ref{section_ECT} and the combined test introduced in Section \ref{section:combination_test} with the weight being set as $r_{N,K}=\dfrac{n}{K^{1.1}}$. 

Throughout the simulation experiment, the results of each setting were based on $B =500$ replications and were conducted in R.
Denote  \(\bm S \) as  the set of dimensions where the alternative hypothesis held,  \(\bm{S}^c\) as the true nulls and \(\bm{\widehat{S}}^c\) as the rejected nulls by some testing procedures. The empirical FWER was defined as \(\mathrm{P}_B\left(\mathbb{I}\left\{|\bm{\widehat{S}}^c \cap \bm{S}^c| \geq 1 \right\}\right)\) and the power was defined as  \(\mathrm{P}_B\left(|\bm{\hat{S}}|\right)\), where \(\mathrm{P}_B\) represented the empirical distribution based on \(B = 500\) replications. For power evaluation, the empirical critical values from the simulation under the null hypothesis were used, ensuring that the empirical FWER of all proposed tests were exactly equal to \(\alpha=0.05\). We consider the number of data blocks $K\in\{10,25,50,75,100,250,350,500,750,1000\}$. For each $1\leq k\leq K,$ the local observations $\{(X_{k,i},Y_{k,i})\}_{i=1}^n$ with $n = 500$ were independently generated from the generalized linear model (GLM):
\be
\EE(Y_{k,i}|X_{k,i}) = g^{-1}(X_{k,i}^{\top}\theta_k^*), \quad X_{k,i}^{(j)} \stackrel{\text{i.i.d}}{\underset{1\leq j\leq p}{\sim}}\mathrm{Pareto}(\eta = 4.1, \zeta = 2),\label{eq: GLM}
\ee
where $X_{k,i}^{(j)}$ is the $j$-th dimension of $X_{k,i}$ and the specific forms of the link functions $g(\cdot)$ will be given in Sections \ref{Section:Simulation_linear} and \ref{Section:Simulation_logistic}, respectively. The probability density function of 
 the Pareto distribution is: 
\be
f(x;\eta,\zeta) = \begin{cases} \dfrac{\eta \zeta^\eta}{x^{\eta + 1}} & \text{if } x \geq\zeta \\ 0 & \text{if } x < \zeta \end{cases}.\nn
\ee
This distribution is heavy-tailed with fourth finite moment, which is aligned with our Assumption \ref{assumption:Smoothness for Hessian}. The dimension of the parameter $p$ was set to be $3$. The parameter vectors \(\{\theta_{k}^{*}\}_{1 \leq k \leq K}\) were constructed with a sparsity level controlled by the parameter \(\beta \in (0,1)\) as defined in Section \ref{section:Power Comparison}. Here, a larger \(\beta\) indicated greater heterogeneity among the parameter vectors. Specifically, the first and second components of each parameter, \(\{\theta_{k}^{(1),*}, \theta_{k}^{(2),*}\}_{1 \leq k \leq K}\), were set to 1 for all \(k\). For the dimension \(j = 3\), \(\{\theta_{k}^{(j),*}\}_{1 \leq k \leq K}\) were i.i.d  with probability \(K^{-\beta}\) equals to \(1 +  4.5K^{(\beta-0.5)/2}n^{-1/2} > 1\), and with probability \(1-K^{-\beta}\) of being 1 to guarantee the power of Wald-type test remains stable in different sparsity levels. 
}

\subsection{Linear Regression Experiment}\label{Section:Simulation_linear}
We first consider linear regression. The observations were independently sampled according to the following GLM model:
 \[
 Y_{k,i} = X_{k, i}^\top \theta_{k}^{*} + \epsilon_{k,i},
\]
with the specific form of the  link function in \eqref{eq: GLM} is  $g(x)=x$. We set  $\epsilon_{k,i}\stackrel{\text{i.i.d}}{{\sim}}\mathrm{Pareto}(\eta = 4.1, \zeta = 2)-\eta \zeta/(\zeta-1)$ as the noise with mean zero. We show the FWER result for $K\geq 100$ and  power results for \(K = 250\) in Figure \ref{figure: Power_and_FWER_5_K_250_linear} for simplicity, while the remaining results are shown in Figures \ref{fig:K<=100_fwer} and \ref{more_power_linear_results_Kgeq100} in the Appendix.  

For the FWER control experiment in the left panel of Figure \ref{figure: Power_and_FWER_5_K_250_linear}, the Wald-type test exhibited FWER values exceeding the ideal level of 0.05 but remained stable. The combined test and the ECT controlled FWER successfully across different values of \(K\).

For power performance in the right panel of Figure \ref{figure: Power_and_FWER_5_K_250_linear}, we ensured the empirical FWER of the proposed tests were exactly equal to the nominal significance level \(\alpha\) with fairness. In scenarios with dense heterogeneous signals (\(\beta \leq 0.5\)), the Wald-type test demonstrated the highest power, followed by the combined test, which maintained robust power across different sparsity levels. In local alternative settings with dense heterogeneity, the ECT exhibited low power, as expected from Theorem \ref{Theorem: local_alternative_theorem}, which shows inconsistency under such conditions. In sparse signal scenarios (\(\beta > 0.5\)), the combined test performed best on power,
followed by the ECT. It's also worth mentioning that the combined test could sometimes be more powerful than the ECT and the Wald test simultaneously in the empirical results, particularly when the ECT and the Wald test had similar power. 
\begin{figure}
    \centering
    \caption{The FWERs  (left panel) and the power (right panel) of the Wald (green triangle), the ECT (blue circle) and the Combined tests (yellow cross) with respect to the block size $K$ (left panel) and the sparsity level $\beta$ for $K=250$ (right panel).}
       \includegraphics[width=1\linewidth]{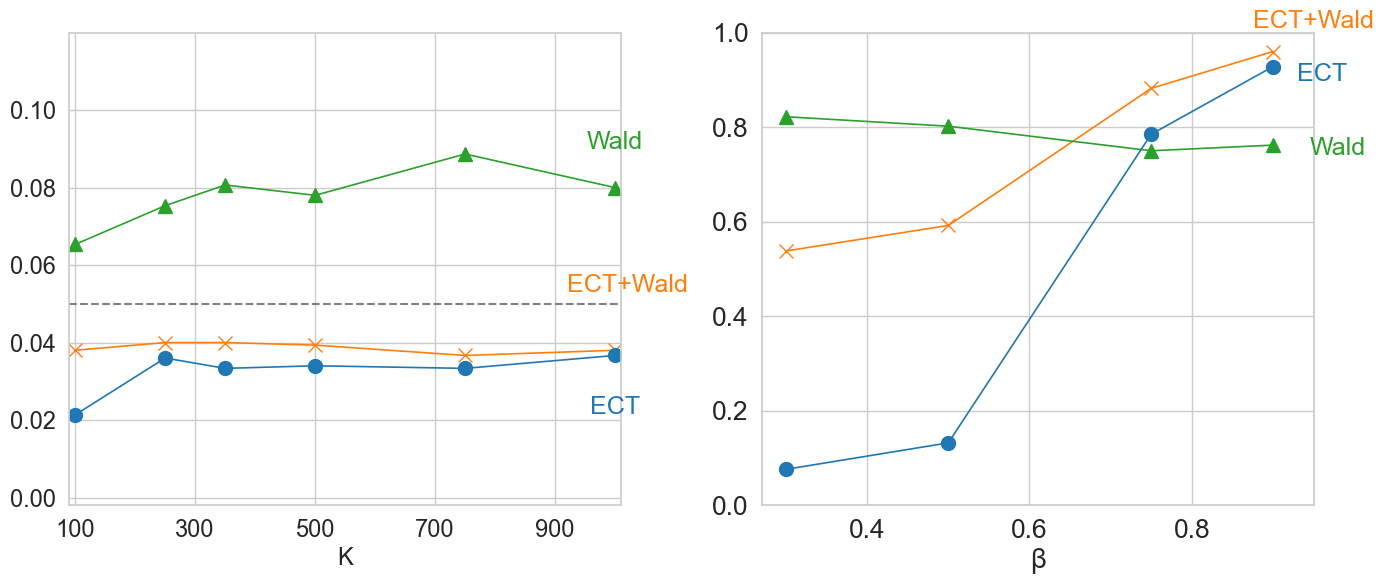}
    \label{figure: Power_and_FWER_5_K_250_linear}
\end{figure}

\subsection{Logistic Regression Experiment}\label{Section:Simulation_logistic}
In the previous sub-section, we compared the proposed methods under linear regression model, where the estimates are unbiased and the remainder term $R_k$ defined in \eqref{eq:Def_for_R_k} equals to zero. We now investigated the performance of the proposed methods in another GLM model where the estimates are consistent but involves non-zero remainder terms $R_k$ as discussed at the end of Section \ref{section:The Wald-type Test}.
The observations were independently sampled according to the logistic regression model:
 \[
 P(Y_{k, i} = 1 | X_{k, i}) = \frac{\exp(X_{k, i}^\top \theta_{k}^{*})}{1 + \exp(X_{k, i}^\top \theta_{k}^{*})},
 \] with the link function defined in \eqref{eq: GLM} is  $g(x)=\log(x/(1-x)).$  The results of the logistic regression experiment with \(n = 500\) samples and \(K\) varying from 100 to 1000 are presented in Table \ref{tab:Wald_coverage_results} and Figure \ref{figure_log_insample} while the remaining results are shown in Figures \ref{fig:K<=100_fwer} and \ref{more_power_logis_results_Kgeq100}. 

\begin{figure}[ht]
    \centering
    \caption{The FWER results for various $K$ (left panel) and power results for various sparsity level $\beta$ and $K=250$ (right panel) in logistic regression experiment for the proposed methods. The settings are followed by Figure \ref{figure: Power_and_FWER_5_K_250_linear}.}
     \includegraphics[width=1\linewidth]{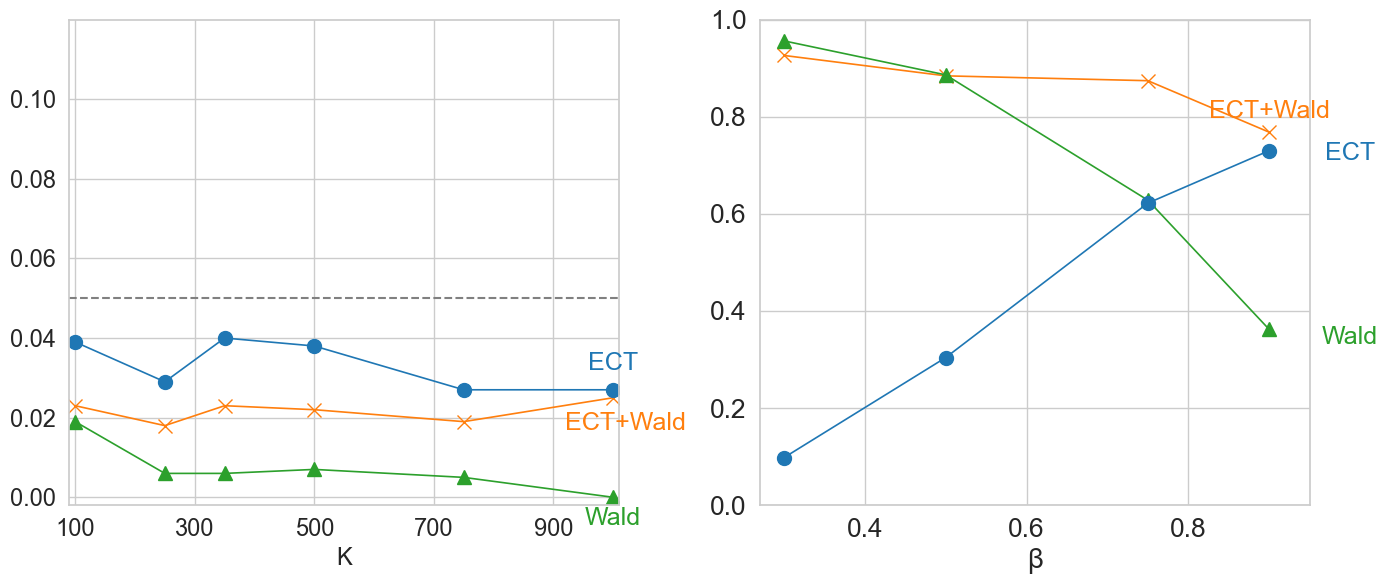}
    \label{figure_log_insample}
\end{figure}
The FWER results shows that the Wald-type test exhibited an extremely low FWER for \(K \geq 250\). In fact, when \(K \succeq n\), the asymptotic standard normality of the re-normalized Wald statistic is no longer theoretically guaranteed. This manifests as the statistic becoming too small when $K$ grows in logistic regression experiments. The empirical coverage of the Wald statistic is shown in Table \ref{tab:Wald_coverage_results}, where it can be seen that as \(K\) increased, the empirical coverage increased and far exceeded the nominal coverage. This phenomenon is attributed to the underestimated variances and the cumulative bias from such underestimations, as previously discussed in the concluding remarks of Section \ref{section:The Wald-type Test}.

\begin{table}[ht]
\centering
\caption{Empirical coverages for the Wald-type statistic $W_{N,K}^{(1)}$ (left panel) and the  combined test statistic $TW_{N,K}^{(1)}$ (right panel)  under the null hypothesis $\mathcal{H}_0^{(1)}$ with respect to different block sizes $K$, and sample size $n = 500$ and $p = 3$ for the logistic regression model at nominal coverage levels   $\tau = 0.95, 0.9, 0.1, 0.05$.}
\label{tab:Wald_coverage_results}
\renewcommand{\arraystretch}{1.3}
\setlength{\tabcolsep}{8pt} 
\begin{tabular}{crrrrrrrrr}
\toprule
& \multicolumn{4}{c}{Empirical Coverage For $W_{N,K}^{(1)}$} & & \multicolumn{4}{c}{Empirical Coverage For $TW_{N,K}^{(1)}$} \\
\cline{2-5} \cline{7-10}
$K$ &  0.95 &   0.9 &   0.1 &  0.05 & &  0.95 &   0.9 &   0.1 &  0.05 \\
\midrule
  10 & 0.955 & 0.927 & 0.055 & 0.003 & & 0.960 & 0.907 & 0.129 & 0.065 \\
  20 & 0.973 & 0.940 & 0.081 & 0.022 & & 0.964 & 0.919 & 0.101 & 0.047 \\
  30 & 0.958 & 0.926 & 0.093 & 0.028 & & 0.965 & 0.908 & 0.117 & 0.057 \\
  40 & 0.964 & 0.929 & 0.106 & 0.041 & & 0.965 & 0.919 & 0.138 & 0.066 \\
  50 & 0.965 & 0.931 & 0.111 & 0.049 & & 0.965 & 0.925 & 0.127 & 0.064 \\
  75 & 0.972 & 0.946 & 0.102 & 0.048 & & 0.975 & 0.927 & 0.142 & 0.073 \\
 100 & 0.966 & 0.935 & 0.118 & 0.057 & & 0.968 & 0.940 & 0.138 & 0.079 \\
 250 & 0.987 & 0.965 & 0.169 & 0.078 & & 0.968 & 0.930 & 0.163 & 0.095 \\
 350 & 0.986 & 0.971 & 0.188 & 0.090 & & 0.966 & 0.928 & 0.140 & 0.060 \\
 500 & 0.989 & 0.970 & 0.213 & 0.101 & & 0.962 & 0.921 & 0.132 & 0.072 \\
 750 & 0.997 & 0.981 & 0.260 & 0.141 & & 0.957 & 0.908 & 0.123 & 0.060 \\
1000 & 0.995 & 0.985 & 0.303 & 0.184 & & 0.964 & 0.909 & 0.136 & 0.067 \\
1250 & 0.998 & 0.993 & 0.312 & 0.176 & & 0.957 & 0.889 & 0.120 & 0.060 \\
1500 & 0.996 & 0.989 & 0.363 & 0.222 & & 0.958 & 0.913 & 0.115 & 0.061 \\
\bottomrule
\end{tabular}
\end{table}
Although the empirical coverage grew too large, indicating that the Wald is no longer appropriate when \(K \succeq n\). For \(K \leq 100\) scenarios that $K$ is smaller than $n/2$, Wald test performed well in terms of coverage, which matched our Theorem \ref{Wald consistency} that its bias accumulation is negligible when \(K = o(n)\). The ECT and combined test consistently exhibited superior FWER control across all values of \(K\). As shown in  \eqref{combination_TW_NK_Definition}, the combined test statistic incorporates a proper weight of the Wald statistic. The weight \(r\) decreased with increasing \(K/n\) avoided the combined test to be affected by the Wald's bias accumulation. As a result, the combined statistic  exhibited ideal coverage across all values of \(K\), which matched the conclusion of Theorem \ref{theorem: combine_method_consistent}.

The logistic regression power result exhibited trends similar to those shown in the right panel of Figure \ref{figure: Power_and_FWER_5_K_250_linear}. Under dense heterogeneity, the Wald-type test demonstrated the best performance. In sparse settings where $\beta > 0.5$, the ECT outperformed the Wald-type test.

For the combined test's performance shown in Figure \ref{figure_log_insample}, under sparse alternative, it exhibited greater power than the Wald test and ECT. In dense alternative where $\beta\leq 0.5$, the combined test matched the Wald test's power while significantly surpassing the ECT.  

Overall, the ECT and combined test showed ideal FWER control across all the  scenarios, which matched  the conclusions of Theorems \ref{theorem: ECT_asym_normal} and \ref{theorem: combine_method_consistent}. For power experiment, the ECT exhibited higher power than the Wald-type test under sparse alternative and the power of the combined test is robust against varying sparsity levels, sometimes outperforms the ECT and the Wald-type test simultaneously especially when Wald has close power. This robustness is valuable especially when the degree of sparsity is unknown, alleviating the need for estimating the unknown sparsity level.

\section{Real Data Analysis}\label{section:case_study}
In this study, we analyzed a publicly available Avazu Click-Through Rate(CTR) Prediction dataset from Kaggle (\url{https://www.kaggle.com/c/avazu-ctr-prediction/data}). The dataset, relevant to online advertising where CTR is a key metric for ad performance evaluation, contains 11 days of Avazu ad interactions and over 40 million data points with features like position of the advertisement on the webpage and the domain of the app where the ad was
displayed etc. 

Our primary objective was to evaluate which features have  heterogeneous effects on CTR across different site categories (e.g., news, entertainment, etc.) in a distributed data environment. Specifically, we conducted  hypothesis testing to determine whether the following features: click time, banner position, weekend or weekdays, device connection type, as well as the masked features {C18}, {C19} and {C21}—affect CTR differently across various site categories. These features were selected due to they are relatively balanced distributed  while other masked features are highly unbalanced distributed within the sample.

The reason for using site categories as the grouping variable was that features such as banner position may have heterogeneous effects to different site categories due to variations in user behavior, content engagement and visual hierarchy preferences. For instance, a prominent top-banner placement might perform well on news websites where users engage in linear content consumption but may be less effective on entertainment platforms where users focus on interactive media. Given these potential differences, heterogeneity of these effects across different sites is crucial for optimizing advertising strategies. 

 The masked site categories are:``3e81413'', ``f66779e6'',``75fa27f'', ``335d28a8'', ``76b2941d'' and ``c0dd3be3'', with sample sizes of 7,377,208, 3,050,306, 252,451, 160,985, 136,463, 104,754 and 42,090, respectively.
The feature banner position represents the position of an advertisement on the webpage, directly influencing its visibility and, consequently, the likelihood of user engagement. Similarly, the device connection type, which reflects the type of network connection accessed by the user (e.g., mobile network or Wi-Fi), may significantly affect clicking behavior due to variation in network speed and stability.

For each site category \(1 \leq k \leq K\), we fitted a logistic regression model to estimate the CTR heterogeneity. To account for temporal patterns, the hour of the day was encoded as \({cos\_hour}\) using a cosine transformation \(\cos(2\pi \cdot {hour}/24)\). The model is defined as:
\[
\log \left( \frac{\mathrm{P}({click} = 1)}{\mathrm{P}({click} = 0)} \right) = \beta_{0k} + \sum_{j=1}^{7} \beta_{jk} X_j,
\]
where \(X_j\) includes {weekend}, {cos\_hour}, and other selected features. Our primary task was to evaluate whether the regression coefficients, which quantify the influence of each feature on {click}, exhibit heterogeneity across different {site\_category} groups. In other words, we aimed at testing the following hypotheses, for \(1 \leq j \leq 7\),

\[
\mathcal{H}_{0}^{(j)} : \beta_{j1} = \beta_{j2} = \dots = \beta_{jK} \quad \text{vs} \quad \mathcal{H}_{1}^{(j)} : \beta_{j1}, \dots, \beta_{jK} \quad \text{are not all equal}.
\]

\begin{table}[ht]
\centering
\caption{Site categories with the maximum and minimum statistics for each feature.} 
\label{table: max_estimate_site_categories} 
\renewcommand{\arraystretch}{1.2} 
\setlength{\tabcolsep}{10pt} 

\begin{tabular}{lcc} 
\toprule 
\textbf{Feature} & \textbf{Max Site Category} & \textbf{Min Site Category} \\
\midrule
weekend & c0dd3be3 & 76b2941d \\
cos\_hour & 28905ebd & 76b2941d \\
device\_conn\_type & 76b2941d & 335d28a8 \\
C18 & 28905ebd & f66779e6 \\
C19 & 335d28a8 & 75fa27f6 \\
C21 & f66779e6 & 76b2941d \\
banner\_pos & 76b2941d & 28905ebd \\
\bottomrule 
\end{tabular}
\end{table}

\begin{table}[ht]
\centering
\caption{ECT and re-normalized Wald test statistics' heterogeneity detection results.}
\label{table: case_study_test_results}
\renewcommand{\arraystretch}{1.3}
\setlength{\tabcolsep}{6pt}

\begin{tabular}{lcccc}
\toprule
\multirow{2}{*}{\textbf{Feature}} & \multicolumn{2}{c}{\textbf{ECT}} & \multicolumn{2}{c}{\textbf{Wald}} \\
\cline{2-5} 
& Statistic & Heterogeneity Detected & Statistic & Heterogeneity Detected \\
\midrule
weekend         & 2.009671  & FALSE                  & 38.01507  & TRUE \\
cos\_hour       & 5.226893  & TRUE                   & 97.49311  & TRUE \\
device\_conn\_type & 8.97209   & TRUE                   & 94.409    & TRUE \\
banner\_pos     & 10.54358  & TRUE                   & 1494.992  & TRUE \\
C21             & 15.47275  & TRUE                   & 2713.357  & TRUE \\
C19             & 23.68508  & TRUE                   & 4223.734  & TRUE \\
C18             & 37.71257  & TRUE                   & 2572.289  & TRUE \\
\bottomrule
\end{tabular}
\end{table}
We tested the heterogeneity via ECT and the Wald-type test, and the results are shown in Tables \ref{table:  case_study_test_results}. 
First, both the ECT and Wald tests consistently identified the weekend feature as demonstrating the lowest heterogeneity in CTR effects across all site categories. This finding was aligned with the fact that user engagement with advertisements remains uniformly high during weekends, as increased leisure time across diverse user segments leads to more consistent CTR performance regardless of site type. Second, for the remaining six features, heterogeneity ranking orders were significantly different between the ECT and Wald tests. This difference is because ECT is more powerful when extreme difference is large whereas Wald-type test emphasizes dense heterogeneity signals. For instance, the cos\_hour feature ranked second-lowest by ECT but third-lowest by Wald, indicating that hourly CTR fluctuations were less extreme yet more densely heterogeneous across sites. 

In conclusion, both our proposed statistics, ECT and Wald are communication-efficient, making them suitable for distributed hypothesis testing with ideal statistical power. However, they possessed distinct characteristics. The testing methods we introduce not only identify heterogeneous and homogeneous parameters but can also help characterizing the nature of heterogeneity—whether it is dense  or sparse via using ECT or re-normalized Wald test. 
For example, in distributed learning and federated learning, when  aggregation the data, we should exclude clients with extreme heterogeneity—specifically those with sparse but extremely large differences. Conversely, if heterogeneity is dense and uniformly distributed across clients, group-wise aggregation may be more appropriate. This approach balances the need to mitigate the impact of outliers while leveraging structured heterogeneity for more robust model convergence. This capability is particularly valuable for federated learning, as it provides a suggestion for implementing personalized models by identifying whether global model uniformity or localized adaptations to some groups are needed.

\section{Conclusions}\label{Section:conclusions}
This paper considers testing for parameter heterogeneity in the context of distributed M-estimation. A re-normalized Wald-type test is proposed and shown to be consistent when $K = o(n_{\min})$ and $\nu_{N,K}^{(j)}\to\infty$ for $j\in\mathcal{H}_1^{(j)}$ as shown in Theorem \ref{Wald consistency}.  By introducing a novel sample splitting technique, an extreme contrast test (ECT) is derived to accommodate a much larger number of blocks $K$. The test statistic for the ECT is asymptotically normal under the null without any restriction on $K$  as shown in Theorem \ref{theorem: ECT_asym_normal}, and is consistent as long as $K  = o(n_{\min}^{v})$  as shown in Theorem \ref{theorem: new_method_consistent}, where $v$ is the moment parameter defined in Assumption \ref{assumption:Smoothness for Hessian}. By contrast, traditional extreme value based test statistics are usually asymptotically Gumbel, which implies their slow convergence rates and the necessity of bootstrap calibration of the corresponding critical values. Under the local alternative framework, it is shown in Theorem \ref{Theorem: local_alternative_theorem} that the Wald-type test is more powerful for dense alternatives while the ECT is more effective for sparse alternatives. 
To enjoy the best of both worlds, a combined test is formulated by a weighted average of both test statistics based on the asymptotic independence of the Wald and the ECT statistic. The combined test has robust power under different levels of sparsity comparing to using the ECT (Wald-type test) alone which has inferior  power under the dense (sparse) alternatives as shown in Theorem \ref{theorem:combined_local_power}.
All the three tests are communication-efficient in the sense that only $\mathcal{O}(p)$ transmission is required for each of the data blocks to construct the test statistics.

The test for heterogeneity is a fundamental research problem in distributed learning, which offers insights for many relevant downstream learning tasks such as the federated learning and multi-task learning. How to explicitly leverage testing results to optimize aggregation strategies in federated learning paradigms we leave to future research.

\vskip 0.2in

\bibliographystyle{apalike}
\bibliography{reference}
\newpage

\vskip 0.2in
\appendix

\renewcommand{\thefigure}{\Alph{section}.\arabic{figure}}


\setcounter{table}{0}
\renewcommand{\thetable}{\Alph{section}.\arabic{table}}

The Appendix is organized as follows. Appendix \ref{section:discussion Recommended_gamma} illustrate the reasons for choosing $\gamma=2/3$ in the ECT's data splitting procedure.  Appendix \ref{app:num_experiment_details} and \ref{appendix: real study details} provides detailed descriptions, supplemental tables and figures for numerical experiments and real study. Appendix \ref{appendix: main_theorem} and \ref{appendix: lemma proofs} presents the proofs of the theorems and lemmas in the paper.

\section{Recommended Proportion in ECT}\label{section:discussion Recommended_gamma}
\setcounter{figure}{0}  
To better detect heterogeneity, we consider how to choose \(\gamma\) to maximize the test's power. We analyze the power in the context of a Gaussian distribution with equal variance. Consider a sequence of i.i.d. random variables \( X_k^{1}, X_k^{2}, \dots, X_k^n \sim \mathcal{N}(\mu_k, 1) \), for \( 1 \leq k \leq K \).

 If there are \( K^{1-\beta} \) indices with a common mean \( \mu_k = \mu > 0 \), while the remaining means \( \mu_k = 0 \), where \( 0.5 \leq \beta  < 1 \) represents the sparsity level. We define the ordering statistic as follows:
\[
T_k^{\text{rank}} = \frac{1}{\sqrt{n(1-\gamma)}} \left(\sum\limits_{i=1}^{[n(1-\gamma)]} X_k^i\right),
\]
where the test part is given by:
\[
T_k^{\text{test}} = \sum\limits_{i=[n(1-\gamma)]+1}^{n} X_k^i \cdot \frac{1}{{n\gamma}},
\]
with the assumption that \( n\gamma \) is an integer and \( 0 < \gamma < 1 \).

The Type II error of this method originates from two sources:

1. The error in the ordering process, where incorrect items are selected. This means there exists at least one random variable $X_k^i$ generated from $\mathcal{N}(0,1)$ is larger than all the variables generated from $\mathcal{N}(\mu,1).$ Take the union bound, the error in the ordering process can be bounded by:
\[
\leq (K-K^{1-\beta}) \cdot \mathrm{P}\left(\mathcal{N}(0,1) > K^{1-\beta} \text{ instances of } \mathcal{N}(\sqrt{n(1-\gamma)} \mu, 1)\right),
\]
which approximates to:
\[
\Phi\left( -\sqrt{n(1-\gamma)}\mu-\sqrt{2(1+\epsilon)} \log  K \right).
\]

2. The error in the testing process, where the correct rejection does not occur, even if the rank selection is correct. This error is given by:
\[
\mathrm{P}\left(\mathcal{N}(\sqrt{n\gamma} \mu, 2) < 0 \right) = \Phi \left( -\sqrt{\frac{n\gamma}{2}}\mu \right).
\]

We observe that as \(\gamma\) increases, the first error term decreases while the second error term increases. The total error is the sum of these two parts. To minimize the total error, we assume the two parts are approximately equal. When \( n \) and \( \mu \) are large, we can neglect the \(\log  K\) term, leading to the relationship:
\[
-\sqrt{n(1-\gamma)}  \approx -\sqrt{\frac{n\gamma}{2}},
\]
which gives \(\gamma \approx \dfrac{2}{3}\).

\section{Numerical Experiment Details}\label{app:num_experiment_details}

\setcounter{figure}{0}
\setcounter{table}{0}

\subsection{More Power Result}\label{Appendix:More Power Result}
Figure \ref{fig:K<=100_fwer} shows FWER results when $K\leq100<n$. It shows that all the proposed tests show ideal FWER control.

Figures  \ref{more_power_linear_results_Kgeq100} and \ref{more_power_logis_results_Kgeq100} shows more power results for GLM models experiments with $n=500$ and $K\in\{100,250,350,500,750,1000\}$. For linear regression model, the ECT and the combined statistic showed superior power when $\beta=0.75,0.9$ as $K$ grew more than $500$ while Wald-type test showed ideal power when $\beta\leq0.5$. For logistic regression model, the combined test still showed robust power under different sparsity levels. As $n$ is fixed and $K$ grows, the ECT dominates the combined test.

\begin{figure}
    \centering
    \caption{The FWER results for  linear(left) and logistic(right)  regression experiment when $K\leq100$ for five methods: ECT(blue circle), Wald(green triangle) and combination of the ECT and Wald(yellow cross). The dashed line represents nominal level $\alpha=0.05 $.}
    \includegraphics[width=1\linewidth]{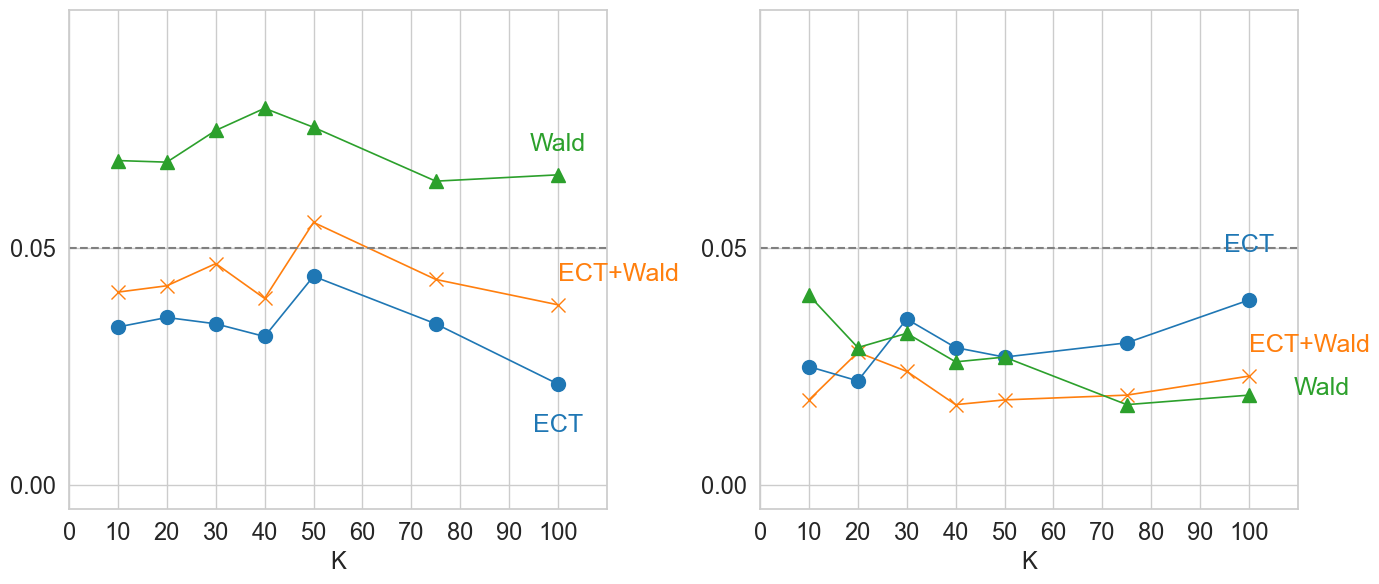}
    \label{fig:K<=100_fwer}
\end{figure}
\begin{figure}
    \centering
    \caption{The linear regression results with \(p = 3\) and \(n = 500\). The left plot shows power results for different methods: ECT(blue circle), Wald(green triangle), combination of the ECT and Wald(yellow cross) across various values of \(K\in\{100,250,350,500,750,1000\}\).}
     \includegraphics[width=0.95\linewidth]{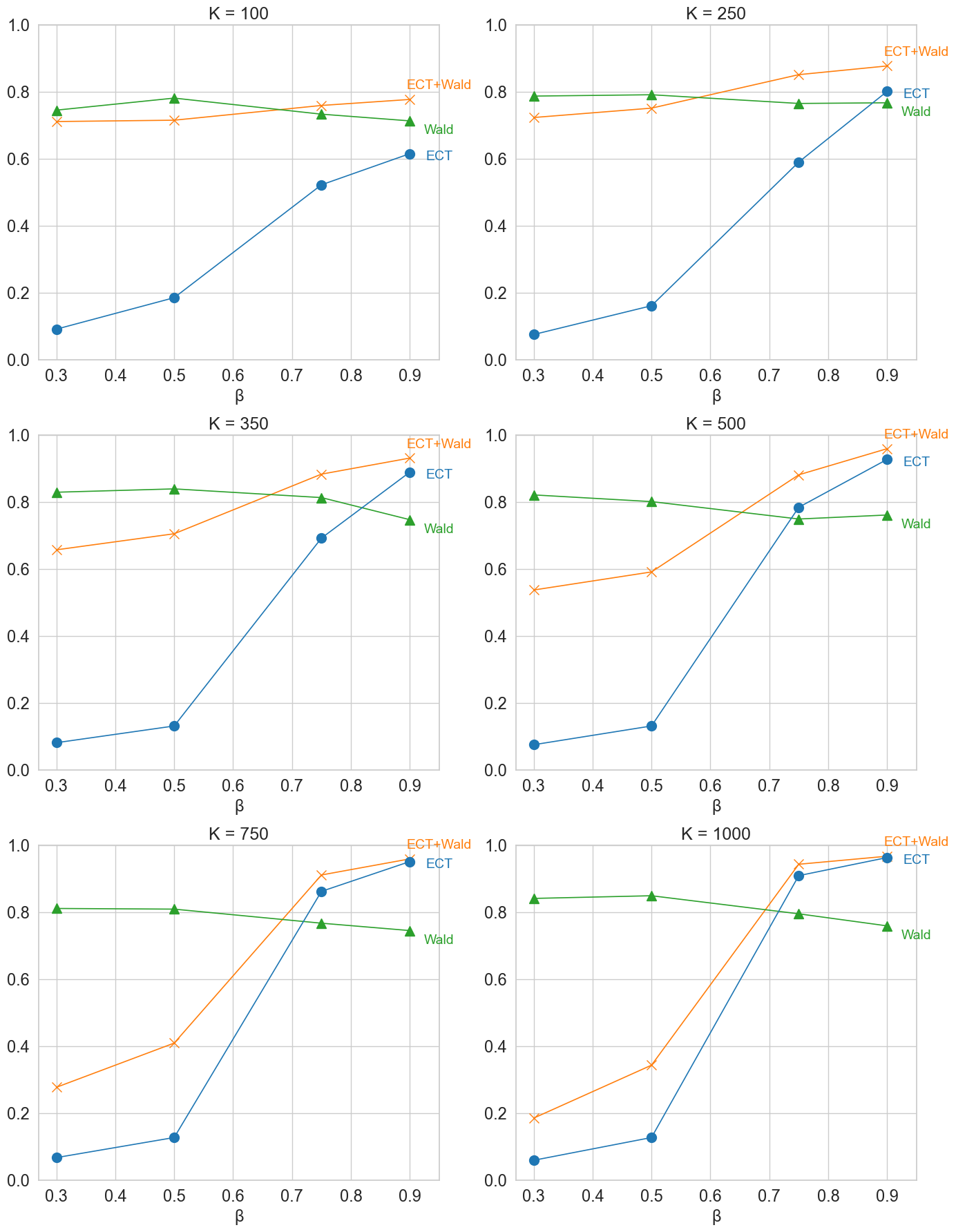}
\label{more_power_linear_results_Kgeq100}
\end{figure}

\begin{figure}
    \centering
     \caption{The power results in logistic regression with \(p = 3\) and \(n = 500\) for different methods across various values of \(K\) and the linetypes are aligned with Figure \ref{more_power_linear_results_Kgeq100}. }
     \includegraphics[width=1\linewidth]{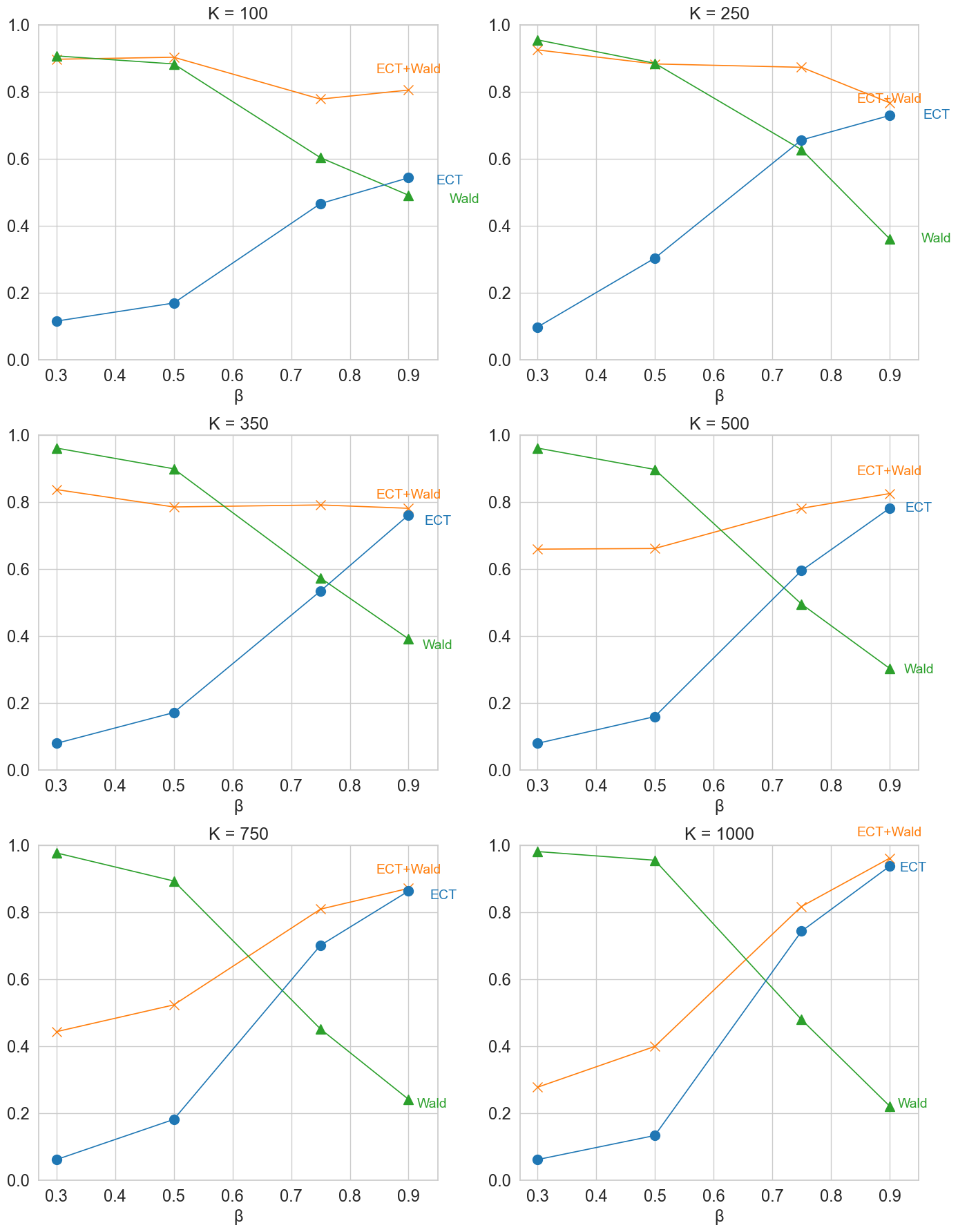}
\label{more_power_logis_results_Kgeq100}
\end{figure}

\section{Real Study Result Details}\label{appendix: real study details}
\setcounter{figure}{0}
\setcounter{table}{0}

In this section, we present the detailed meaning of each feature and the results of logistic regression for the two groups after sample splitting, broken down by site category, as shown in Tables \ref{table: logistic_results_coefficients} and \ref{table: logistic_results_std_errors}.

\begin{table}[]
\centering
\caption{Features, descriptions and encoding types in the Avazu click-through rate dataset.}
\label{table: ad_character}
\begin{tabular}{|l|p{9.5cm}|l|}
\hline
\textbf{Feature} & \textbf{Description} & \textbf{Encoding Type} \\
\hline
{id} & A unique identifier for each ad impression. & Masked  \\
\hline
{click} & A binary variable indicating whether the ad was clicked (1 if clicked, 0 otherwise). & Discrete  \\
\hline
{hour} & The timestamp of the ad impression in YYMMDDHH format (Year, Month, Day, Hour). & Numeric  \\
\hline
{C1} & An anonymized categorical feature (unknown meaning). & Discrete  \\
\hline
{banner\_pos} & The position of the ad on the webpage (e.g., top, side, etc.). & Discrete \\
\hline
{site\_domain} & The domain of the website where the ad was shown. & Masked  \\
\hline
{site\_category} & The category of the website (e.g., news, sports, etc.). & Masked  \\
\hline
{app\_id} & A unique identifier for the app where the ad was displayed (if shown in an app). & Masked  \\
\hline
{device\_model} & The model of the device used for the ad impression (e.g., iPhone 11, Samsung Galaxy, etc.). & Masked \\
\hline
{device\_conn\_type} & The type of network connection used (e.g., Wi-Fi, mobile data). & Discrete \\
\hline
{C14}-{C21} & Anonymized categorical features (unknown meanings). & Discrete  \\
\hline
\end{tabular}
\end{table}

\begin{table}
\centering
\fontsize{10pt}{10pt}\selectfont
\caption{Logistic regression coefficients of features across two groups for each site category.}
\begin{tabular}{|c|c|c|c|}
\hline
\textbf{Site Category} & \textbf{Feature} & \textbf{Group $1$ Coefficient} & \textbf{Group $2$ Coefficient} \\
\hline
28905ebd & {weekend} & 0.09666066 & 0.10102885 \\
28905ebd & {cos\_hour} & 0.07813501 & 0.07848706 \\
28905ebd & {device\_conn\_type} & 0.13353454 & 0.13286049 \\
28905ebd & {banner\_pos} & -1.28969920 & -1.31061934 \\
28905ebd & {C18} & 0.04971742 & 0.05183141 \\
28905ebd & {C19} & 0.03328290 & 0.03390778 \\
28905ebd & {C21} & -0.03407175 & -0.03341361 \\
\hline
3e814130 & {weekend} & 0.00019428 & 0.00311024 \\
3e814130 & {cos\_hour} & 0.05461391 & 0.06081830 \\
3e814130 & {device\_conn\_type} & 0.02820926 & 0.03999413 \\
3e814130 & {banner\_pos} & -0.87617363 & -0.87812688 \\
3e814130 & {C18} & -0.22638426 & -0.22025544 \\
3e814130 & {C19} & -0.38406075 & -0.37794136 \\
3e814130 & {C21} & -0.36260043 & -0.35870090 \\
\hline
f66779e6 & {weekend} & -0.08843932 & -0.18430991 \\
f66779e6 & {cos\_hour} & -0.02091556 & -0.05164434 \\
f66779e6 & {device\_conn\_type} & 0.21790652 & 0.13009402 \\
f66779e6 & {banner\_pos} & 0.45818045 & 0.43827614 \\
f66779e6 & {C18} & -0.51412530 & -0.53290644 \\
f66779e6 & {C19} & -0.42454448 & -0.40037968 \\
f66779e6 & {C21} & 0.02054264 & 0.01189972 \\
\hline
75fa27f6 & {weekend} & 0.05531754 & 0.03756613 \\
75fa27f6 & {cos\_hour} & 0.02232968 & 0.05218150 \\
75fa27f6 & {device\_conn\_type} & 0.16007865 & 0.11954274 \\
75fa27f6 & {banner\_pos} & -0.94038970 & -0.90383007 \\
75fa27f6 & {C18} & -0.30854908 & -0.32542333 \\
75fa27f6 & {C19} & -0.58967302 & -0.56381381 \\
75fa27f6 & {C21} & -0.20544393 & -0.24982930 \\
\hline
335d28a8 & {weekend} & -0.04507724 & -0.07016884 \\
335d28a8 & {cos\_hour} & 0.03221640 & 0.03264525 \\
335d28a8 & {device\_conn\_type} & 0.00005344 & 0.01589622 \\
335d28a8 & {banner\_pos} & 0.42997630 & 0.40720831 \\
335d28a8 & {C18} & -0.12464660 & -0.13848740 \\
335d28a8 & {C19} & 0.07514346 & 0.05453338 \\
335d28a8 & {C21} & -0.17655060 & -0.17227720 \\
\hline
76b2941d & {weekend} & -0.49368570 & -0.23481570 \\
76b2941d & {cos\_hour} & -0.30805620 & -0.29040840 \\
76b2941d & {device\_conn\_type} & 0.24159600 & 0.22902780 \\
76b2941d & {banner\_pos} & 1.36100920 & 1.00875070 \\
76b2941d & {C18} & -0.49312720 & -0.49706620 \\
76b2941d & {C19} & -0.31906440 & -0.42217870 \\
76b2941d & {C21} & -0.88178830 & -0.78746900 \\
\hline
c0dd3be3 & {weekend} & 0.16344740 & 0.08067979 \\
c0dd3be3 & {cos\_hour} & 0.05043350 & 0.10102225 \\
c0dd3be3 & {device\_conn\_type} & 0.11736960 & -0.06634701 \\
c0dd3be3 & {banner\_pos} & 0.87598390 & 0.95104335 \\
c0dd3be3 & {C18} & -0.09997430 & -0.05794371 \\
c0dd3be3 & {C19} & -0.16151230 & -0.09487071 \\
c0dd3be3 & {C21} & -0.20257900 & -0.17607583 \\
\hline
\end{tabular}
\label{table: logistic_results_coefficients}
\end{table}

\begin{table}
\centering
\fontsize{10pt}{10pt}\selectfont 
\caption{Group $1$ and group $2$'s standard errors for each site category.}
\begin{tabular}{|c|c|c|c|}
\hline
\textbf{Site Category} & \textbf{Feature} & \textbf{Group1 Std Error} & \textbf{Group2 Std Error} \\
\hline
28905ebd & {weekend} & 0.00322268 & 0.00322198 \\
28905ebd & {cos\_hour} & 0.00198461 & 0.00198566 \\
28905ebd & {device\_conn\_type} & 0.00257250 & 0.00257389 \\
28905ebd & {banner\_pos} & 0.01144794 & 0.01158948 \\
28905ebd & {C18} & 0.00168396 & 0.00168602 \\
28905ebd & {C19} & 0.00142932 & 0.00142916 \\
28905ebd & {C21} & 0.00147985 & 0.00148108 \\
\hline
3e814130 & {weekend} & 0.00449624 & 0.00449150 \\
3e814130 & {cos\_hour} & 0.00284695 & 0.00285155 \\
3e814130 & {device\_conn\_type} & 0.00521045 & 0.00518629 \\
3e814130 & {banner\_pos} & 0.00574542 & 0.00573991 \\
3e814130 & {C18} & 0.00274487 & 0.00274172 \\
3e814130 & {C19} & 0.00326183 & 0.00325474 \\
3e814130 & {C21} & 0.00309767 & 0.00309760 \\
\hline
f66779e6 & {weekend} & 0.05200229 & 0.05387511 \\
f66779e6 & {cos\_hour} & 0.02524815 & 0.02538091 \\
f66779e6 & {device\_conn\_type} & 0.03362108 & 0.03582427 \\
f66779e6 & {banner\_pos} & 0.07220437 & 0.07461655 \\
f66779e6 & {C18} & 0.01530066 & 0.01541318 \\
f66779e6 & {C19} & 0.02007389 & 0.01968903 \\
f66779e6 & {C21} & 0.01781726 & 0.01790785 \\
\hline
75fa27f6 & {weekend} & 0.03188831 & 0.03188470 \\
75fa27f6 & {cos\_hour} & 0.01794041 & 0.01796704 \\
75fa27f6 & {device\_conn\_type} & 0.02350179 & 0.02384143 \\
75fa27f6 & {banner\_pos} & 0.03513160 & 0.03528451 \\
75fa27f6 & {C18} & 0.01471221 & 0.01478651 \\
75fa27f6 & {C19} & 0.02272012 & 0.02212920 \\
75fa27f6 & {C21} & 0.01935124 & 0.01935578 \\
\hline
335d28a8 & {weekend} & 0.03622492 & 0.03680584 \\
335d28a8 & {cos\_hour} & 0.01888585 & 0.01899468 \\
335d28a8 & {device\_conn\_type} & 0.02185696 & 0.02202663 \\
335d28a8 & {banner\_pos} & 0.02932805 & 0.02954671 \\
335d28a8 & {C18} & 0.01146598 & 0.01155342 \\
335d28a8 & {C19} & 0.01364806 & 0.01385191 \\
335d28a8 & {C21} & 0.01647628 & 0.01657973 \\
\hline
76b2941d & {weekend} & 0.16809977 & 0.14875233 \\
76b2941d & {cos\_hour} & 0.04324290 & 0.04106791 \\
76b2941d & {device\_conn\_type} & 0.03445967 & 0.03431482 \\
76b2941d & {banner\_pos} & 0.21553216 & 0.21967395 \\
76b2941d & {C18} & 0.03649958 & 0.03520861 \\
76b2941d & {C19} & 0.04436360 & 0.04482910 \\
76b2941d & {C21} & 0.05034596 & 0.04846002 \\
\hline
c0dd3be3 & {weekend} & 0.04903794 & 0.05018129 \\
c0dd3be3 & {cos\_hour} & 0.03894017 & 0.04029747 \\
c0dd3be3 & {device\_conn\_type} & 0.07355204 & 0.08565594 \\
c0dd3be3 & {banner\_pos} & 0.04562823 & 0.04712095 \\
c0dd3be3 & {C18} & 0.03622369 & 0.03709145 \\
c0dd3be3 & {C19} & 0.04102937 & 0.04161196 \\
c0dd3be3 & {C21} & 0.03830035 & 0.03844438 \\
\hline
\end{tabular}
\label{table: logistic_results_std_errors}
\end{table}

\section{Proofs for Main Theorem}\label{appendix: main_theorem}
\setcounter{figure}{0}
\setcounter{table}{0}
 
Firstly, follows the expansion in \cite{jmlr2013}, we present the expansion for $\widehat{\theta}_k-\theta_k^*$: 
$$
\widehat{\theta}_k-\theta_k^* =-\mathbb{E} \nabla_{\theta_k}^2 M_k(X_k, \theta_k^*)^{-1} \sum\limits_{i=n_k(1-\gamma)}^{n_k} \nabla_{\theta_k} M_k(X_{k,i}, \theta_k^*) + \mathbb{E} \nabla_{\theta_k}^2 M_k(X_k, \theta_k^*)^{-1} (P + Q)(\widehat{\theta}_k-\theta_k^*)
$$
Here, $
P = \mathbb{E} \nabla_{\theta_k}^2 M_k(X_k, \theta_k^*) -\sum\limits_{i=n_k(1-\gamma)}^{n_k} \nabla_{\theta_k} M_k(X_{k,i}, \theta_k^*), \quad \text{and} \quad Q = \sum\limits_{i=n_k(1-\gamma)}^{n_k} \nabla_{\theta_k} M_k(X_{k,i}, \theta_k^*)-\sum\limits_{i=n_k(1-\gamma)}^{n_k} \nabla^2_{\theta_k} M_k(X_{k,i}, \widehat{\theta}_k).
$

For $\epsilon>0$, denote  $\delta_\rho^\epsilon=\min\{\rho,\frac{\rho^-}{8G},\frac{\rho^-}{4},\epsilon\}$ and define good events as follows :\begin{itemize}
    \item $E_{0,k}:=\left\{\dfrac{1}{n_k} \sum\limits_{i=1}^{n_k}G(x_{k,i})\le 2G\right\}$;
    \item $E^\epsilon_{1,k}:=\left\{\left\|\dfrac{1}{n_k} \sum\limits_{i=1}^{n_k}\nabla_{\theta_k}^2M_k(X_{k,i},\theta^*_k)-\mathbf{E}\nabla_{\theta_k}^2M_k(X_k,\theta^*_k)\right\|_2\le \dfrac{\rho_-}{4}\right\}$;
    \item $E_{2,k}^\epsilon:=\left\{\left\|\dfrac1{n}\sum^n_{i=1}\nabla_{\theta_k} M_k(X_{k,i},\theta_k^*)\right\|_2\le\rho^-\delta_\rho^\epsilon/4\right\}$.
\end{itemize}
By regularity assumptions and Markov inequality and Appendix \ref{appendix: lemma proofs}, we can easily obtain that:  \[\mathrm{P}(E_{0,k})\geq 1-c_3 n_{\min}^{-v}, \mathrm{P}(E_{1,k}^\epsilon)\geq 1-c_{3,\epsilon}\cdot n_{\min}^{-v},\mathrm{P}(E^\epsilon_{2,k})=1-c_{3,\epsilon}n_{\min}^{-v}.\]

In \cite{jmlr2013}, Lemma 6, they establish local strong convexity and bound for $\|\widehat{\theta}_k-\theta_k^*\|$ around the true parameter $\theta_k^*$ under the good events. The following is a refined proof for the Lemma 6 of \cite{jmlr2013} that we have:
\begin{lemma}\label{lemma:good_event}
    Under events $E_{0,k}\cap E^\epsilon_{1,k}\cap E^\epsilon_{2,k}$, we have\[\|\widehat{\theta}_k-\theta^*_k\|_2\le\delta_\rho^\epsilon\le{\epsilon}, \quad \dfrac{1}{n_k} \sum\limits_{i=1}^{n_k}\nabla_{\theta_k}^2M_k(X_{k,i},\widehat{\theta}_k) \succeq \frac{\rho_-}{2}I_{p}.\]
\end{lemma}
\begin{proof}
 Firstly, in the ball $\{\theta:\|\theta-\theta^*\|\leq\delta_{\rho}\}$, under good events $E_{0,k}$ and $E^\epsilon_{1,k}$, we have $$\left\|\dfrac{1}{n_k} \sum\limits_{i=1}^{n_k}\nabla_{\theta_k}^2M_k(X_{k,i},\theta)-\mathbf{E}\nabla_{\theta_k}^2M_k(X_k,\theta^*_k)\right\|_2\leq 2G\|\theta-\theta^*_k\|+\delta_{\rho}^\epsilon<\frac{\rho^-+\rho^-}{4}=\frac{\rho^-}{2},$$
thus we have the local strong-convexity: $\dfrac{1}{n_k} \sum\limits_{i=1}^{n_k}\nabla_{\theta_k}^2M_k(X_{k,i},\theta)\succeq \frac{\rho^-}{2}I$ for $\theta$ in the ball $\{\theta:\|\theta-\theta^*\|\leq\delta_{\rho}\}.$

    If $\|\widehat{\theta}_k-\theta^*_k\|_2>\delta_{\rho}^\epsilon,$ we prove there's a contradiction with $E^\epsilon_{2,k}$. Denote $t=\dfrac{\delta_{\rho}^\epsilon}{\|\widehat{\theta}_k-\theta^*_k\|_2}\in(0,1)$,  due to global convexity, we have $$\dfrac{t}{n}\sum^n_{i=1}M_k(X_{k,i},\widehat{\theta}_k)+\dfrac{1-t}{n}\sum^n_{i=1}M_k(X_{k,i},{\theta}^*)\geq\dfrac{1}{n}\sum^n_{i=1}M_k(X_{k,i},(1-t){\theta}^*+t\widehat{\theta}_k).$$
    Because $(1-t){\theta}^*+t\widehat{\theta}_k $ lies in the ball $\{\theta:\|\theta-\theta^*\|\leq\delta_{\rho}\},$ then we can use local strong convexity to deal with the RHS of the inequality above and  obtain:
    $$\dfrac{1}{n}\sum^n_{i=1}M_k(X_{k,i},(1-t){\theta}^*+t\widehat{\theta}_k)\geq\dfrac{1}{n}\sum^n_{i=1}M_k(X_{k,i},{\theta}^*)-\|\dfrac{1}{n}\sum^n_{i=1}\nabla_\theta M_k(X_{k,i},{\theta}^*)\|\cdot t\|\theta^*-\widehat{\theta}_k\|+\frac{1}{2}\frac{\rho^-}{2}t^2\|\theta^*-\widehat{\theta}_k\|_2^2$$
    Combining the two inequalities above and  $\dfrac{1}{n}\sum^n_{i=1}M_k(X_{k,i},{\theta}^*)$ appears  on both side, we have $$\dfrac{t}{n}\left(\sum^n_{i=1}M_k(X_{k,i},\widehat{\theta}_k)-\sum^n_{i=1}M_k(X_{k,i},{\theta}^*)\right)\geq-\|\dfrac{t}{n}\sum^n_{i=1}\nabla_\theta M_k(X_{k,i},{\theta}^*)\|\|\theta^*-\widehat{\theta}_k\|+\frac{1}{2}\frac{\rho^-}{2}(\delta_\rho^\epsilon)^2$$
    Due to $\hat\theta_k$ is the minimizer of the empirical loss, thus
    $$\|\dfrac{t}{n}\sum^n_{i=1}\nabla_\theta M_k(X_{k,i},{\theta}^*)\|\|\theta^*-\widehat{\theta}_k\|>\frac{1}{2}\frac{\rho^-}{2}(\delta_\rho^\epsilon)^2.$$
    Using the definition of $\delta_\rho^\epsilon$ and $t$, we have  $\|\dfrac{1}{n}\sum^n_{i=1}\nabla_\theta M_k(X_{k,i},{\theta}^*)\|>\dfrac{\rho^-}{4}\delta_\rho^\epsilon$, which is a contradiction with the good event $E_{2,k}^\epsilon.$\end{proof}
Moreover, we define more good events.
\begin{itemize}
    \item $E_{3,k}:=\left\{\dfrac{1}{n_k} \sum\limits_{i=1}^{n_k}B(X_{k,i})\le 2B\right\}$;
    \item $E_{4,k}^\epsilon:=\left\{\left\|\dfrac{1}{n_k} \sum\limits_{i=1}^{n_k}Z_k(X_{k,i},\theta^*_k)-\mathbf{E}(Z_k(X_k,\theta_k^*))\right\|_2\le\epsilon\right\}$.
\end{itemize}
Similarly we can easily obtain that for a fix $\epsilon>0$, $\mathrm{P}(E_{3,k})=1-O\left(n_{\min}^{-v}\right),\mathrm{P}(E^\epsilon_{4,k})=1-O\left(n_{\min}^{-v}\right)$ under Assumptions \ref{assumption:identifiability_compactness}-\ref{assumption: Local Strong Convexity}. Moreover, under Assumption \ref{assumption:identifiability_compactness}-\ref{assumption: Local Strong Convexity}, we have  $0 < \sigma^- \leq\sigma_k^{(j)}\leq \sigma^+ < \infty$ for all $k,j$.

\begin{lemma}\label{theorem: variance_estimates_thereom1}
    If we estimate the variance of each estimator
     using \ref{X^-1YX}, under the assumptions \ref{assumption:identifiability_compactness}-\ref{assumption: Local Strong Convexity},  then we have $\underset{1 \leq j \leq p,1\leq k \leq K}{\max} |\widehat{\sigma}_k^{(j)}-{\sigma}_k^{(j)}|=o_p(1)$ as $n_{\min} \to \infty$  for $K=o(n_{\min}^v)$.
\end{lemma}

\subsection{Proof for Lemma \ref{theorem: variance_estimates_thereom1}}\label{appendix: theorem1_variance_proof}
\begin{proof}
We only need to prove $\mathrm{P}(\max_{1\leq j \leq p}|(\sigma_k^{(j)})^2-(\widehat{\sigma}_k^{(j)})^2|>\epsilon)=\mathcal{O}(n_{\min}^{-v})$ for each $k$ and $\epsilon>0$, then $\mathrm{P}(\max_{1\leq j \leq p,1\leq k \leq K}|(\sigma_k^{(j)})^2-(\widehat{\sigma}_k^{(j)})^2|>\epsilon)=\mathcal{O}(\dfrac{K}{n_{\min}^v})=o(1)$, which indicates $ \max_{1\leq j \leq p,1\leq k \leq K}|(\sigma_k^{(j)})^2-(\widehat{\sigma}_k^{(j)})^2|\xrightarrow[]{P}0$ for $K=o(n_{\min}^v)$. 

Thus we only need to prove\begin{align}\label{cov_t1}
&\mathrm{P}\Bigg(\Bigg\|\Bigg(\dfrac{1}{n}\sum\limits_{i=1}^n \nabla_{\theta_k}^2 M_k(X_{k,i},\widehat{\theta}_k)\Bigg)^{-1} 
\Bigg(\dfrac{1}{n}\sum\limits_{i=1}^n \nabla_{\theta_k} M_k(X_{k,i},\widehat{\theta}_k) \nabla_{\theta_k} M_k(X_{k,i},\widehat{\theta}_k)^\top \Bigg) \\
&\qquad \Bigg(\dfrac{1}{n}\sum\limits_{i=1}^n \nabla_{\theta_k}^2 M_k(X_{k,i},\widehat{\theta}_k)\Bigg)^{-1} 
- \mathbf{V}_k (\theta_k^*) \Bigg\|_2 >\epsilon \Bigg)=\mathcal{O}(n_{\min}^{-v})
\end{align}

    For data block $k$, under good events, we have\begin{align*}
        &\quad\left\|\dfrac{1}{n_k} \sum\limits_{i=1}^{n_k}\nabla_{\theta_k}^2M_k(X_{k,i},\widehat{\theta}_k)-\mathbf{E}\nabla_{\theta_k}^2M_k(X_k,\theta^*_k)\right\|_2
        \\&\le\dfrac{1}{n_k} \sum\limits_{i=1}^{n_k}G(X_{k,i})\cdot\left\|\widehat{\theta}_k-\theta^*_k\right\|+\left\|\dfrac{1}{n_k} \sum\limits_{i=1}^{n_k}\nabla_{\theta_k}^2M_k(X_{k,i},\theta^*_k)-\mathbf{E}\nabla^2_{\theta_k}M_k(X_k,\theta^*_k)\right\|
        \\&\le{\rho^-\cdot 2G}{\epsilon}+\dfrac{\rho\rho^-}{2}\epsilon 
    \end{align*}

    We choose $2\epsilon<({\rho^-\cdot 2G}+\dfrac{\rho\rho^-}{2})^{-1} $ Thus \[\displaystyle\dfrac{1}{n_k} \sum\limits_{i=1}^{n_k}\nabla_{\theta_k}^2M_k(X_{k,i},\widehat{\theta}_k)\succeq\left(1-{\rho^-\cdot 2G}{\epsilon}-\dfrac{\rho\rho^-}{2}\epsilon\right)\rho^-I_{\mathrm{P}\times P}\succeq \dfrac{\rho_-}{2}I_{\mathrm{P}\times P} \] and \[\displaystyle\left\|\left[\dfrac{1}{n_k} \sum\limits_{i=1}^{n_k}\nabla^2_{\theta_k}M_k(X_{k,i},\widehat{\theta}_k)\right]^{-1}\right\|_2\le\dfrac1{\rho^-(1-(2G+\rho/2)\cdot\epsilon)}.  \]
Similarly we can obtain Under events $E_{0,k}\cap E^\epsilon_{1,k}\cap E^\epsilon_{2,k}\cap E_{3,k}\cap E_{4,k}^\epsilon$,\[\left\|\dfrac{1}{n_k} \sum\limits_{i=1}^{n_k}Z_k(X_{k,i},\widehat{\theta}_k)-\mathbf{E}Z_k(X_k,\theta^*_k)\right\|_2\le (2B\rho_-+2)\epsilon\]  
    For non-singular matrix $X,Y$, $\hat{X},\hat{Y}$, we  have\begin{align*}
        &\quad\|\hat{X}^{-1}\hat{Y}(\hat{X}^\top)^{-1}-X^{-1}Y{X}^{-1}\|_2
        \\&\le\|\hat{X}^{-1}-X^{-1}\|_2\cdot\|\hat{Y}(\hat{X}^\top)^{-1}\|_2+\|X^{-1}\hat{Y}(\hat{X}^{-1}-{X}^{-1})\|_2+\|X^{-1}(\hat{Y}-Y){X}^{-1}\|_2
    \end{align*}
and\[\|\hat{X}^{-1}-X^{-1}\|_2=\|\hat{X}^{-1}(\hat{X}-X)X^{-1}\|_2\le \|\hat{X}-X\|_2\cdot\|\hat{X}^{-1}\|_2\|X^{-1}\|_2.\]
Thus
\begin{align*}
        &\quad\|\hat{X}^{-1}\hat{Y}(\hat{X}^\top)^{-1}-X^{-1}Y{X}^{-1}\|_2
        \\&\le \|\hat{X}^{-1}\|_2^2 \cdot \|X^{-1}\|_2 \cdot \|\hat{X}-X\|_2\cdot \|\hat{Y}\|_2+\|\hat{X}^{-1}\|_2 \cdot \|X^{-1}\|_2^2 \cdot \|\hat{X}-X\|_2\cdot \|\hat{Y}\|_2\\
        &+\|X^{-1}\|_2^2 \cdot \|\hat{Y}-Y\|_2
    \end{align*}
Applying the expression to estimate \( \mathbf{V}_k (\theta_k^*) \) in  \eqref{cov_t1}, we substitute \( \hat{X} = \frac{1}{n} \sum\limits_{i=1}^n \nabla_{\theta_k}^2 M_k(X_{k,i}, \widehat{\theta}_k) \) into the equation. Additionally, we substitute 
$\hat{Y} = \frac{1}{n} \sum\limits_{i=1}^n \nabla_{\theta_k} M_k(X_{k,i}, \widehat{\theta}_k) \nabla_{\theta_k} M_k(X_{k,i}, \widehat{\theta}_k)^\top$
into the corresponding term. Given these substitutions, it becomes straightforward to verify that the desired result holds:
\[
\| \widehat{\sigma}_k(\widehat{\theta}_k)-\Sigma_k(\theta_k) \|_2 \leq C(B,G,\rho_{\sigma},\rho,\rho_-,\sigma_{\max},\sigma_{\min})\epsilon
\]
with probability 
$O\left(\frac{1}{n_{\min}^v}\right),$
where $C(B,G,\rho_{\sigma},\rho,\rho_-,\sigma_{\max},\sigma_{\min})$ is a constant dependent solely on the terms within the parentheses.

    Also recall that for a symmetric semi-positive matrix, the maximum of its digonal element is no more than the matrix's 2-norm, then the proof is done. \end{proof}

In the remaining proof, we denote the following. Since the expression for the difference between the estimator \( \widehat{\theta}_k \) and the true parameter \( \theta_k^* \) can be written as  \citep{jmlr2013,jiagu2023}:

\[
\widehat{\theta}_k-\theta_k^* = \frac{1}{n_k} [\mathbb{E} \nabla_{\theta_k}^2 M_k(X_k, \theta_k^*)]^{-1} \sum\limits_{i=1}^{n_k } \nabla_{\theta_k} M_k(X_{k,i}, \theta_k^*) + R_k =\dfrac{\sum\limits_{i=1}^{n_k}f_k(X_{k,i}, \theta_k^*)}{n_k}
\]
Here, \( f_k(X_{k,i}, \theta_k^*) = \mathbb{E} \nabla_{\theta_k}^2 M_k(X_k, \theta_k^*)^{-1} \nabla_{\theta_k} M_k(X_{k,i}, \theta_k^*) \) and the set \( \{f_k(X_{k,i}, \theta_k^*)\}_{i=1,2,\dots,n} \) are i.i.d. \( p \)-dimensional random variables. \[R_k=-(\mathbb{E}\nabla^2_{\theta_k}M_k(X_{k,i},\theta_k^*))^{-1}\cdot\left(\mathbb{E}\nabla^2_{\theta_k}M_k(X_{k,i},\theta_k^*)-\frac{1}{n_k}\sum_{i = 1}^{n_k}\nabla_{\theta_k}M_k(X_{k,i},\widehat{\theta}_k)\right)(\widehat{\theta}_k-\theta_k^*)\] is the remainder term. We also denote:
\[P_k=-(\mathbb{E}\nabla^2_{\theta_k}M_k(X_{k,i},\theta_k^*))^{-1}\cdot\left(\mathbb{E}\nabla^2_{\theta_k}M_k(X_{k,i},\theta_k^*)-\frac{1}{n_k}\sum_{i = 1}^{n_k}\nabla^2_{\theta_k}M_k(X_{k,i},{\theta}^*_k)\right),\]
\[Q_k=-(\mathbb{E}\nabla^2_{\theta_k}M_k(X_{k,i},\theta_k^*))^{-1}\cdot\left(\frac{1}{n_k}\sum_{i = 1}^{n_k}\nabla_{\theta_k}M_k(X_{k,i},{\theta}^*_k)-\frac{1}{n_k}\sum_{i = 1}^{n_k}\nabla^2_{\theta_k}M_k(X_{k,i},\hat{\theta}_k)\right).\]

\subsection{Proof for Theorem \ref{Wald consistency}}\label{appendix: Wald consistency proof}
\begin{proof}
We're going to prove the asymptotic normality of the  Wald-type test under the null. 

Using Lemma \ref{lemma:Projection_Matrix_expansion}, for the Wald-type test we have: $(R\widehat{\theta_{j}})^{\top}(R\widehat{V_j}R^{\top})^{-1}R\widehat{\theta_{j}}$
\begin{equation}\label{equation: expansion_Wald_when_not_identity}
     = \sum\limits_{1 \leq k_1 < k_2 \leq K}(\widehat{\theta}_{k_1}^{(j)}-\widehat{\theta}_{k_2}^{(j)})^{2}\cdot \hat{r}^{(j)}_{k_1,k_2},
    \quad \hat{r}^{(j)}_{k_1,k_2} = \dfrac{\prod\limits_{k\ne k_1,k_2}n_{k}^{-1}(\widehat{\sigma}_{k}^{(j)})^{2} }{\sum\limits_{1 \leq l \leq K}\prod\limits_{k\ne l}n_{k}^{-1}(\widehat{\sigma}_{k}^{(j)})^{2}}, 
    \quad r_{k_1,k_2}^{(j)} = \dfrac{\prod\limits_{k\ne k_1,k_2}n_k^{-1}({\sigma}_{k}^{(j)})^{2} }{\sum\limits_{1 \leq l \leq K}\prod\limits_{k\ne l}n_{k}^{-1}({\sigma}_{k}^{(j)})^{2}}.
\end{equation}
Recall that $\widehat{\theta}_{k}^{(j)}-\theta_k^{(j),*}=\dfrac{1}{n_k}\sum\limits_{i=1}^{n_k}f_k^{(j)} (X_{k,i},\theta_k^{*})+R_k^{(j)}.$ Under null hypothesis, $\theta_{k_1}^{(j),*}=\theta_{k_2}^{(j),*}$, we will demonstrate which terms in Wald-type statistic contribute to the asymptotic normality and which terms asymptotically converge to zero.

$$\begin{aligned}\label{expansion_thetak1_thetak2}
    (\widehat{\theta}_{k_1}^{(j)}-\widehat{\theta}_{k_2}^{(j)})^2 =& \underbrace{\dfrac{1}{n_{k_1}^2}\left(\sum\limits_{i=1}^{n_{k_1}}f_{k_1}^{(j)} (X_{k_1,i},\theta_{k_1}^{*})\right)^{2} + \dfrac{1}{n_{k_2}^2}\left(\sum\limits_{i=1}^{n_{k_2}}f_{k_2}^{(j)} (X_{k_2,i},\theta_{k_2}^{*})\right)^{2}}_{A_{k_1,k_2}^{(j)}} \\
    -& \underbrace{\dfrac{1/\gamma}{n_{k_1}n_{k_2}}\left(\sum\limits_{i=1}^{n_{k_1}}f_{k_1}^{(j)} (X_{k_1,i},\theta_{k_1}^{*})\right)\left(\sum\limits_{i=1}^{n_{k_2}}f_{k_2}^{(j)} (X_{k_2,i},\theta_{k_2}^{*})\right)}_{B_{k_1,k_2}^{(j)}} \\ 
    +& \underbrace{2\left(\dfrac{1}{n_{k_1}}\sum\limits_{i=1}^{n_{k_1}}f_{k_1}^{(j)} (X_{k_1,i},\theta_{k_1}^{*})-\dfrac{1}{n_{k_2}}\sum\limits_{i=1}^{n_{k_2}}f_{k_2}^{(j)} (X_{k_2,i},\theta_{k_2}^{*})\right)(R_{k_1}^{(j)}-R_{k_2}^{(j)})}_{C_{k_1,k_2}^{(j)}} \\
    +& \underbrace{(R_{k_1}^{(j)}-R_{k_2}^{(j)})^2}_{D_{k_1,k_2}^{(j)}}
\end{aligned}$$
We show the asymptotic normality for $\dfrac{1}{\sqrt{2K-2}}\left(\sum\limits_{1 \leq k_1 < k_2 \leq K}(\widehat{\theta}_{j}^{k_1}-\widehat{\theta}_{j}^{k_2})^{2}\cdot \hat{r}^{(j)}_{k_1,k_2}-(K-1)\right)$. Specifically, from Lemma \ref{lemma: A_k_1,k_2}, \ref{lemma: BCD_k1k2}, we have proven that 
\[
\dfrac{1}{\sqrt{2K-2}}\left(\sum\limits_{1 \leq k_1 < k_2 \leq K} A_{k_1k_2}\widehat{r}_{k_1k_2}^{(j)}-(K-1)\right) \xrightarrow[]{d} N(0,1),
\]
and that the remaining terms related to $B_{k_1,k_2}^{(j)},C_{k_1,k_2}^{(j)},D_{k_1,k_2}^{(j)}$ in the test statistic asymptotically converge to zero as $n_{\text{min}} \to \infty$, provided that $K = o(n_{\text{min}})$. Thus, the asymptotically standard normality of the Wald-type test under the null distribution is established. \end{proof}

\subsection{Proof for Theorem \ref{theorem: ECT_asym_normal}}\label{appendix: proof for ECT consistent}
\begin{proof}

We start by defining the test statistic $t_{k_1 k_2}^{(j),[2]}$ as:

\[
t_{k_1 k_2}^{(j),[2]} = \frac{\widehat{\theta}_{k_1}^{(j),[2]}-\widehat{\theta}_{k_2}^{(j),[2]}}{\sqrt{\frac{1/\gamma}{n_{k_1}} \left( \widehat{\sigma}_{k_1}^{(j)} \right)^2 + \frac{1/\gamma}{n_{k_2}} \left( \widehat{\sigma}_{k_2}^{(j)} \right)^2}}.
\]

Expanding $t_{k_1 k_2}^{(j),[2]}$:

\[
t_{k_1 k_2}^{(j),[2]} = \frac{\frac{1}{n_{k_1}} \sum\limits_{i=n_{k_1}*(1-\gamma)}^{n_{k_1}} f^{(j)}(X_{k_1,i}, \theta_{k_1}^*)-\frac{1}{n_{k_2}} \sum\limits_{i=n_{k_2}*(1-\gamma)}^{n_{k_2}} f^{(j)}(X_{k_2,i}, \theta_{k_2}^*) + R_{k_1}^{(j)}-R_{k_2}^{(j)}+\theta_{k_1}^*-\theta_{k_2}^*}{\sqrt{\frac{1/\gamma}{n_{k_1}} \left( \widehat{\sigma}_{k_1}^{(j)} \right)^2 + \frac{1/\gamma}{n_{k_2}} \left( \widehat{\sigma}_{k_2}^{(j)} \right)^2}}.
\]
We also denote in the Theorem \ref{theorem: ECT_asym_normal} that:
\[
\widetilde{t}_{k_1 k_2}^{(j),[2]} = \frac{\frac{1}{n_{k_1}} \sum\limits_{i=n_{k_1}*(1-\gamma)}^{n_{k_1}} f^{(j)}(X_{k_1,i}, \theta_{k_1}^*)-\frac{1}{n_{k_2}} \sum\limits_{i=n_{k_2}*(1-\gamma)}^{n_{k_2}} f^{(j)}(X_{k_2,i}, \theta_{k_2}^*) + R_{k_1}^{(j)}-R_{k_2}^{(j)}}{\sqrt{\frac{1/\gamma}{n_{k_1}} \left( \widehat{\sigma}_{k_1}^{(j)} \right)^2 + \frac{1/\gamma}{n_{k_2}} \left( \widehat{\sigma}_{k_2}^{(j)} \right)^2}}.
\]
Next, we define a t-statistic, denoted as $\widetilde{\tilde{t}}_{k_1 k_2}^{(j),[2]}$, which eliminates the bias and divides by the true variance, rather than an estimated variance, thus ensuring that the variance equals 1 and expectation equals to 0.

\[
\widetilde{\tilde{t}}_{k_1 k_2}^{(j),[2]} = \frac{\frac{1}{\gamma \cdot n_{k_1}} \sum\limits_{i=n_{k_1}*(1-\gamma)}^{n_{k_1}} f^{(j)}(X_{k_1,i}, \theta_{k_1}^*)-\frac{1}{\gamma\cdot n_{k_2}} \sum\limits_{i=n_{k_2}}^{n_{k_2}*(1-\gamma)} f^{(j)}(X_{k_2,i}, \theta_{k_2}^*)}{\sqrt{\frac{1/\gamma}{n_{k_1}} \left( \sigma_{k_1}^{(j)} \right)^2 + \frac{1/\gamma}{n_{k_2}} \left( \sigma_{k_2}^{(j)} \right)^2}}.
\]

It is easy to observe that $\widetilde{\tilde{t}}_{k_1 k_2}^{(j),[2]}$ has variance 1 and mean 0, which simplifies the proof of asymptotic normality.

We divide the entire proof into four steps. The first step is to prove the asymptotic normality of the refined t-statistic has been substituted:
\[
\sup_x \left| \mathrm{P}\left( \widetilde{\tilde{t}}_{k_1 k_2}^{(j)} \leq x \right)-\Phi(x) \right| \lesssim O\left( n_{\min}^{-\frac{1}{2}} \right).
\]
Where $\mathbb{I}_{k_1, k_2}^{(j)}=\mathbb{I}\{k_1=\arg\max_k \widehat{\theta}_k^{(j),[1]},k_2=\arg\min_k \widehat{\theta}_k^{(j),[1]}\}$. The second step is two prove:
\[
\sup_x \left| \mathrm{P}\left( \sum\limits_{k_1, k_2} \mathbb{I}_{k_1, k_2} ^{(j)}\widetilde{\tilde{t}}_{k_1 k_2}^{(j)} \leq x \right)-\Phi(x) \right| \lesssim O\left( n_{\min}^{-\frac{1}{2}} \right).
\]
The third step is to prove:
\[
\sum\limits_{k_1, k_2} \mathbb{I}_{k_1, k_2} ^{(j)}(\widetilde{\tilde{t}}_{k_1 k_2}^{(j)}-\widetilde{{t}}_{k_1 k_2}^{(j)}) \xrightarrow{P} 0.
\]
And the fourth is to prove:\[
\sup_x \left| \mathrm{P}\left( \sum\limits_{k_1, k_2}\mathbb{I}_{k_1, k_2}^{(j)}{t}_{k_1 k_2}^{(j)} \leq x \right)-\Phi(x) \right| \lesssim O\left( n_{\min}^{-\frac{v}{2(v+2)}} \right).
\]

For the first step, this requires proving the asymptotic normality of \(\sum\limits_{k_1, k_2} \mathbb{I}_{k_1, k_2} \widetilde{\tilde{t}}_{k_1 k_2}^{(j)}\). Since \(\{\widetilde{\tilde{t}}_{k_1 k_2}^{(j)}\}_{1 \leq i \leq n}\) are i.i.d. random variables with zero mean and unit variance, we apply the Berry-Esseen theorem.

Firstly, we use the Berry-Esseen theorem to bound the convergence rate of each t-statistic with index $k_1, k_2$. We can rewrite $\widetilde{\tilde{t}}_{k_1 k_2}^{(j)}$ as

\[
\widetilde{\tilde{t}}_{k_1 k_2}^{(j)} = \frac{n_{k_2}\sum\limits_{i=n_{k_1}*(1-\gamma)}^{n_{k_1}} f^{(j)}(X_{k_1,i}, \theta_{k_1}^*)-n_{k_1}\sum\limits_{i=n_{k_2}*(1-\gamma)}^{n_{k_2}} f^{(j)}(X_{k_2,i}, \theta_{k_2}^*)}{\sqrt{\lceil\gamma n_{k_1}\rceil\cdot n_{k_2}^2({\sigma}_{k_1}^{(j)})^2 + \lceil\gamma n_{k_2}\rceil n_{k_1}^2({\sigma}_{k_2}^{(j)})^2}}.
\]

Observing that the equation above, the denominator is the sum of the variances of each independent term in the numerator and the expected value of each term in the numerator is zero, by Berry-Essen,  we have $\sup_{x \in \mathbb{R}} \left| \mathrm{P}\left( \widetilde{\tilde{t}}_{k_1 k_2}^{(j)} \leq x \right)-\Phi(x) \right| $
\begin{align*}\label{BERRYESSEN}
&\leq C_1 \frac{\sum\limits_{i=n_{k_1}*(1-\gamma)}^{n_{k_1}}\mathbb{E}|n_{k_2}f^{(j)}(X_{k_1,i}, \theta_{k_1}^*)|^3 
+ \sum\limits_{i=n_{k_2}*(1-\gamma)}^{n_{k_2}}\mathbb{E}|n_{k_1}f^{(j)}(X_{k_2,i}, \theta_{k_2}^*)|^3}{\big(\lceil\gamma n_{k_1}\rceil n_{k_2}^2({\sigma}_{k_1}^{(j)})^2 
+ \lceil\gamma n_{k_2}\rceil n_{k_1}^2({\sigma}_{k_2}^{(j)})^2\big)^{\frac{3}{2}}} \\
&\leq C_1 \frac{R^3}{\sqrt{\gamma}\sigma_{\min}^3} 
\frac{\gamma (n_{k_1}n_{k_2}^3+n_{k_1}^3n_{k_2})}{\big(\lceil\gamma n_{k_1}\rceil n_{k_2}^2 + \lceil\gamma n_{k_2}\rceil n_{k_1}^2\big)^{\frac{3}{2}}} \\
&\leq C_1 \frac{R^3}{\sqrt{\gamma}\sigma_{\min}^3} 
\frac{n_{k_1}^2 + n_{k_2}^2}{\sqrt{n_{k_1}n_{k_2}} (n_{k_1} + n_{k_2})^{1.5}}\leq C_1 \frac{R^3}{\sqrt{\gamma}\sigma_{\min}^3} 
\frac{n_{k_1}^2 + n_{k_2}^2}{\sqrt{n_{k_1}n_{k_2}} (n_{k_1}^{1.5} + n_{k_2}^{1.5})} \\
&= C_1 \frac{R^3}{\sqrt{\gamma}\sigma_{\min}^3} 
\frac{n_{k_1}^2 + n_{k_2}^2}{n_{k_1}^2\sqrt{n_{k_2}} + n_{k_2}^2\sqrt{n_{k_1}}}\leq\frac{\rho_0}{\sqrt{\min\{n_{k_1}, n_{k_2}\}} }\leq \frac{\rho_0}{\sqrt{n_{\min}} }.
\end{align*}

where $C_1$ is the constant derived from the Berry-Esseen bound and $\rho_0$ is a constant composed of $\sigma_{\min}$, $R$, $C_1$ and $\gamma$. Note that $\gamma$ is a predetermined constant used for data splitting.

In the second step, we bound the convergence of the overall statistic.

\[
|\mathrm{P}( \sum\limits_{k_1, k_2} \mathbb{I}_{k_1 k_2}  \widetilde{t}_{k_1 k_2}^{(j)} \leq x) -\Phi(x) |.
\]

Recall that we obtained the indicators and $\widetilde{t}_{k_1 k_2}^{(j)}$ via independent sample group $D_1$ and $D_2$,  thus expression above can be expanded using the law of iterated expectations and conditioned on the indicators as:

\[
= | \mathbb{E}_{D_1,D_2} (\mathbb{I}\{ \sum\limits_{k_1, k_2} \mathbb{I}_{k_1 k_2}^{(j)} ( \widetilde{t}_{k_1 k_2}^{(j)} \leq x ) \})- \Phi(x)  |
\]

\[
= | \mathbb{E}_{D_1}( \mathbb{E}_{D_2} (\mathbb{I}\{ \sum\limits_{k_1, k_2 \leq K} \mathbb{I}_{k_1 k_2} \widetilde{t}_{k_1 k_2}^{(j)} \leq x  \})|\{\mathbb{I}_{k_1 k_2}\}_{1\leq k_1,k_2\leq K} )-\Phi(x) |
\]

Once knowing the set of indicator variables $\{\mathbb{I}_{k_1 k_2}\}$, which signify the selection of the data block with the maximum or minimum estimator, implies that one of the indicators is equal to 1 while the others are 0, we can deduce the following:

\[
= | \mathbb{E}_{D_1}( \mathbb{E}_{D_2} (\mathbb{I}\{ \sum\limits_{k_1, k_2 \leq K}  \widetilde{t}_{k_1 k_2}^{(j)} \leq x  \})|\{\mathbb{I}^{(j)}_{k_1 k_2}\}_{1\leq k_1,k_2\leq K} )-\Phi(x) |
\]

By the independence of $\{\mathbb{I}^{(j)}_{k_1 k_2}\}$ and $\{\widetilde{t}_{k_1 k_2}\}$, the expression can be simplified by
:
\[
= | \mathbb{E}_{D_1}( \mathbb{E}_{D_2} (\mathbb{I}\{ \sum\limits_{k_1, k_2 \leq K}  \widetilde{t}_{k_1 k_2}^{(j)} \leq x  \}) )-\Phi(x) |
\]
\[
= | \mathbb{E}_{D_1}( \mathrm{P} ( \sum\limits_{1\leq k_1, k_2 \leq K}  \widetilde{t}_{k_1 k_2}^{(j)} \leq x  ) )-\Phi(x) |\leq \mathbb{E}_{D_1}| \mathrm{P} ( \sum\limits_{1\leq k_1, k_2 \leq K}  \widetilde{t}_{k_1 k_2}^{(j)} \leq x  )-\Phi(x)|
\]
Due to independence of $D_1$ and $D_2$, we can bound the equation above
 via 
\[
\leq \sup_{x \in \mathbb{R},1\leq k_1, k_2 \leq K} \left| \mathrm{P}\left( \widetilde{t}_{k_1 k_2}^{(j)} \leq x \right)-\Phi(x) \right|  \leq \frac{\rho_0}{\sqrt{n_{\min}}}
\]

Hence, we have:

\[
\sup_{x \in \mathbb{R},1\leq k_1, k_2 \leq K}\left|\mathrm{P}\left(\sum\limits_{k_1 k_2} \mathbb{I}^{(j)}_{k_1,k_2} \widetilde{t}_{k_1 k_2}^{(j)} \leq x \right)-\Phi(x)\right| \leq \frac{\rho_0}{\sqrt{n_{\min}}}.
\]

Thus, the sum converges in distribution to $N(0,1)$.

Now, we prove the third step that

\begin{equation}\label{eq:Tj-and-tildeTj}
    \left|T^{(j)}-\widetilde{T}^{(j)}\right|=\left|\sum\limits_{k_1 k_2} \mathbb{I}^{(j)}_{k_1,k_2}(t_{k_1 k_2}^{(j),[2]}-\tilde{t}_{k_1 k_2}^{(j),[2]})\right| \xrightarrow{P} 0,
\end{equation}

which shows the equivalence of $\sum\limits_{k_1 k_2} \mathbb{I}^{(j)}_{k_1,k_2}t_{k_1 k_2}^{(j),[2]}$ and $  \sum\limits_{k_1 k_2} \mathbb{I}^{(j)}_{k_1,k_2}\tilde{t}_{k_1 k_2}^{(j),[2]}$  thus prove the asymptotic normality of the test statistic.

To prove  \eqref{eq:Tj-and-tildeTj}
, we can bound  the difference between $t_{k_1 k_2}^{(j),[2]}$ and $\tilde{t}_{k_1 k_2}^{(j),[2]}$ by

\begin{align*}
&\mathbb{E}_{D_1, D_2} \left| \sum\limits_{k_1, k_2} \mathbb{I}_{k_1, k_2} (t_{k_1 k_2}^{(j),[2]}-\tilde{t}_{k_1 k_2}^{(j),[2]}) \right|^{v} 
= \mathbb{E}_{D_1, D_2} \sum\limits_{k_1, k_2} \mathbb{I}_{k_1, k_2} \left| (t_{k_1 k_2}^{(j),[2]}-\tilde{t}_{k_1 k_2}^{(j),[2]}) \right|^{v} \\
&= \sum\limits_{k_1, k_2} \mathbb{E}_{D_1} \mathbb{I}_{k_1, k_2} \mathbb{E}_{D_2} \left| (t_{k_1 k_2}^{(j),[2]}-\tilde{t}_{k_1 k_2}^{(j),[2]}) \right|^{v} \leq \max_{k_1, k_2} \mathbb{E}_{D_2} \left| (t_{k_1 k_2}^{(j),[2]}-\tilde{t}_{k_1 k_2}^{(j),[2]}) \right|^{v} = \mathcal{O}(n_{\min}^{-v/2}).
\end{align*}

The first equality holds because the indicators $\{\mathbb{I}_{k_1, k_2}\}$ are independent of $\{t^{(j)}_{k_1, k_2}, \tilde{t}^{(j)}_{k_1, k_2}\}$. The second equality is due to the fact that any cross terms of $\{\mathbb{I}_{k_1, k_2}\}_{k_1, k_2}$ are zero (as exactly one $\mathbb{I}_{k_1, k_2}$ equals 1). The inequality follows from Lemma \ref{lem:fangsuo_tk1k2}, which holds uniformly for any $k_1, k_2$. By lemma \ref{lem:fangsuo_tk1k2}, we know that $\mathbb{E}_{D_2} \left| (t_{k_1 k_2}^{(j),[2]}-\tilde{t}_{k_1 k_2}^{(j),[2]}) \right|^{v} =\mathcal{O}(n_{\min}^{-v/2})$.

Thus, we establish that the two statistics are asymptotically equivalent for any large $K$ and we have
\[
\sum\limits_{k_1, k_2} \mathbb{I}_{k_1, k_2} \tilde{t}_{k_1 k_2}^{(j),[2]} \xrightarrow{d} N(0, 1).
\]
For the fourth step, by Lemma \ref{lem:KS_bound}, we have
$sup_x|\mathrm{P}(\sum\limits_{k_1, k_2} \mathbb{I}_{k_1, k_2} \tilde{t}_{k_1 k_2}^{(j),[2]}\leq x)-\Phi(x)|\leq \rho_1 n_{\min}^{-\frac{v}{2(v+2)}}$ where $\rho_1$ is independent to $n,K,p.$ This completes the proof for the asymptotic normality and convergence rate of the test statistic under the null. 

Next, we prove the second part of Theorem \ref{theorem: ECT_asym_normal}, demonstrating the consistency of the ECT. Recalling we have defined  the maximum difference of true parameter: $\Delta_j = \underset{k_1 \ne k_2}{\sup} |\theta_{k_1}^{(j),*}-\theta_{k_2}^{(j),*}|$. We can express the difference between the estimator from $D_2$ and the true parameter for each \( j \)-th dimension as $
\widehat{\theta}_k^{(j),[2]}-\theta_k^{(j),*} = \frac{\sum\limits_{i=n_k(1-\gamma)+1}^{n_k} f_k^{(j)}(X_{k,i}, \theta_k^*)}{\gamma\cdot n_k} + R_k^{(j),[2]}$. Now, let us define the sets \( A_j \) and \( B_j \) as:
\[
A_j = \left\{ k' \mid \theta_{k'}^{(j),*} > \underset{k}{\max} \theta_k^{(j),*}-\frac{\Delta_j}{3} \right\}, \quad B_j = \left\{ k' \mid \theta_{k'}^{(j),*} \leq \underset{k}{\min} \theta_k^{(j),*} + \frac{\Delta_j}{3} \right\}
\]

Under the alternative hypothesis, it holds that \( |A_j| \geq 1 \), \( |B_j| \geq 1 \) and \( |A_j| + |B_j| \leq K \).

Given two device indices $1 \leq k_1 \neq k_2 \leq K$ such that $\theta_{k_1}^* > \theta_{k_2}^*$, we analyze the probability $\mathrm{P}( \widehat{\theta}_{k_1}^{(j),[1]} < \widehat{\theta}_{k_2}^{(j),[1]} )$. For $\epsilon>0$, the Markov inequality and Lemma \ref{lemma: E theta_k-theta_k* diverge p} gives the bound $\mathrm{P}(|\widehat{\theta}_{k_1}^{(j),[1]}-\theta_{k_1}^{(j),*}| > \epsilon) = O\left(\dfrac{1}{\epsilon^{2v} n_{\min}^v}\right)$ and similarly $\mathrm{P}(|\widehat{\theta}_{k_2}^{(j),[1]}-\theta_{k_2}^{(j),*}| > \epsilon) = O\left(\dfrac{1}{\epsilon^{2v} n_{\min}^v}\right)$. Moreover, denote $2\epsilon = \theta_{k_1}^{(j),*}-\theta_{k_2}^{(j),*}$, we bound the probability $\mathrm{P}\left( \widehat{\theta}_{k_1}^{(j),[1]} \leq \widehat{\theta}_{k_2}^{(j),[1]} \right)$ by noting that 
\[
\mathrm{P}\left( \widehat{\theta}_{k_1}^{(j),[1]} > \widehat{\theta}_{k_2}^{(j),[1]} \right)\geq \mathrm{P}\left(  \left\{\widehat{\theta}_{k_1}^{(j),[1]} > \theta_{k_1}^{(j),*}-\epsilon\right\} \cap \left\{\widehat{\theta}_{k_2}^{(j),[1]} < \theta_{k_2}^{(j),*} + \epsilon \right\}\right).
\]
Taking the complement on both sides gives 
\[
\mathrm{P}\left( \widehat{\theta}_{k_1}^{(j),[1]} \leq \widehat{\theta}_{k_2}^{(j),[1]} \right) \leq p\left( \left\{ \widehat{\theta}_{k_1}^{(j),[1]} \leq \theta_{k_1}^{(j),*}-\epsilon\right\} \cup \left\{\widehat{\theta}_{k_2}^{(j),[1]} \geq \theta_{k_2}^{(j),*} + \epsilon \right\}\right).
\]
Finally, applying the union bound leads to 
\[
\mathrm{P}\left( \widehat{\theta}_{k_1}^{(j),[1]} \leq \widehat{\theta}_{k_2}^{(j),[1]} \right) \leq p\left( |\widehat{\theta}_{k_1}^{(j),[1]}-\theta_{k_1}^{(j),*}| > \epsilon \right) + \mathrm{P}\left( |\widehat{\theta}_{k_2}^{(j),[1]}-\theta_{k_2}^{(j),*}| > \epsilon \right) = O\left(\dfrac{1}{\epsilon^{2v} n_{\min}^v}\right).
\]
Noting that the conclusion above holds for all dimension and data block uniformly due to the same constants in regularity assumptions, thus we get the probability to choose approximately wrong:
\begin{equation}\label{consistentbound}
    \PP(\widehat{\theta}_{k_1}^{(j),[1]} < \widehat{\theta}_{k_2}^{(j),[1]})=O\left( \frac{(\theta_{k_1}^{(j),*}-\theta_{k_2}^{(j),*})^{-2v}}{n_{\min}^v} \right) \text{ holds uniformly for }1\leq j \leq p, 1\leq k_1 \neq k_2 \leq K.
\end{equation}

Continually, if we have chosen the data block approximately right, i.e.  \( \PP(k_{max}^{(j)} \in A_j) \), meaning that $\theta_{k_{max}^{(j)}}^{(j),*} > \underset{k \in A_j}{\max} \theta_k^{(j),*}-\dfrac{\Delta_j}{3},$ then we say that our selection of the maximum estimator is approximately correct.

Next, we bound the probability \( \PP(k_{max}^{(j)} \notin A_j) \):

\[
  \PP(k_{max}^{(j)} \notin A_j) = \mathrm{P}\left( \exists k_2 \notin A_j \text{ such that } \widehat{\theta}_{k_2}^{(j),[1]} > \widehat{\theta}_{k'}^{(j),[1]} = \underset{k \in A_j}{\max} \widehat{\theta}_k^{(j),[1]} \right)
\]

Note that \( k' \) does not necessarily equal \( k_{max}^{(j)} \) and if \( k' = k_{max}^{(j)} \), then our selection of the maximum estimator is approximately correct. By Boole's inequality, we have:

\[
  \leq \sum\limits_{k_2 \notin A_j} \mathrm{P}\left( \widehat{\theta}_{k_2}^{(j)} > \widehat{\theta}_{k'}^{(j),[1]} = \underset{k \in A_j}{\max} \widehat{\theta}_k^{(j),[1]} \right)
\]

Let \( k'' = \underset{k \in A_j}{\arg \max} \theta_k^{(j),*} \), which is the index of the data block that maximizes the true parameter value. Since \( k'' \in A_j \), we have:

\[
  \widehat{\theta}_{k''}^{(j),[1]} \leq \widehat{\theta}_{k'}^{(j),[1]} = \underset{k \in A_j}{\max} \widehat{\theta}_k^{(j),[1]}.
\]

By the monotonicity of probability, we then have:

\[
  \leq \sum\limits_{k_2 \notin A_j} \mathrm{P}\left( \widehat{\theta}_{k_2}^{(j),[1]} > \widehat{\theta}_{k''}^{(j),[1]} \right)
\]

Since for \( k \notin A_j \), \( \theta_k^{(j),*} \leq \underset{k \in A_j}{\arg \max} \theta_k^{(j),*}-\frac{\Delta_j}{3} \) by the definition of \( A_j \) and use \eqref{consistentbound}, we get:

\[
  \leq (K-1) \cdot O\left( \frac{(\Delta_j/3)^{-2v}}{n_{\min}^v} \right) = O\left( \frac{K \Delta_j^{-2v}}{n_{\min}^v} \right).
\]

Similarly, we can derive $
\PP(k_{min}^{(j)} \notin B_j) = O\left( \frac{K \Delta_j^{-2v}}{n_{\min}^v} \right)$. Thus, we obtain: $\PP(k_{max}^{(j)} \in A_j \cap k_{min}^{(j)} \in B_j) = 1-O\left( \frac{K \Delta_j^{-2v}}{n_{\min}^v} \right)$. For any \( x > 0 \) and \( k_1 \in A_j \), \( k_2 \in B_j \), we have \( \theta_{k_1}^{(j),*}-\theta_{k_2}^{(j),*} > \frac{\Delta_j}{3} \). Therefore: $\PP(|t_{k_1 k_2}^{(j),[2]}| < x)  = O\left( \frac{(\Delta_j/3)^{-2v}}{n_{\min}^v} \right) $. Finally, we conclude that: $\PP(|T^{(j)}| < x) = O\left( \frac{K \Delta_j^{-2v}}{n_{\min}^v} \right)$ under the alternative hypothesis \( \mathcal{H}_{1}^{(j)} \).\end{proof}

\subsection{Proof for Theorem \ref{theorem:tilde_W_NK_normal_independence_ECT}}\label{appendix: proof for combination consistent}
\begin{proof}
Without loss of generality (WLOG), we assume $\gamma = 0.5$ in the following. The proof for other fixed $\gamma\in(0,1)$ is similar. Denote $\widetilde{\theta}_{k}^{(j)}-\theta_k^{(j),*}=\dfrac{1}{n_k}\sum_{i = 1}^{n_k}f_k^{(j)}(X_{k,i},\theta_k^{*})$, the influence function without bias term a and $\widetilde{W}_{N,K}^{(j)}$ is a modified version of Wald statistic using $\{\widetilde{\theta}_{k}^{(j)}\}_{k \geq 1},\{r_{k_1,k_2}\}_{k_1 \ne k_2}$ instead of $\{\widehat{\theta}_{k}^{(j)}\}_{k \geq 1},\{\widehat{r}_{k_1,k_2}\}_{k_1 \ne k_2}$. 
In the proof \ref{appendix: Wald consistency proof} and proof for Lemma \ref{lemma: A_k_1,k_2}, Lemma \ref{lemma: BCD_k1k2}, we known that $\widetilde{W}_{N,K}^{(j)}\overset{d}{\to} \mathcal{N}(0,1).$ We also denote a $\breve{\theta}_{k}^{(j)}-\theta_k^{(j),*}=\dfrac{1}{n_k/2}\sum\limits_{i = 1}^{n_k/2}f_k^{(j)}(X_{k,i},\theta_k^{*})$ and $\check{W}_{K,k_1,k_2}^{(j)}$ is a modified version of $\widetilde{W}_{N,K}^{(j)}$ using $\breve{\theta}_{k_1}^{(j)}$ and $\breve{\theta}_{k_2}^{(j)}$ instead of $\widetilde{\theta}_{k_1}^{(j)}$ and $\widetilde{\theta}_{k_2}^{(j)}$ for a fixed pair $(k_1,k_2)$. We defined $\mathbb{I}_{k_1,k_2}^{(j)}\check{W}_{K,k_1,k_2}^{(j)}$ as $\check{W}_{N,K}^{(j)}$.

Firstly we show that  $\widetilde{W}_{N,K}^{(j)}-\sum\limits_{1\leq k_1 \ne k_2 \leq K}\mathbb{I}_{k_1,k_2}^{(j)}\check{W}_{K,k_1,k_2}^{(j)}\to 0$ as $K\to\infty$ where $\mathbb{I}_{k_1, k_2}^{(j)}=\mathbb{I}\{k_1=\arg\max_k \widehat{\theta}_k^{(j),[1]},k_2=\arg\min_k \widehat{\theta}_k^{(j),[1]}\}$. 
We have the following expansion:
\begin{align*}
(\breve{\theta}_{k_1}^{(j)}-\widetilde{\theta}_{k}^{(j)})^2-(\breve{\theta}_{k_1}^{(j)}-\widetilde{\theta}_{k}^{(j)})^2 &= \underbrace{\dfrac{2Q_{k_1}}{n_{k_1}}\sum_{i = 1}^{n_{k_1/2}}f_{k_1}^{(j)} (X_{k_1,i},\theta_{k_1}^{*})}_{J_{k_1,k}}\underbrace{-\frac{2Q_{k_1}}{n_{k}}\sum_{i = 1}^{n_{k}}f_{k}^{(j)} (X_{k,i},\theta_{k}^{*})}_{L_{k_1,k}}+\underbrace{Q_{k_1}^2}_{M_{k_1}}
\end{align*}
Where $Q_{k_1}=\dfrac{1}{n_{k_1}}\sum_{i = 1+n_{k_1/2}}^{n_{k_1}}f_{k_1}^{(j)} (X_{k_1,i},\theta_{k_1}^{*})=O_p(1/\sqrt{n_{k_1}})$ is measureble to the observation $D_{k_1}^{[2]}$. Since the difference of $\widetilde{W}_{N,K}^{(j)}$ and $\check{W}_{K,k_1,k_2}^{(j)}$ only appears on the terms related to indices $k_1,k_2$, thus we only need to prove that $\sum_{k\neq k_1} \dfrac{1}{\sqrt{2K-2}}\sum\limits_{1\leq k_1 \ne k_2 \leq K}\mathbb{I}_{k_1,k_2}^{(j)}J_{k_1,k}\cdot r_{k_1, k} \to 0$,  $\sum_{k\neq k_1} \dfrac{1}{\sqrt{2K-2}}\sum\limits_{1\leq k_1 \ne k_2 \leq K}\mathbb{I}_{k_1,k_2}^{(j)}L_{k_1,k}\cdot r_{k_1, k} \to 0$, and $\sum_{k\neq k_1} \dfrac{1}{\sqrt{2K-2}}\sum\limits_{1\leq k_1 \ne k_2 \leq K}\mathbb{I}_{k_1,k_2}^{(j)}M_{k_1,k}\cdot r_{k_1, k} \to 0$. However, these
have already been proven in Lemma \ref{lemma: JLM_K1K2}, and thus the claim is established. 

Then, we prove the asymptotic independence of $\check{W}_{N,K}^{(j)}$ and $T_{N,K}^{(j)}$ when $K = o(n_{\min})$. Notice from the proof of Theorem \ref{theorem: ECT_asym_normal} we have the KS distance between $T_{N,K}^{(j)}$ and standard normal is $\mathcal{O}(n_{\min}^{-\frac{v}{2(v+2)}})$. 

We start with the joint probability:
\[
\mathrm{P}\left(\check{W}_{N,K}^{(j)} \leq t, T_{N,K}^{(j)} \leq s\right)=\mathbb{E}_{k_1, k_2} \left[ \mathrm{P}\left(\check{W}_{N,K}^{(j)} \leq t, T_{N,K}^{(j)} \leq s \,\big|\, \hat{k}^{(j)}_{\max} = k_1, \hat{k}^{(j)}_{\min} = k_2\right) \right].
\]

Given \(\hat{k}^{(j)}_{\max}=k_1\) and \(\hat{k}^{(j)}_{\min}=k_2\), \(T_{N,K}^{(j)}\) is measurable with respect to the \(\sigma -\)algebra generated by \(D_{k_1}^{[2]}\) and \(D_{k_2}^{[2]}\), while \(\check{W}_{N,K}^{(j)}\) is measurable with respect to the \(\sigma -\)algebra generated by \(D\setminus (D_{k_1}^{[2]} \cup D_{k_2}^{[2]})\). Since the \(\sigma -\)algebras generated by these non-overlapping data subsets are independent in the relevant sense, we can rewrite the inner conditional probability as a product. So,

\begin{align*}
&\mathbb{E}_{k_1, k_2} \left[ \mathrm{P}\left(\check{W}_{N,K}^{(j)} \leq t, T_{N,K}^{(j)} \leq s \,\big|\, \hat{k}^{(j)}_{\max} = k_1, \hat{k}^{(j)}_{\min} = k_2\right) \right]\\
=&\mathbb{E}_{k_1, k_2} \left[ \mathrm{P}\left(\check{W}_{N,K}^{(j)} \leq t \,\big|\, \hat{k}^{(j)}_{\max} = k_1, \hat{k}^{(j)}_{\min} = k_2\right) \cdot \mathrm{P}\left(T_{N,K}^{(j)} \leq s \,\big|\, \hat{k}^{(j)}_{\max} = k_1, \hat{k}^{(j)}_{\min} = k_2\right) \right].
\end{align*}

According to the proof in \(\ref{appendix: proof for ECT consistent}\), for any given \(\hat{k}^{(j)}_{\max}=k_1\) and \(\hat{k}^{(j)}_{\min}=k_2\), 
\[
\mathrm{P}\left(T_{N,K}^{(j)} \leq s \,\big|\, \hat{k}^{(j)}_{\max} = k_1, \hat{k}^{(j)}_{\min} = k_2\right)=\Phi(s)+O\left(n_{\min}^{-\frac{v}{2(v + 2)}}\right).
\]
uniformly. We can then factor out this known conditional probability of \(T_{N,K}^{(j)}\) from the expectation, obtaining

\begin{align*}
&\mathbb{E}_{k_1, k_2} \left[ \mathrm{P}\left(\check{W}_{N,K}^{(j)} \leq t \,\big|\, \hat{k}^{(j)}_{\max} = k_1, \hat{k}^{(j)}_{\min} = k_2\right) \cdot \mathrm{P}\left(T_{N,K}^{(j)} \leq s \,\big|\, \hat{k}^{(j)}_{\max} = k_1, \hat{k}^{(j)}_{\min} = k_2\right) \right]\\
=&\mathbb{E}_{k_1, k_2} \left[ \mathrm{P}\left(\check{W}_{N,K}^{(j)} \leq t \,\big|\, \hat{k}^{(j)}_{\max} = k_1, \hat{k}^{(j)}_{\min} = k_2\right) \right]\cdot \left(\Phi(s) + \mathcal{O}( n_{\min}^{-\frac{v}{2(v + 2)}})\right).
\end{align*}

By the law of total probability again,
\[
\mathbb{E}_{k_1, k_2} \left[ \mathrm{P}\left(\check{W}_{N,K}^{(j)} \leq t \,\big|\, \hat{k}^{(j)}_{\max} = k_1, \hat{k}^{(j)}_{\min} = k_2\right) \right]=\mathrm{P}\left(\check{W}_{N,K}^{(j)} \leq t\right).
\]
So, $\mathbb{E}_{k_1, k_2} \left[ \mathrm{P}\left(\check{W}_{N,K}^{(j)} \leq t \,\big|\, \hat{k}^{(j)}_{\max} = k_1, \hat{k}^{(j)}_{\min} = k_2\right) \right]\cdot \left(\Phi(s) + \mathcal{O}( n_{\min}^{-\frac{v}{2(v + 2)}})\right)$
\[
=\mathrm{P}\left(\check{W}_{N,K}^{(j)} \leq t\right) \left(\Phi(s) + \mathcal{O}( n_{\min}^{-\frac{v}{2(v + 2)}})\right).
\]

Aboveall, we proved the asymptotic independence of $\check{W}_{N,K}^{(j)}$ and $T_{N,K}^{(j)}$, we also proved that ${W}_{N,K}^{(j)}-\check{W}_{N,K}^{(j)}\to 0,$ thus Theorem \ref{theorem:tilde_W_NK_normal_independence_ECT} is proved.
\end{proof}

\subsection{Proof for Theorem \ref{Theorem: local_alternative_theorem}}\label{appendix: PROOF FOR GAUSSIAN comparison}
\begin{proof}
The above proof are for all dimension $j\in[p]$, so with a little abuse of notation we omit the subscript $(j)$ here. Denote set $A=\{k|\theta_k^*=\theta_0+\mu(K,n)\}.$ Denote $k_{\max}=\arg\max_k \widehat{\theta}_k^{[1]},k_{\min}=\arg\min_k \widehat{\theta}_k^{[1]}$ and event $E=\{k_{max}\in A,k_{\min}\notin A\}$ which means we choose the right indices in the first stage of ECT. Since $P(T>Z_\alpha)=P(E\cap \{T>Z_\alpha\})+P(E^c\cap \{T>Z_\alpha\})$ we will prove that when $\beta>0.5,$ $P(E)\to 1$ and $P(E\cap \{T>Z_\alpha\})\to 1$,  when $\beta>0.5,$ $P(E)\to 0$ and $P(T>Z_\alpha)\to P(E^c\cap \{T>Z_\alpha\})\to \alpha.$

For $\epsilon>0$ to be specified later, by Lemma \ref{lem:maxima_logK}, we have $$\PP(\max_{k\in A} \sqrt{(1-\gamma)n}(\widehat{\theta}^{[1]}_k-\theta_0-\mu(K,n))/{\sigma_k}\in (\sqrt{2(1-\epsilon)(1-\beta)\log K}, \sqrt{2(1+\epsilon)(1-\beta)\log K}))\to 1,$$ 
$$\PP(\max_{k\notin A} \sqrt{(1-\gamma)n}(\widehat{\theta}^{[1]}_k-\theta_0)/{\sigma_k}\in (\sqrt{2(1-\epsilon)\log K}, \sqrt{2(1+\epsilon)\log K}))\to 1.$$ 

Thus we have $\PP( \widehat{\theta}^{[1]}_k-\theta_0-\mu(K,n)$
$$\in(-\sigma^{+}\sqrt{2(1+\epsilon)(1-\beta)(1-\gamma)n\log K},\sigma^{+}\sqrt{2(1+\epsilon)(1-\beta)(1-\gamma)n\log K} )\text{ for all }k\in A)\to1,$$

$\PP( \widehat{\theta}^{[1]}_k-\theta_0$
$$\in(-\sigma^{+}\sqrt{2(1+\epsilon)(1-\gamma)n\log K},\sigma^{+}\sqrt{2(1+\epsilon)(1-\gamma)n\log K} )\text{ for all }k\notin A)\to1.$$

Take $\epsilon>0$ small enough and we get the events in the two probability expression above are disjoint if $\mu(K,n)>\sigma^+\sqrt{2n\rho(\beta)(1-\gamma)\log K}$ which means $c_n>\rho(\beta)(\sigma^+)^2(1-\gamma)^{-1}.$ The disjoint events means we select the correct indices with maximum and minimum blocks, thus
 we have $ET_{N,K}^{(j)}=\sum_{}\one^{[1]}t_{k_1k_2}^{[2]}\to\sum_{}1_{k_1 k_2,k_1\in A,k_2\in B}t_{k_1k_2}^{[2]}\asymp\sqrt{c_n\log K}$. And by proof for Theorem \ref{theorem: ECT_asym_normal}. We also have that $T_{N,K}^{(j)}-\EE T_{N,K}^{(j)}\to \mathcal{N}(0,1)$.

 When $\mu(K,n)<\sigma^-\sqrt{2n\rho(\beta)(1-\gamma)\log K}$, due to we have $$\PP(\max\limits_{k\in A}\widehat\theta_k^{[1]}>\theta_0+\mu(K,n)+\sigma^-\sqrt{2(1-\epsilon)(1-\beta)\log K/(n(1-\gamma))})\to 1,$$ and $$\PP(\max\limits_{k\notin A}\widehat\theta_k^{[1]}<\theta_0+\sigma^+\sqrt{2(1-\epsilon)\log K/(n(1-\gamma))})\to 1,$$ 
 thus we have $\PP (\arg\max_k \theta_k^{[1]}\notin A)\to 1$ which means $ET_{N,K}^{(j)}\to 0.$ Thus we have $T_{N,K}^{(j)}\to \mathcal{N}(0,1)$ by proof for Theorem \ref{theorem: ECT_asym_normal}. Then the proof for ECT's asymptotic power under the local alternative is done. 

Next, we prove the power calculation for the re-normalized Wald. Noting that $W_{N,K}=\sum\limits_{1\leq k_1<k_2\leq K} \widehat r_{k_1,k_2} (\widehat{\theta}_{k_1}-\widehat{\theta}_{k_2})^2$ where $\widehat r_{k_1,k_2}=\dfrac{n(\widehat\sigma_{k_1})^{-2}(\widehat\sigma_{k_2})^{-2}}{\sum\limits_{l=1}^K(\widehat\sigma_{l})^{-2}}$. Under alternative hypothesis,
\begin{align*}
      (\widehat{\theta}_{k_1}-\widehat{\theta}_{k_2})^2 =& \underbrace{\dfrac{1}{n^2}\left(\sum\limits_{i=1}^{n}f_{k_1} (X_{k_1,i},\theta_{k_1}^{*})\right)^{2} + \dfrac{1}{n^2}\left(\sum\limits_{i=1}^{n}f_{k_2} (X_{k_2,i},\theta_{k_2}^{*})\right)^{2}}_{A_{k_1,k_2}^{(j)}} \\
    -& \underbrace{\dfrac{1/\gamma}{n^2}\left(\sum\limits_{i=1}^{n}f_{k_1} (X_{k_1,i},\theta_{k_1}^{*})\right)\left(\sum\limits_{i=1}^{n}f_{k_2} (X_{k_2,i},\theta_{k_2}^{*})\right)}_{B_{k_1,k_2}^{(j)}} \\ 
    +& \underbrace{2\left(\dfrac{1}{n}\sum\limits_{i=1}^{n}f_{k_1} (X_{k_1,i},\theta_{k_1}^{*})-\dfrac{1}{n}\sum\limits_{i=1}^{n}f_{k_2} (X_{k_2,i},\theta_{k_2}^{*})\right)(R_{k_1}-R_{k_2})}_{C_{k_1,k_2}^{(j)}} \\
    +& \underbrace{(R_{k_1}-R_{k_2})^2}_{D_{k_1,k_2}^{(j)}}+\underbrace{2\left(\dfrac{1}{n}\sum\limits_{i=1}^{n}f_{k_1} (X_{k_1,i},\theta_{k_1}^{*})-\dfrac{1}{n}\sum\limits_{i=1}^{n}f_{k_2} (X_{k_2,i},\theta_{k_2}^{*})+R_{k_1}-R_{k_2}\right)(\theta_{k_1}^*-\theta_{k_2}^*)}_{G_{k_1,k_2}}\\
    +&\underbrace{(\theta_{k_1}^*-\theta_{k_2}^*)^2}_{H_{k_1,k_2}}
\end{align*}
In the proof \ref{appendix: Wald consistency proof} we know that $\sum\limits_{1\leq k_1<k_2\leq K} \widehat r_{k_1,k_2}A_{k_1,k_2}^{(j)}\to N(0,1)$ and with Lemma \ref{lem:G_H_k1k2_local_noncentral},  the sum of terms related to $B,C,D,G$ is $o(1)$, also:
$$\dfrac{\dfrac{1}{\sqrt{2K-2}}\sum\limits_{1\leq k_1<k_2\leq K} \widehat r_{k_1,k_2}H_{k_1,k_2}}{2^{-1/2}nK^{0.5-\beta}\mu^2\sigma_j^{-2}}\to1.$$
Noting that $\mu(K,n)^2\sim c_{\infty}\log K/n,$ by definition, thus when $\beta\leq0.5$, we have $\nu_{N,K}^{(j)}$ diverges and we have ${W_{N,K}^{(j)}-\nu_{N,K}^{(j)}}\to  \mathcal{N}(0,1)$ 
 by Slutsky's theorem where $\nu_{N,K}^{(j)}\approx 2^{-1/2}nK^{0.5-\beta}\mu^2\sigma_j^{-2}=c_n2^{-1/2}K^{0.5-\beta}\sigma_j^{-2}\log K$ by Lemma 27.

For the combined test's SNR under the local alternative, since when $K=o(\sqrt{n})$ the weight equals to 1, thus it's a equal weighted sum of Wald and ECT, and the conclusion is derived from the proof above for Wald and ECT.
\end{proof}

\section{Proofs for lemmas}\label{appendix: lemma proofs}
\begin{lemma}\label{lemma:Projection_Matrix_expansion}
    Recall $R_K$  in  \eqref{lem:R_k_fixed_p},  for a positive-definite diagonal matrix $\Lambda=\text{diag}\{\lambda_1^2,\cdots,\lambda_K^2\}$, for $1\leq i,j\leq K$, we have 
   \[
  (R_K^\top(R_K\Lambda R_K^\top)^{-1}R_K)_{i,j} =
  \begin{cases}
    \dfrac{\prod\limits_{k\ne i,j}\lambda_k^2}{\sum\limits_{l=1}^{K} \prod\limits_{k\ne l}\lambda_k^2}=\dfrac{\lambda_i^{-2}\lambda_j^{-2}}{\sum\limits_{l=1}^{K} \lambda_l^{-2}} & \text{when } i \ne j, \\
    \dfrac{\sum\limits_{s \ne i}\prod\limits_{k\ne s,i}\lambda_k^2}{\sum\limits_{l=1}^{K} \prod\limits_{k\ne l}\lambda_k^2} & \text{when } i = j.
  \end{cases}
\]

\end{lemma}
Following lemma shows the relationship between $\nu_{N,K}^{(j)}$ and $\Delta_j$.
\begin{lemma}\label{lem:min_square_diff}
For the true parameters \(\theta_1^{(j),*},\theta_2^{(j),*},\cdots,\theta_K^{(j),*}\), the following inequality holds:
\[
\dfrac{K^2\Delta_j^2}{4}\geq\sum\limits_{1\leq k_1 < k_2\leq K}(\theta_{k_1}^{(j),*}-\theta_{k_2}^{(j),*})^2\geq\frac{K}{2}\Delta_j^2.
\]
When $\Delta_j>0$, suppose  \(\theta_{1}^{(j),*}\) is the minimum and \(\theta_K^{(j),*}\) is the maximum (we can arbitarily choose one if there're mutliple extreme values),  the latter equality holds if and only if \(\theta_2^{(j),*}=\cdots=\theta_{K-1}^{(j),*}=\frac{\theta_1^{(j),*}+\theta_K^{(j),*}}{2}\).
\end{lemma}

\begin{proof}
Without loss of generality, assume that \(\theta_1^{(j),*}\leq\theta_2^{(j),*}\leq\cdots\theta_{K-1}^{(j),*}\leq\theta_K^{(j),*}\).

Firstly, we consider the lower bound. For the sum \(\sum\limits_{1\leq k_1 < k_2\leq K}(\theta_{k_1}^{(j),*}-\theta_{k_2}^{(j),*})^2\), we can bound it from below. First, we discard most of the terms and only keep some of them. We have \(\sum\limits_{1\leq k_1 < k_2\leq K}(\theta_{k_1}^{(j),*}-\theta_{k_2}^{(j),*})^2\geq\sum\limits_{m = 2}^{K-1}[(\theta_1^{(j),*}-\theta_m^{(j),*})^2+(\theta_K^{(j),*}-\theta_m^{(j),*})^2]+(\theta_1^{(j),*}-\theta_K^{(j),*})^2\). Then, by the inequality \(a^2 + b^2\geq\frac{1}{2}(a-b)^2\), we get \(\sum\limits_{m = 2}^{K-1}[(\theta_1^{(j),*}-\theta_m^{(j),*})^2+(\theta_K^{(j),*}-\theta_m^{(j),*})^2]+(\theta_1^{(j),*}-\theta_K^{(j),*})^2\geq\sum\limits_{m = 2}^{K-1}[\frac{1}{2}(\theta_1^{(j),*}-\theta_K^{(j),*})^2]+(\theta_1^{(j),*}-\theta_K^{(j),*})^2=(K-2)\times\frac{1}{2}\Delta_j^2+\Delta_j^2=\frac{K}{2}\Delta_j^2\).

The equality holds if and only if for all \(m\in\{2,\cdots,K-1\}\), \(\theta_m^{(j),*}=\frac{\theta_1^{(j),*}+\theta_K^{(j),*}}{2}\). In this case, the differences among the middle terms are automatically zero, and the compatibility with the discarded terms is verified.

For the upper bound, we will prove that the maximum of $\sum\limits_{1\leq k_1 < k_2\leq K}(\theta_{k_1}^{(j),*}-\theta_{k_2}^{(j),*})^2$ is $\dfrac{K^2\Delta_j^2}{4}$ if $K$ is even and $\dfrac{(K^2-1)\Delta_j^2}{4}$ if $K$ is odd.  This bound is trivial when $K=1,2,3.$ For $K\geq 3,$ it's easy to see if we replace $\theta_2^{(j),*}$ with $\theta_1^{(j),*}$, the sum can be bigger. That's because after replacement, $(\theta_2^{(j),*}-\theta_k^{(j),*})^2$ becomes bigger where $k\ne 1,K$. And $(\theta_2^{(j),*}-\theta_1^{(j),*})^2+(\theta_2^{(j),*}-\theta_K^{(j),*})^2$ also becomes bigger after replacement because $\theta_1^{(j),*}\leq\theta_2^{(j),*}\leq\theta_K^{(j),*}$. The remaining terms doesn not change. After that it is worth mentioning after replacement we can not replace replace $\theta_3^{(j),*}$ with $\theta_1^{(j),*}$ because the sum can not become bigger after this replacement. But  we can replace $\theta_{K-1}^{(j),*}$ with $\theta_K^{(j),*}$ for same reason at this time. Thus WLOG we can assume $\theta_{1}^{(j),*}=\theta_2^{(j),*}$, and $\theta_{K-1}^{(j),*}=\theta_K^{(j),*}$. Similarly we can assume $\theta_{1}^{(j),*}=\theta_2^{(j),*}=\theta_{[K/2]}^{(j),*}$ and $\theta_{[K/2]+1}^{(j),*}=\theta_{[K/2]+2}^{(j),*}=\theta_{K}^{(j),*}$. Under this condition it can be shown that the sum is $\dfrac{K^2\Delta_j^2}{4}$ if $K$ is even and $\dfrac{(K^2-1)\Delta_j^2}{4}$ if $K$ is odd, which is actually the maximum.
\end{proof}

\begin{lemma}\label{lem:KS_bound}
The Kolmogorov-Smirnov distance between two random variables $U$ and $V$ is defined as $d_{KS}(U, V)=\sup_{t\in\mathbb{R}}|P(U\leq t)-P(V\leq t)|$, which measures the maximum vertical distance between the cumulative distribution functions of $U$ and $V$.

Let $X_n$ be a sequence of random variables and $Y\sim N(0,1)$ such that: 
\begin{enumerate}[label=(\roman*)]
    \item There exist positive constants $M_1$  such that  $\sup_{t\in\mathbb{R}} |P(X_n \leq t)-P(Y \leq t)|\leq M_1n^{-1/2}$.
    \item There exists an integer $v \geq 1$ and positive constants $M_2$ such that $E|X_n-Z_n|^{v/2}\leq M_2n^{-v/4}$, where $Z_n$ is another random variable.
\end{enumerate}
Then the Kolmogorov-Smirnov distance between $Z_n$ and $Y$ satisfies:
\[
\sup_{t\in\mathbb{R}} |P(Z_n \leq t)-P(Y \leq t)| = \mathcal{O}\left(n^{-\frac{v}{2(v+2)}}\right).
\]
\end{lemma}

\begin{proof}
See the equation (10.2) in \cite{nonlinear_shao}.
\end{proof}

\begin{lemma}\label{lem:maxima_logK}
 When $n_1=n_2=\cdots=n_K=n,$   under  Assumption  \ref{assumption:identifiability_compactness}-\ref{assumption: Local Strong Convexity} and both null hypothesis and local alternative defined in  \eqref{eq:gaussian_hypothesis}, when $K=o(\sqrt{n})$, for $\forall \epsilon>0$, we have $P(\max_k \sqrt{n_k}(\widehat{\theta}^{(j)}_k-\theta_k^{(j),*})/{\sigma}_k^{(j)}\in(\sqrt{2(1-\epsilon)\log K},\sqrt{2(1+\epsilon)\log K}))\to 1$ as $K\to\infty$.
\end{lemma}
\begin{proof}
   Noting that  $$\sqrt{n}(\widehat{\theta}^{(j)}_k-{\theta}_k^{(j),*})=\dfrac{\sum_{i=1}^{n}f_k^{(j)}(X_{k,i},{\theta}_k^{*})}{\sqrt{n}}+\sqrt{n}R_k^{(j)}.$$
   Since $f_k^{(j)}(X_{k,i},{\theta}_k^{*})$ are i.i.d with mean zero and variance $\sigma_k^{(j)}\in[\sigma^{-},\sigma^{+}]$, thus by Berry-Esseen theorem, we have $$\sup_x|\PP(\dfrac{\sum_{i=1}^{n}f^{(j)}(X_{k,i},{\theta}_k^{*})}{\sqrt{n}\sigma_k^{(j)}}\leq x)-\Phi(x)|\leq C_1\cdot\dfrac{R^3}{(\rho^-)^3\sqrt{n}}$$ where $C_1$ is the constant derived from the Berry-Esseen bound, we have \[\sup_x\biggl|\prod\limits_{k=1}^K\PP(\dfrac{\sum_{i=1}^{n}f^{(j)}(X_{k,i},{\theta}_k^{*})}{\sqrt{n}\sigma_k^{(j)}}\leq x)-\Phi^K(x)\biggl|\leq C_1\cdot\dfrac{R^3}{(\rho^-)^3}\sum\limits_{k=1}^K n^{-1/2}.\]

   Thus \[\PP\left(\max_k\dfrac{\sum_{i=1}^{n}f^{(j)}(X_{k,i},{\theta}_k^{*})}{\sqrt{n}\sigma_k^{(j)}}\leq \sqrt{2(1+\epsilon)\log K}\right)=\prod\limits_{k=1}^K\PP\left(\dfrac{\sum_{i=1}^{n}f^{(j)}(X_{k,i},{\theta}_k^{*})}{\sqrt{n}\sigma_k^{(j)}}\leq \sqrt{2(1+\epsilon)\log K}\right)\]
   \[ \geq \Phi^K(\sqrt{2(1+\epsilon)\log K})-C_1\cdot\dfrac{R^3}{(\rho^-)^3}\sum\limits_{k=1}^K n^{-1/2}.\]

   When $K=o(\sqrt{n})$ and $K,n\to\infty$, we have  $\Phi^K(\sqrt{2(1+\epsilon)\log K})\to 1$ and $\sum\limits_{k=1}^K n^{-1/2}\to 0,$ which means $\PP\left(\max_k\dfrac{\sum_{i=1}^{n}f^{(j)}(X_{k,i},{\theta}_k^{*})}{\sqrt{n}\sigma_k^{(j)}}\leq \sqrt{2(1+\epsilon)\log K}\right)\to 1.$ Similarly we have $\PP\left(\max_k\dfrac{\sum_{i=1}^{n}f^{(j)}(X_{k,i},{\theta}_k^{*})}{\sqrt{n}\sigma_k^{(j)}}\geq \sqrt{2(1-\epsilon)\log K}\right)\to 1.$ 

   For the remainder term $R_k^{(j)},$ since Lemma \ref{lem:R_k_fixed_p}, we have $\EE(|R_k^{(j)}|^{v/2})=\mathcal{O}(\dfrac{1}{n^{v/2}}),$ uniformly holds, thus for $\epsilon>0$, $\PP(\sqrt{n}|R_k^{(j)}|>\epsilon)\leq\dfrac{\EE(\sqrt{n}|R_k^{(j)}|)^{v/2}}{\epsilon^{v/2}}=\mathcal{O}(\dfrac{1}{n^{v/4}})$, which means $\PP(\max_k\sqrt{n}|R_k^{(j)}|>\epsilon)=\mathcal{O}(\dfrac{K}{n^{v/4}})$, thus $\max_k\sqrt{n}|R_k^{(j)}|\to 0$ when $K=\mathcal{O}({n^{v/4}})=\mathcal{O}(\sqrt{n})$ since  $v\geq 2.$
   Thus  $\max_k \sqrt{n}(\widehat{\theta}^{(j)}_k-{\theta}_k^{*})/{\sigma}_k^{(j)}\in(\sqrt{2(1-\epsilon)\log K},\sqrt{2(1+\epsilon)\log K})$. 
\end{proof}

\begin{lemma}\label{lemma: E theta_k-theta_k* diverge p}
    Under Assumptions \ref{assumption:identifiability_compactness}–\ref{assumption: Local Strong Convexity}, $\mathbb{E}\|\widehat{\theta}_k-\theta^*_k\|_2^{2v}=O\left({n_{k}^{-v}}\right)$.
\end{lemma}
\begin{proof}
  See the proof for Lemma \ref{lemma:good_event} and the proof for Lemma 8 in \cite{jmlr2013}.\end{proof}

For estimating the variance $\{\sigma_k^{(j)}\}_{1 \leq k \leq K, 1 \leq j \leq p}$, we need the following lemma.
\begin{lemma}\label{lemma: variance_diverge_p}
Under the given Assumptions \ref{assumption:identifiability_compactness}–\ref{assumption: Local Strong Convexity}, we have
\begin{align*}
\mathbb{E}\left\|P_k+Q_k\right\|_2^{v}=\mathbb{E}\left\| \dfrac{1}{n_k} \sum\limits_{i=1}^{n_k} \nabla_{\theta_k}^2 M_k(X_{k,i}, \widehat{\theta}_k)-\mathbf{E} \nabla_{\theta_k}^2 M_k(X_k, \theta^*_k) \right\|_2^{v} &= O\left( n_{k}^{-v/2} \right), \\
\mathbb{E}\left\| \dfrac{1}{n_k} \sum\limits_{i=1}^{n_k} Z_k(X_{k,i}, \widehat{\theta}_k)-\mathbf{E} Z_k(X_k, \theta^*_k) \right\|_2^{v} &= O\left(  n_{k}^{-v/2} \right).
\end{align*}

uniformly in $1\leq k \leq K.$ Thus, we directly obtain that \(\mathbb{E}  \left\| (\widehat{\sigma}_k)^2-(\sigma_k)^2 \right\|_2^{v} = \mathcal{O}( n_{\min}^{-v/2})\), which holds uniformly for \(1 \leq k \leq K\).

\end{lemma}
\begin{proof}
The first one can be derived from the proof for Lemma 9 in \cite{jmlr2013}. Additionally, 
\begin{align*}
\left\|\dfrac{1}{n_k} \sum\limits_{i=1}^{n_k}Z_k(X_{k,i},\widehat{\theta}_k)-\mathbf{E}Z_k(X_k,\theta^*_k)\right\|_2 &\le\dfrac{1}{n_k} \sum\limits_{i=1}^{n_k}B(X_{k,i})\cdot\left\|\widehat{\theta}_k-\theta^*_k\right\|_2\\
&+\left\|\dfrac{1}{n_k} \sum\limits_{i=1}^{n_k}Z_k(X_{k,i},\theta^*_k)-\mathbf{E}Z_k(X_k,\theta^*_k)\right\|_{2}
\end{align*}
And the rest is same to the first one.
 \end{proof}
\begin{lemma}\label{lem:fangsuo_tk1k2}
For a fixed tuple $(k_1, k_2)$, we have $ \left\| t_{k_1k_2}-\tilde{t}_{k_1k_2} \right\|_2^{v/2}=\mathcal{O}(n_{\min}^{-v/4})$. Moreover, when $v\geq 4$, we have $\sum_{j\in\mathcal{H}_0}|t^{(j)}_{k_1k_2}-\tilde{t}^{(j)}_{k_1k_2} |^2=\mathcal{O}(1/n_{\min})$ uniformly holds.

\end{lemma}
\begin{proof}
For $j \in \mathcal{H}_0,$ $t_{k_1k_2}^{(j)}-\tilde{t}_{k_1k_2}^{(j)}=\biggl( \frac{1}{n_{k_1}} \sum\limits_{i=1}^{n_{k_1}} f^{(j)}(X_{k_1,i}, \theta_{k_1}^*)-\frac{1}{n_{k_2}} \sum\limits_{i=1}^{n_{k_2}} f^{(j)}(X_{k_2,i}, \theta_{k_2}^*)\biggl)\cdot$
\begin{align*}
    &\biggl( 
\dfrac{1}{\sqrt{\frac{1/\gamma}{n_{k_1}}(\widehat{\sigma}^{(j)}_{k_1})^2 + \frac{1/\gamma}{n_{k_2}}(\widehat{\sigma}^{(j)}_{k_2})^2}}- \dfrac{1}{\sqrt{\frac{1/\gamma}{n_{k_1}}({\sigma}^{(j)}_{k_1})^2 + \frac{1/\gamma}{n_{k_2}}({\sigma}^{(j)}_{k_2})^2}}\biggl)\\
&+\dfrac{R_{k_1}^{(j)}-R_{k_2}^{(j)}}{\sqrt{\frac{1/\gamma}{n_{k_1}}({\sigma}^{(j)}_{k_1})^2 + \frac{1/\gamma}{n_{k_2}}({\sigma}^{(j)}_{k_2})^2}}.
\end{align*}
$\text{ Since }\biggl( \frac{1}{n_{k_1}} \sum\limits_{i=1}^{n_{k_1}} f^{(j)}(X_{k_1,i}, \theta_{k_1}^*)-\frac{1}{n_{k_2}} \sum\limits_{i=1}^{n_{k_2}} f^{(j)}(X_{k_2,i}, \theta_{k_2}^*)\biggl)=\mathcal{O}(n_{k_1}^{-1/2}+n_{k_2}^{-1/2}),$ and

 $\biggl( 
\dfrac{1}{\sqrt{\frac{1}{n_{k_1}}(\widehat{\sigma}^{(j)}_{k_1})^2 + \frac{1}{n_{k_2}}(\widehat{\sigma}^{(j)}_{k_2})^2}}- \dfrac{1}{\sqrt{\frac{1}{n_{k_1}}({\sigma}^{(j)}_{k_1})^2 + \frac{1}{n_{k_2}}({\sigma}^{(j)}_{k_2})^2}}\biggl)\preceq\sqrt{\dfrac{n_{k_1}n_{k_2}}{n_{k_1}+n_{k_2}}}(|({\sigma}^{(j)}_{k_1})^2-(\widehat{\sigma}^{(j)}_{k_1})^2|+|({\sigma}^{(j)}_{k_2})^2-(\widehat{\sigma}^{(j)}_{k_2})^2|)$, also noting that $\sum_j |({\sigma}^{(j)}_{k_1})^2-(\widehat{\sigma}^{(j)}_{k_1})^2|^{v}=||({\sigma}_{k_1})^2-(\widehat{\sigma}_{k_1})^2||_2^{v/2}\preceq \mathcal{O}(n^{-v/4})$, together with cauchy-schwarz inequality, the proof is done.
\end{proof}

\begin{lemma}\label{lem:R_k_fixed_p}
    For each individual $k$, $\mathbb{E}\|R_k\|_2^{v/2}=\mathcal{O}( n_k^{-v/2})$.
\end{lemma}
\begin{proof}
    Using the conclusions in Lemma \ref{lemma: variance_diverge_p}, we have $\mathbb{E}\|P_k\|_2^{v}=\mathcal{O}( n_{k}^{-v/2})$ and $\mathbb{E}\|Q_k\|_2^{v}=\mathcal{O}( n_{k}^{-v/2})$. Thus $\mathbb{E}\|(P_k+Q_k)(\widehat{\theta}_k-\theta^*_k)\|_2^{v/2}\leq\sqrt{\mathbb{E}\|(P_k+Q_k)\|_2^{v}\mathbb{E}\|\widehat{\theta}_k-\theta^*_k\|_2^{v}}$
    
    $\leq\sqrt{2^{v}\cdot\left(\mathbb{E}\|P_k\|_2^{v}+\mathbb{E}\|Q_k\|_2^{v}\right)\mathbb{E}\|\widehat{\theta}_k-\theta^*_k\|_2^{v}}=\mathcal{O}( n_{k}^{-v/2}).$
    \end{proof}

\begin{lemma}\label{lemma: A_k_1,k_2}
 Denote \( r^{(j)}_{k_1} = \sum\limits_{k_2 \ne k_1} r_{k_1,k_2}^{(j)}\), then
    $$\dfrac{1}{\sqrt{2K-2}}\left(\sum\limits_{k=1}^{K}\dfrac{r_k}{n_{k}^2}\left(\sum\limits_{i=1}^{n_{k}}f_{k_1}^{(j)} (X_{k,i},\theta_{k}^{*})\right)^{2}-(K-1)\right)\xrightarrow[]{}\mathcal{N}(0,1). $$
    
Moreover, the result remains valid when replacing \( r_k \) with its estimate \( \hat{r}^{(j)}_k = \sum\limits_{k_2 \ne k_1} \hat{r}^{(j)}_{k_1,k_2}  \).

\end{lemma}

\begin{proof}
$\EE\left(\sum\limits_{i=1}^{n_{k_1}}f_{k_1}^{(j)} (X_{k_1,i},\theta_{k_1}^{*})\right)^{2}=n_{k_1}(\sigma_{k_1}^{(j)})^2,$ thus we have $$\EE\sum\limits_{k=1}^{K}\dfrac{r_k}{n_{k_1}^2}\left(\sum\limits_{i=1}^{n_{k_1}}f_{k_1}^{(j)} (X_{k_1,i},\theta_{k_1}^{*})\right)^{2}=\sum\limits_{k=1}^{K}n_{k_1}^{-1}(\sigma_{k}^{j})^2{r_k}$$
$$=\sum\limits_{k_1=1}^{K}\left(1-\dfrac{\prod\limits_{k\ne k_1}n_k^{-1}({\sigma}_{k}^{(j)})^{2} }{\sum\limits_{1 \leq l \leq K}\prod\limits_{k\ne l}n_k^{-1}({\sigma}_{k}^{(j)})^{2}}\right) =K-1.$$

Similarly, we have $\EE\sum\limits_{k=1}^{K}\dfrac{{r}^{(j)}_k}{n}\left(\sum\limits_{i=1}^{n_{k_1}}f_{k_1}^{(j)} (X_{k_1,i},\theta_{k_1}^{*})\right)^{2}=\sum\limits_{k=1}^{K}{{r}^{(j)}_k}(\sigma_{k}^{i})^2$
    $$\EE\left(\sum\limits_{k=1}^{K}\dfrac{{r}^{(j)}_k}{n}\left(\sum\limits_{i=1}^{n_k}f_{k}^{(j)} (X_{k_1,i},\theta_{k_1}^{*})\right)^{2}\right)^2=\EE\sum\limits_{k=1}^{K}\dfrac{r^{(j)}_k}{n^2}\left(\left(\sum\limits_{i=1}^{n_k}f_{k}^{(j)} (X_{k_1,i},\theta_{k_1}^{*})\right)^{2}\right)^2$$
$$=\sum\limits_{1 \leq l \leq K}\frac{\prod\limits_{k \neq l} \left( {\sigma}_i^{k} \right)^2(K-1) }{\sum\limits_{1 \leq l \leq K} \prod\limits_{k \neq l} \left( {\sigma}_i^{k} \right)^2}=(K-1)$$

Each term's variance $\sim 2(r^{(j)}_{k_1})^2n_{k_1}^{-2}(\sigma_{k_1}^{i})^4\sim 2\left(1-\dfrac{\prod\limits_{k\ne k_1}n_k^{-1}({\sigma}_{k}^{(j)})^{2} }{\sum\limits_{1 \leq l \leq K}\prod\limits_{k\ne l}n_k^{-1}({\sigma}_{k}^{(j)})^{2}}\right)^2\overset{\triangle}{=}2(1-d_k)^2$ where $\sum_k d_k=1$. For $0<\delta<0.5,$ sum of term's $(2+2\delta)$-th moment $\sim$ $K^{2+2\delta}$. Similarly, by  Assumption \ref{assumption:Smoothness for Hessian}, each term's $(2+2\delta)$-th moment $\sim 2(1-d_k)^{2+2\delta}$ summing to $\mathcal{O}(K)=o(K^{2+2\delta}),$ by Lyapunov's theorem for CLT we get proved.
Thus it goes to standard normal. When we replace $r_k$ with $\widehat{r}_k$, noting we've proved in Lemma \ref{lemma:RK-hatRK_AK1K2} that each term has a difference at a rate $O\left(\frac{1}{\sqrt{n_{k}}}\right)$ while there're K terms, and the total difference after replacing $r_k$ with $\widehat{r}_k$ is $\mathcal{O}(\sqrt{1/(2K-2)}\cdot K\frac{1}{\sqrt{n_{\min}}})$ and is negligable when $K=o(n_{\text{min}}).$
\end{proof}
\begin{lemma}\label{lemma:rk1k2-hat{rk1k2}}
  Under Assumptions \ref{assumption:identifiability_compactness}–\ref{assumption: Local Strong Convexity}, $\sum_j\mathbb{E}|r_{k_1,k_2}^{(j)}-\widehat{r}_{k_1,k_2}^{(j)}|=\mathcal{O}(\dfrac{n_{k_1}\sqrt{n_{k_1}}+n_{k_2}\sqrt{n_{k_2}}}{N})$
\end{lemma}
\begin{proof}
    Since $r^{(j)}_{k_1,k_2}-\widehat{r}_{k_1,k_2}^{(j)}=\dfrac{n_{k_1} (\sigma_{k_1}^{(j)})^2 n_{k_2} (\sigma_{k_2}^{(j)})^2}{\sum\limits_{l=1}^{K}n_l (\sigma_l^{(j)})^2}-\dfrac{n_{k_1} (\widehat{\sigma}_{k_1}^{(j)})^2 n_{k_2} (\widehat{\sigma}_{k_2}^{(j)})^2}{\sum\limits_{l=1}^{K}n_l (\widehat{\sigma}_l^{(j)})^2}$
    \begin{align*}
r_{k_1,k_2}^{(j)}-\widehat{r}_{k_1,k_2}^{(j)}&=\frac{n_{k_1} n_{k_2}(\sigma_{k_1}^{(j)})^2 (\sigma_{k_2}^{(j)})^2\sum\limits_{l=1}^{K}n_l (\widehat{\sigma}_l^{(j)})^2-n_{k_1} n_{k_2}(\widehat{\sigma}_{k_1}^{(j)})^2 (\widehat{\sigma}_{k_2}^{(j)})^2\sum\limits_{l=1}^{K}n_l (\sigma_l^{(j)})^2}{(\sum\limits_{l=1}^{K}n_l (\sigma_l^{(j)})^2)(\sum\limits_{l=1}^{K}n_l (\widehat{\sigma}_l^{(j)})^2)}\\
&=\frac{n_{k_1} n_{k_2}(\sigma_{k_1}^{(j)})^2 (\sigma_{k_2}^{(j)})^2\left(\sum\limits_{l=1}^{K}n_l (\widehat{\sigma}_l^{(j)})^2-\sum\limits_{l=1}^{K}n_l (\sigma_l^{(j)})^2\right)}{(\sum\limits_{l=1}^{K}n_l (\sigma_l^{(j)})^2)(\sum\limits_{l=1}^{K}n_l (\widehat{\sigma}_l^{(j)})^2)}\\
&\quad+\frac{n_{k_1} n_{k_2}\sum\limits_{l=1}^{K}n_l (\sigma_l^{(j)})^2\left((\sigma_{k_1}^{(j)})^2 (\sigma_{k_2}^{(j)})^2-(\widehat{\sigma}_{k_1}^{(j)})^2 (\widehat{\sigma}_{k_2}^{(j)})^2\right)}{(\sum\limits_{l=1}^{K}n_l (\sigma_l^{(j)})^2)(\sum\limits_{l=1}^{K}n_l (\widehat{\sigma}_l^{(j)})^2)}\\
\vert r_{k_1,k_2}^{(j)}-\widehat{r}_{k_1,k_2}^{(j)}\vert&\leqslant\frac{2n_{k_1} n_{k_2}(\sigma^{+})^5\left(\sum\limits_{l = 1}^{K}n_l\vert\widehat{\sigma}_l^{(j)}-\sigma_l^{(j)}\vert+\sum\limits_{l = 1}^{K}n_l\left(\vert\sigma_{k_1}^{(j)}-\widehat{\sigma}_{k_1}^{(j)}\vert+\vert\sigma_{k_2}^{(j)}-\widehat{\sigma}_{k_2}^{(j)}\vert\right)\right)}{(\sum\limits_{l = 1}^{K}n_l(\sigma^{-})^2)^2}
\end{align*}
Thus we have $\sum_j\mathbb{E}\vert r_{k_1,k_2}^{(j)}-\widehat{r}_{k_1,k_2}^{(j)}\vert =$ $$ O\left( \frac{2n_{k_1} n_{k_2}(\sigma^{+})^5\left(\sum\limits_{l = 1}^{K}\sqrt{n_l}+\sum\limits_{l = 1}^{K}n_l\left(n_{k_1}^{-1/2}+n_{k_2}^{-1/2}\right)\right)}{(\sum\limits_{l = 1}^{K}n_l(\sigma^{-})^2)^2} \right)=\mathcal{O}((\dfrac{n_{k_1}\sqrt{ n_{k_1}}+n_{k_2}\sqrt{ n_{k_2}}}{N})).$$
\end{proof}
\begin{lemma}\label{lemma:rk-hat{rk}}
    For each $1\leq k\leq K$, $\mathbb{E}\sum_j|r_{k}^{(j)}-\widehat{r}_{k}^{(j)}|^v=\mathcal{O}(n_k^{v/2})$ holds uniformly.
\end{lemma}
\begin{proof}
    \begin{align*}
|r^{(j)}_{k}-\widehat{r}^{(j)}_{k}|&=\biggl | \frac{n_{k}(\sigma_{k}^{(j)})^{-2}\cdot\left(\sum_{l\neq k}n_{l}\cdot(\sigma_{l}^{(j)})^{-2}\right)}{\sum\limits_{l = 1}^{K}n_{l}\cdot(\sigma_{l}^{(j)})^{-2}}-\frac{n_{k}(\widehat{\sigma}_{k}^{(j)})^{-2}\left(\sum\limits_{l\neq k}n_{l}\cdot(\widehat{\sigma}_{l}^{(j)})^{-2}\right)}{\sum_{l = 1}^{K}n_{l}(\widehat{\sigma}_{l}^{(j)})^{-2}}\biggl | 
 \\
&\leq\frac{[n_{k}(\sigma^{-})^{-2}\cdot\sum\limits_{l\neq k}n_{l}\cdot ((\sigma_{l}^{(j)})^{-2}-(\widehat\sigma_{l}^{(j)})^{-2})+N(\sigma^{-})^{-4}\cdot n_{k}\cdot(\sigma^{+})\cdot(\sigma_{k}^{(j)}-\widehat\sigma_{k}^{(j)})]}{N\cdot(\sigma^{+})^{-2}}\\
&\preceq  n_k\cdot |\sigma_{k}^{(j)}-\widehat\sigma_{k}^{(j)}|
\end{align*}
Due to $\mathbb{E}\sum_{j}|\sigma_{k}^{(j)}-\widehat\sigma_{k}^{(j)}|^{v}=\mathcal{O}(n_k^{-v/2})$ uniformly holds then the proof is done.

\end{proof}
\begin{lemma}\label{lemma:RK-hatRK_AK1K2}
   For each $1\leq k\leq K$, $\mathbb{E}\sum\limits_{1\leq j\leq p}\biggl|\dfrac{r_k^{(j)}-\widehat{r}^{(j)}_k}{n_{k}^2}\left(\sum\limits_{i=1}^{n_{k}}f_{k_1}^{(j)} (X_{k,i},\theta_{k}^{*})\right)^{2}\biggl|=\mathcal{O}(n_k^{-1/2})$.
\end{lemma}
\begin{proof} By Cauchy-Schwarz inequality,
    $$\mathbb{E}\sum\limits_{1\leq j\leq p}\biggl|\dfrac{r^{(j)}_k-\widehat{r}^{(j)}_k}{n_{k}^2}\left(\sum\limits_{i=1}^{n_{k}}f_{k_1}^{(j)} (X_{k,i},\theta_{k}^{*})\right)^{2}\biggl|\leq \sqrt{\sum\limits_{1\leq j\leq p}\mathbb{E}\biggl|\dfrac{r^{(j)}_k-\widehat{r}^{(j)}_k}{n_{k}^2}\biggl|^2\mathbb{E}\biggl|\left(\sum\limits_{i=1}^{n_{k}}f_{k_1}^{(j)} (X_{k,i},\theta_{k}^{*})\right)^{2}\biggl|^2}$$
Which is $\mathcal{O}(n_k^{-1/2})$  due to Lemma \ref{lemma:rk-hat{rk}}.\end{proof}
\begin{lemma}\label{var_of_B_{k1k2}}
For the term $B_{k_1 k_2}^{(j)}$,  we have
    $$Var\left(\sum\limits_{1 \leq k_1 < k_2 \leq K}\dfrac{r_{k_1 k_2}^{(j)}}{{n_{k_1}n_{k_2}}}\left(\sum\limits_{i=1}^{n_{k_1}}f_{k_1}^{(j)} (X_{k_1,i},\theta_{k_1}^{*})\cdot\sum\limits_{i=1}^{n_{k_2}}f_{k_2}^{(j)} (X_{k_2,i},\theta_{k_2}^{*})\right)\right)<1$$
\begin{proof}
\text{We begin with some simple probability calculation.}
$$
\begin{aligned}
    &Var\left(\sum\limits_{1 \leq k_1 < k_2 \leq K}\dfrac{r_{k_1 k_2}^{(j)}}{{n_{k_1}n_{k_2}}}\left(\sum\limits_{i=1}^{n_{k_1}}f_{k_1}^{(j)} (X_{k_1,i},\theta_{k_1}^{*})\cdot\sum\limits_{i=1}^{n_{k_2}}f_{k_2}^{(j)} (X_{k_2,i},\theta_{k_2}^{*})\right)\right)\\
    =&\mathbf{E}\left(\sum\limits_{1 \leq k_1 < k_2 \leq K}\dfrac{r_{k_1 k_2}^{(j)}}{{n_{k_1}n_{k_2}}}\left(\sum\limits_{i=1}^{n_{k_1}}f_{k_1}^{(j)} (X_{k_1,i},\theta_{k_1}^{*})\cdot\sum\limits_{i=1}^{n_{k_2}}f_{k_2}^{(j)} (X_{k_2,i},\theta_{k_2}^{*})\right)\right)^2
\end{aligned}
$$
 When $k_1\neq k_3$, due to independence,  we have \[\mathbf{E}\left(\sum\limits_{i = 1}^{n_{k_1}}f_{k_1}^{(j)} (X_{k_1,i},\theta_{k_1}^{*})\cdot\sum\limits_{i = 1}^{n_{k_2}}f_{k_2}^{(j)} (X_{k_2,i},\theta_{k_2}^{*})\right)\left(\sum\limits_{i = 1}^{n_{k_2}}f_{k_2}^{(j)} (X_{k_2,i},\theta_{k_2}^{*})\cdot\sum\limits_{i = 1}^{n_{k_3}}f_{k_3}^{(j)} (X_{k_3,i},\theta_{k_3}^{*})\right)=0.\] 
Thus we have
$$
\begin{aligned}
    &\mathbf{E}\left(\sum\limits_{1 \leq k_1 < k_2 \leq K}\dfrac{r_{k_1 k_2}^{(j)}}{{n_{k_1}n_{k_2}}}\left(\sum\limits_{i=1}^{n_{k_1}}f_{k_1}^{(j)} (X_{k_1,i},\theta_{k_1}^{*})\cdot\sum\limits_{i=1}^{n_{k_2}}f_{k_2}^{(j)} (X_{k_2,i},\theta_{k_2}^{*})\right)\right)^2\\
    =&\sum\limits_{1 \leq k_1 < k_2 \leq K}\mathbf{E}\left(\dfrac{r_{k_1 k_2}^{(j)}}{{n_{k_1}n_{k_2}}}\sum\limits_{i=1}^{n_{k_1}}f_{k_1}^{(j)} (X_{k_1,i},\theta_{k_1}^{*})\cdot\sum\limits_{i=1}^{n_{k_2}}f_{k_2}^{(j)} (X_{k_2,i},\theta_{k_2}^{*})\right)^2\\
    =&\sum\limits_{1 \leq k_1 < k_2 \leq K}{(r_{k_1 k_2}^{(j)})^2}\mathbf{E}\left(\dfrac{\sum\limits_{i=1}^{n_{k_1}}f_{k_1}^{(j)} (X_{k_1,i},\theta_{k_1}^{*})}{n_{k_1}}\right)^2\cdot\mathbf{E}\left(\dfrac{\sum\limits_{i=1}^{n_{k_2}}f_{k_2}^{(j)} (X_{k_2,i},\theta_{k_2}^{*})}{n_{k_2}}\right)^2\\
    =&\sum\limits_{1 \leq k_1 < k_2 \leq K}{(r_{k_1 k_2}^{(j)})^2}n_{k_1}^{-1}(\sigma_{k_1}^{(j)})^2 n_{k_2}^{-1}(\sigma_{k_2}^{(j)})^2\\
    =&\dfrac{\sum\limits_{1 \leq k_1 < k_2 \leq K}\left(\prod\limits_{k\ne k_1,k_2}n_k^{-1}({\sigma}_{k}^{(j)})^{2}\right)^2 n_{k_1}^{-1}(\sigma_{k_1}^{(j)})^2 n_{k_2}^{-1}(\sigma_{k_2}^{(j)})^2}{\left(\sum\limits_{1 \leq l \leq K}\prod\limits_{k\ne l}n_k^{-1}({\sigma}_{k}^{(j)})^{2}\right)^2}\\
    =&1-\dfrac{\sum\limits_{1 \leq l \leq K}\left(\prod\limits_{k\ne l}n_k^{-1}({\sigma}_{k}^{(j)})^{2}\right)^2}{\left(\sum\limits_{1 \leq l \leq K}\prod\limits_{k\ne l}n_k^{-1}({\sigma}_{k}^{(j)})^{2}\right)^2}<1
\end{aligned}
$$
    \end{proof}
    
\end{lemma}

\begin{lemma}\label{lemma: BCD_k1k2}
    For the remainder terms, under Assumptions \ref{assumption:identifiability_compactness}–\ref{assumption: Local Strong Convexity} we have
    \[
\biggl|\sum\limits_{1 \leq k_1 < k_2 \leq K} \dfrac{1}{\sqrt{2K-2}} X^{(j)}_{k_1 k_2} {r}^{(j)}_{k_1,k_2}\biggl| \to 0,\biggl|\sum\limits_{1 \leq k_1 < k_2 \leq K} \dfrac{1}{\sqrt{2K-2}} X^{(j)}_{k_1 k_2} \widehat{r}^{(j)}_{k_1,k_2}\biggl| \to 0, \quad \text{for } X \in \{B, C, D\}.
\]

\end{lemma}

\begin{proof}
By Lemma \ref{var_of_B_{k1k2}}, we have \( \text{Var}\left(\sum\limits_{1 \leq k_1 < k_2 \leq K} r_{k_1,k_2}^{(j)}B_{k_1,k_2}^{(j)}\right) < 1\), which implies:

\[
\dfrac{1}{\sqrt{2K-2}} \sum\limits_{1 \leq k_1 < k_2 \leq K} r_{k_1,k_2}^{(j)}B_{k_1,k_2}^{(j)} =O_p(\frac{1}{\sqrt{K}})\xrightarrow[]{} 0.
\]
Since Lemma \ref{lemma:rk1k2-hat{rk1k2}} holds uniformly for all $k_1 \neq k_2$ up to some constants, we get:

We start with the expression \(\frac{1}{\sqrt{2K-2}}\sum_{1\leq k_1<k_2\leq K}(r_{k_1,k_2}^{(j)}-\widehat{r}_{k_1,k_2}^{(j)})B_{k_1,k_2}^{(j)}\). By the lemma, it equals \(\frac{1}{\sqrt{2K-2}}O\left(\sum_{1\leq k_1<k_2\leq K}\frac{n_{k_2}\sqrt{n_{k_1}}+n_{k_1}\sqrt{n_{k_2}}}{N\sqrt{n_{k_1}n_{k_2}}}\right)\).

Simplifying the fraction inside the sum gives \(\frac{1}{\sqrt{2K-2}}O\left(\sum_{1\leq k_1<k_2\leq K}\left(\frac{\sqrt{n_{k_1}}}{N}+\frac{\sqrt{n_{k_2}}}{N}\right)\right)=O\left( K^{-\frac{1}{2}}\sum_{1\leq k_1<k_2\leq K}\frac{\sqrt{n_{k_1}}+\sqrt{n_{k_2}}}{N}\right)\).

Bounding this sum leads to \(O\left(\frac{\sqrt{K}}{\sqrt{n_{\min}}}\right)\), where \(n_{\min}=\min\{n_1,n_2,\cdots,n_K\}\).

Thus the difference is negligible when \(K = \mathcal{O}(n_{\min})\) and we have \[
\dfrac{1}{\sqrt{2K-2}} \sum\limits_{1 \leq k_1 < k_2 \leq K} \widehat{r}_{k_1,k_2}^{(j)}B_{k_1,k_2}^{(j)} \xrightarrow[]{} 0.
\]
Additionally, since \( \mathbb{E}\left((R_{k_1}-R_{k_2})^2\right) = O\left(\dfrac{1}{n_{k_1}^2}+\dfrac{1}{n_{k_2}^2}\right) \) uniformly in $k_1,k_2$, it follows that:

\[
\sum_{j}\biggl|\sum\limits_{1 \leq k_1 < k_2 \leq K} {r}^{(j)}_{k_1 k_2}\mathbb{E}    \left((R_{k_1}-R_{k_2})^2\right)\biggl| =  \sum\limits_{1 \leq k_1 < k_2 \leq K} O\left((\dfrac{1}{n_{k_1}^2}+\dfrac{1}{n_{k_2}^2})\cdot{r}^{(j)}_{k_1 k_2}\right).
\]
Now, we evaluate the sum, since the variances are bounded and bounded away from zero, recall that ${r}^{(j)}_{k_1,k_2} = \dfrac{\prod\limits_{k\ne k_1,k_2}n_{k}^{-1}(\sigma_{k}^{(j)})^{2} }{\sum\limits_{1 \leq l \leq K}\prod\limits_{k\ne l}n_{k}^{-1}(\sigma_{k}^{(j)})^{2}}$, thus $\sum_{k\ne k_1}\dfrac{{r}^{(j)}_{k_1,k}}{n_{k_1}} = \sum_{k\ne k_1}\dfrac{n_{k_1}^{-1}\prod\limits_{l\ne k_1,k}n_{l}^{-1}(\sigma_{j}^{l})^{2} }{\sum\limits_{1 \leq l \leq K}\prod\limits_{k\ne l}n_{k}^{-1}(\sigma_{k}^{(j)})^{2}}\leq \dfrac{\sum_{k\ne k_1}\prod\limits_{l\ne k}n_{l}^{-1}(\sigma_{j}^{l})^{2} }{\sum\limits_{1 \leq l \leq K}\prod\limits_{k\ne l}n_{k}^{-1}(\sigma_{k}^{(j)})^{2}}<1$.

Thus,  we obtain:
\[\sum\limits_{1 \leq k_1 < k_2 \leq K} (\dfrac{1}{n_{k_1}^2}+\dfrac{1}{n_{k_2}^2})\cdot{r}^{(j)}_{k_1 k_2}\leq \dfrac{1}{n_{\min}}\sum\limits_{1 \leq k_1 < k_2 \leq K} (\dfrac{1}{n_{k_1}}+\dfrac{1}{n_{k_2}})\cdot{r}^{(j)}_{k_1 k_2}\leq \dfrac{2K}{n_{\min}  (\sigma^{-})^2}.\]
\[
\dfrac{1}{\sqrt{2K-2}} \sum\limits_{1 \leq k_1 < k_2 \leq K} \mathbb{E} \left[ {r}^{(j)}_{k_1 k_2} \cdot (R_{k_1}-R_{k_2})^2  \right] = O\left(\dfrac{\sqrt{K}}{n_{\text{min}}}\right) \xrightarrow[]{} 0,
\]

which implies:

\[
\sum\limits_{1 \leq k_1 < k_2 \leq K} \dfrac{1}{\sqrt{2K-2}} D_{k_1,k_2}^{(j)} {r}^{(j)}_{k_1 k_2} \to 0.
\]
Also we have\[
\sum\limits_{1 \leq k_1 < k_2 \leq K} (\dfrac{1}{n_{k_1}^2}+\dfrac{1}{n_{k_2}^2})\cdot({r}^{(j)}_{k_1 k_2}-\hat{r}^{(j)}_{k_1 k_2})=\mathcal{O}(\dfrac{K}{n_{\min}\sqrt{{n_{\min}}}})
\]
Which is negligable when $K=o(n_{\min}).$

Finally, let us consider the terms related to \( \{C_{k_1 k_2}^{(j)}\} \). Since

\[
\mathbb{E} \left| \sum\limits_{i=1}^{n_k} f_{k}^{(j)} (X_{k,i}, \theta_{k}^{*}) \right|^2 = n_k(\sigma_{k}^{(j)})^2 \leq n_k \sigma_{\max}^{2},
\]

we have \( \mathbb{E} \left| \sum\limits_{i=1}^{n_k} f_{k}^{(j)} (X_{k,i}, \theta_{k}^{*}) \right| = \mathcal{O}(\sqrt{n_k}) \). Therefore, when considering the terms related to \( \{D_{k_1,k_2}^{(j)}\} \) in the test statistic and given that \( \mathbb{E} \sum_j|R_k^{(j)}| = O\left(\dfrac{1}{n_k}\right),\mathbb{E} \sum_j \sum\limits_{i=1}^{n_{k_1}} f_{k_1}^{(j)} (X_{k_1,i}, \theta_{k_1}^{*})=\mathcal{O}(\sqrt{n_k}) \), we obtain the expression 

\[
\mathbb{E} \left[ \dfrac{1}{\sqrt{2K-2}} \sum\limits_{1 \leq k_1 < k_2 \leq K} r^{(j)}_{k_1 k_2} \left| \left( \dfrac{1}{{n_{k_1}}}\sum\limits_{i=1}^{n_{k_1}} f_{k_1}^{(j)} (X_{k_1,i}, \theta_{k_1}^{*})-\dfrac{1}{{n_{k_2}}}\sum\limits_{i=1}^{n} f_{k_2}^{(j)} (X_{k_2,i}, \theta_{k_2}^{*}) \right)(R_{k_1}^{(j)}-R_{k_2}^{(j)}) \right| \right],
\]

which can be similarly bounded by :

\[
= O\left(\dfrac{1}{\sqrt{K}}\right) \sum\limits_{1 \leq k_1 < k_2 \leq K} (\dfrac{1}{n_{k_1}}+\dfrac{1}{n_{k_2}})(\dfrac{1}{\sqrt{n_{k_1}}}+\dfrac{1}{\sqrt{n_{k_2}}})r^{(j)}_{k_1 k_2}.
\]
From the proof for $D_{k_1,k_2}^{(j)}$, we know that $ \sum\limits_{1 \leq k_1 < k_2 \leq K} (\dfrac{1}{n_{k_1}}+\dfrac{1}{n_{k_2}})r^{(j)}_{k_1 k_2} =\mathcal{O}(K)$, and $(\dfrac{1}{\sqrt{n_{k_1}}}+\dfrac{1}{\sqrt{n_{k_2}}})=\mathcal{O}(\dfrac{1}{\sqrt{{n_{\min}}}})$, thus the whole expression can be bound by $O\left(\dfrac{K}{\sqrt{Kn_{\text{min}}}}\right)=\mathcal{O}(\dfrac{\sqrt{K}}{\sqrt{n_{\text{min}}}})$.
Thus, when \( K = o(n_{\text{min}}) \), this implies that:  $\sum\limits_{1 \leq k_1 < k_2 \leq K} \dfrac{1}{\sqrt{2K-2}} C_{k_1 k_2}^{(j)} r_{k_1 k_2}^{(j)} \to 0. $ We can similarly conclude that $\sum\limits_{1 \leq k_1 < k_2 \leq K} \dfrac{1}{\sqrt{2K-2}} C_{k_1 k_2}^{(j)} \hat{r}^{(j)}_{k_1 k_2} \to 0$ via Lemma \ref{lemma:rk1k2-hat{rk1k2}}.\end{proof}
\begin{lemma}\label{lem:G_H_k1k2_local_noncentral}
    Considering local alternative  \eqref{eq:gaussian_hypothesis} where  $n_1=n_2=\cdots=n_K$, for  the remaining terms related to $G_{k_1,k_2}$ and non-central component related to $H_{k_1,k_2},$ under $K=o(\sqrt{n})$ we have: 
      \[
\sum\limits_{1\leq k_1<k_2\leq K} \dfrac{1}{\sqrt{2K-2}} G_{k_1,k_2} \widehat{r}^{(j)}_{k_1,k_2}=o(\sqrt{K}) \text{, and }\]
\[{\sum\limits_{1\leq k_1<k_2\leq K} \dfrac{1}{\sqrt{2K-2}} H_{k_1,k_2} \widehat{r}_{k_1,k_2}},\text{ and }\nu_{N,K}^{(j)}\] \[\text{ are asymptotically equivalent to } \sqrt{2}c_{\infty}\sigma_j^{-2}K^{0.5-\beta}\log K,\] thus $\text{SNR}_{\text{Wald}}/\sqrt{2}c_{\infty}\sigma_j^{-2}K^{0.5-\beta}\log K \to 1.$ 

\end{lemma}
\begin{proof}
The proof is the same for each dimension so we omit the superscript $(j)$ here. Since $ K=o(\sqrt{n}),$ due to proof for  Lemma \ref{theorem: variance_estimates_thereom1}, the variance estimates are consistent, so we only have to consider the case where $\widehat{r}_{k_1,k_2}$ are replaced by  ${r}_{k_1,k_2}.$ By the proof for Theorem \ref{Wald consistency}, we know that the remainder terms in $G_{k_1,k_2}$ are negligible, thus we only have to consider $G_{k_1,k_2}$ without $R_{k_1},R_{k_2}.$

For \(G_{k_1,k_2}\), recall that $r_{k_1,k_2}=\dfrac{n(\sigma_{k_1})^{-2}(\sigma_{k_2})^{-2}}{\sum\limits_{l=1}^K(\sigma_{l})^{-2}}$, denote $S=\sum\limits_{l=1}^K(\sigma_{l})^{-2}$, and consider 
$\sum_{1\leq k_1<k_2\leq K}r_{k_1,k_2}\left(\dfrac{1}{n}\sum\limits_{i=1}^{n}f_{k_1} (X_{k_1,i},\theta_{k_1}^{*})-\dfrac{1}{n}\sum\limits_{i=1}^{n}f_{k_2} (X_{k_2,i},\theta_{k_2}^{*})\right)(\theta_{k_1}^*-\theta_{k_2}^*)$

$=\sum\limits_{k=1}^{K}n\sigma_k^{-2}S^{-1}\sum\limits_{h\ne k}\sigma_h^{-2}(\theta_{k}^*-\theta_h^*)\dfrac{\sum\limits_{i=1}^{n}f_{k}}{n};$

Since $E(\theta_k^*-\theta_h^*)^2=K^{-\beta}\mu^2$ and the randomness of the true parameters' difference is set to independent to the random variables of M-functions, recall the definition of $S$ that $S/\sum\limits_{h\ne k}\sigma_h^{-2}\to 1$ as $K\to\infty,$ thus, $[S^{-1}\sum\limits_{h\ne k}\sigma_h^{-2}(\theta_{k}^*-\theta_h^*)]^2 \to K^{1-\beta}\mu^2$, in conclusion we have, $\dfrac{Var(\sum\limits_{k=1}^{K}n\sigma_k^{-2}S^{-1}\sum\limits_{h\ne k}\sigma_h^{-2}(\theta_{k}^*-\theta_h^*)\dfrac{\sum\limits_{i=1}^{n}f_{k}}{n})}{\sum\limits_{k=1}^{K}n\sigma_k^{-2}K^{1-\beta}\mu^2}\to 1.$ Moreover,
$$\EE\sum_{1\leq k_1<k_2\leq K}r_{k_1,k_2}\left(\dfrac{1}{n}\sum\limits_{i=1}^{n}f_{k_1} (X_{k_1,i},\theta_{k_1}^{*})-\dfrac{1}{n}\sum\limits_{i=1}^{n}f_{k_2} (X_{k_2,i},\theta_{k_2}^{*})\right)(\theta_{k_1}^*-\theta_{k_2}^*)=0$$ due to $\EE f_k(X_{k,i},\theta_k^*)=0.$ In conclusion we have:
\[\dfrac{\dfrac{1}{\sqrt{2K-2}}\sum\limits_{1\leq k_1<k_2\leq K} \widehat r_{k_1,k_2}G_{k_1,k_2}}{2\sqrt{K}}\to 0.
\]

Which means the asymptotic variance of Wald statistic retains 1. We also have under $K=o(\sqrt{n})$ and the local alternative, $$\sqrt{2K-2}\EE\nu_{N,K}^{(j)}=\EE\sum\limits_{1\leq k_1 < k_2\leq K} \widehat{r}_{k_1k_2}^{(j)}\cdot \EE({\theta}_{k_1}^*-\theta_{k_2}^*)^2\sim \sum\limits_{1\leq k_1 < k_2\leq K} \dfrac{n(\sigma_{k_1}^{(j)})^{-2}(\sigma_{k_2}^{(j)})^{-2}}{\sum\limits_{l=1}^K(\sigma_{l}^{(j)})^{-2}}\cdot \EE({\theta}_{k_1}^*-\theta_{k_2}^*)^2$$
$$=\sum\limits_{1\leq k_1 < k_2\leq K} \dfrac{n(\sigma_{k_1}^{(j)})^{-2}(\sigma_{k_2}^{(j)})^{-2}}{\sum\limits_{l=1}^K(\sigma_{l}^{(j)})^{-2}}K^{-\beta}\mu^2\to\sum\limits_{k=1}^{K}n(\sigma_k^{(j)})^{-2}K^{-\beta}\mu^2=2c_{\infty}\sigma_j^{-2}K^{1-\beta}\log K.$$ Thus $\text{SNR}_{\text{Wald}}/\sqrt{2}c_{\infty}\sigma_j^{-2}K^{0.5-\beta}\log K \to 1.$
\end{proof}
\begin{lemma}\label{lemma: JLM_K1K2}
    For the remainder terms in modify Wald statistic for combination, we have
    \[
\sum\limits_{k\ne (1)} \dfrac{1}{\sqrt{2K-2}} X_{(1),k} \widehat{r}_{(1), k} \to 0, \quad \text{for } X \in \{J,L,M\}.
\]
\end{lemma}
\begin{proof}
 For $J_{k_1,k}$ we need to deal with 
\[
\mathbb{E} \left[ \dfrac{1}{\sqrt{2K-2}}\sum\limits_{1\leq k_1 \ne k_2 \leq K}\mathbb{I}_{k_1,k_2}^{(j)} \sum\limits_{k\ne k_1} r_{k_1,k} \left| \left( \dfrac{1}{{n_{k_1}}}\sum\limits_{i=1}^{n_{k_1}/2} f_{k_1}^{(j)} (X_{k_1,i}, \theta_{k_1}^{*}) \right)Q_{k_1}\right| \right].
\]
The difficulty here stems from two main reasons. In comparison to the previous case, \(Q_{k_1}=\mathcal{O}(n_{k_1}^{-1/2})\) instead of \(\mathcal{O}(n_{k_1}^{-1})\), which implies that \(Q_{k_1}\) has a larger order of magnitude. However, there is an advantage: in this context, there are only \(K\) terms, as opposed to \(K^2\) terms.

Second, the term \(\left(\dfrac{1}{n_{k_1}}\sum_{i = 1}^{n_{k_1}/2}f_{k_1}^{(j)}(X_{k_1,i},\theta_{k_1}^{*})\right)\) cannot be scaled as \(O_p(n_{k_1}^{-1/2})\) conditional on the event \(k_1 = \hat{k}^{(j)}_{\max}\). Nevertheless, we can bound it using the union-bound of the \(2v\)-th moment. We know that \[\mathbb{E}\left(\max_k\left|\dfrac{1}{n_{k_1}}\sum_{i = 1}^{n_{k}/2}f_{k}^{(j)}(X_{k,i},\theta_{k}^{*})\right|^{2v}\right)\leq\mathbb{E}\left(\sum_k\left|\dfrac{1}{n_{k_1}}\sum_{i = 1}^{n_{k}/2}f_{k}^{(j)}(X_{k,i},\theta_{k}^{*})\right|^{2v}\right)=\mathcal{O}\left(\sum_{k = 1}^{K}\dfrac{1}{n_k^v}\right).\]

Consequently, we have \(\mathbb{E}\left|\dfrac{1}{n_{k_1}}\sum_{i = 1}^{n_{k_1}/2}f_{k_1}^{(j)}(X_{k_1,i},\theta_{k_1}^{*})\right|=\mathcal{O}\left(\left(\sum_{k = 1}^{K}\dfrac{1}{n_k^v}\right)^{\frac{1}{2v}}\right)\). Since \(\dfrac{1}{n_{k_1}}\sum_{i = 1}^{n_{k_1}/2}f_{k_1}^{(j)}(X_{k_1,i},\theta_{k_1}^{*})\) is independent of \(Q_{k_1}=\dfrac{1}{n_{k_1}}\sum_{i = 1 + n_{k_1}/2}^{n_{k_1}}f_{k_1}^{(j)}(X_{k_1,i},\theta_{k_1}^{*})\) (they are respectively measurable with respect to non-overlapping observations \(D_{k_1}^{[1]},D_{k_1}^{[2]}\). So, \(\mathbb{E}\left|\dfrac{1}{n_{k_1}}\sum_{i = 1}^{n_{k_1}/2}f_{k_1}^{(j)}(X_{k_1,i},\theta_{k_1}^{*})\right|Q_{k_1}=\mathcal{O}\left(\dfrac{1}{\sqrt{n_{k_1}}}\left(\sum_{k = 1}^{K}\dfrac{1}{n_k^v}\right)^{\frac{1}{2v}}\right)=O\left(\dfrac{1}{n_{k_1}}\left(1+\left(\sum_{k = 1}^{K}\dfrac{n_{k_1}^v}{n_k^v}\right)^{\frac{1}{2v}}\right)\right)\).

Then, consider the following expectation:
\[
\mathbb{E}\left[\dfrac{1}{\sqrt{2K-2}}\sum\limits_{1\leq k_1 \ne k_2 \leq K}\mathbb{I}_{k_1,k_2}^{(j)}\sum_{k\neq k_1}r_{k_1,k}\biggl|\left(\dfrac{1}{n_{k_1}}\sum_{i = 1}^{n_{k_1}/2}f_{k_1}^{(j)}(X_{k_1,i},\theta_{k_1}^{*})\right)Q_{k_1}\biggl| \Biggl| k_1,k_2\right].
\]
\[
=\mathbb{E}\left[\dfrac{1}{\sqrt{2K-2}}\sum_{k\neq k_1}r_{k_1,k}\biggl|\left(\dfrac{1}{n_{k_1}}\sum_{i = 1}^{n_{k_1}/2}f_{k_1}^{(j)}(X_{k_1,i},\theta_{k_1}^{*})\right)Q_{k_1}\biggl| \Biggl| k_1,k_2\right].
\]
It can be scaled as follows:
\[
\mathcal{O}(K^{-1/2})\cdot\sum_{k\neq k_1}r_{k_1,k}O\left(\left(\sum_{k = 1}^{K}\dfrac{1}{n_k^v}\right)^{\frac{1}{2v}}\right)\cdot\dfrac{1}{\sqrt{n_{k_1}}}=\mathcal{O}(K^{-1/2})\cdot\sum_{k\neq k_1}\dfrac{r_{k_1,k}}{n_{k_1}}\left(1+\left(\sum_{k = 1}^{K}\dfrac{n_{k_1}^v}{n_k^v}\right)^{\frac{1}{2v}}\right)
\]
Since we have already proved that \(\sum_{k\neq k_1}\dfrac{r_{k_1,k}}{n_{k_1}}<1\), we only need \(\left(\sum_{k = 1}^{K}\dfrac{n_{k_1}^v}{n_k^v}\right)^{1/2v}=o(\sqrt{K})\). This condition can be satisfied when \(\max_{k_3\neq k_4}\dfrac{n_{k_3}}{n_{k_4}}=o(K^{1-1/v})\). This is because in this case, each \(\dfrac{n_{k_1}^v}{n_k^v}\) is \(o(K^{v-1})\), and when summed up, the result is \(o(K^{v})\). Taking the \(1/2v\)-th power of \(o(K^{v})\) gives \(o(\sqrt{K})\). The terms related to $k_2$ can be treated in a exactly same way. While the above analysis is uniform to all $k_1,k_2$, thus we have
\[
\mathbb{E}_{k_1,k_2}\mathbb{E}\left[\dfrac{1}{\sqrt{2K-2}}\sum_{k\neq k_1}r_{k_1,k}\biggl|\left(\dfrac{1}{n_{k_1}}\sum_{i = 1}^{n_{k_1}/2}f_{k_1}^{(j)}(X_{k_1,i},\theta_{k_1}^{*})\right)Q_{k_1}\biggl| \Biggl| k_1,k_2\right]\to 0
\]

 when \(\max_{k_3\neq k_4}\dfrac{n_{k_3}}{n_{k_4}}=o(K^{1-1/v})\) and $K\to\infty$. 

For the terms related to $L_{k_1,k},M_{k_1,k}$, we need to show
\[
\mathbb{E} \left[ \dfrac{1}{\sqrt{2K-2}}\sum\limits_{1\leq k_1 \ne k_2 \leq K}\mathbb{I}_{k_1,k_2}^{(j)} \sum\limits_{k\ne k_1} r_{k_1,k} (Q_{k_1}^2+|Q_{k_1}\dfrac{1}{n_k}\sum\limits_{i=1}^{n_k}f^{(j)}(X_{k,i},\theta_k^*)|)\right] \to 0.
\]
Since $Q_{k_1}^2=O_p(1/{n_{k_1}})$ regardless of $\mathbb{I}_{k_1,k_2}^{(j)}$ and $|Q_{k_1}\dfrac{1}{n_k}\sum\limits_{i=1}^{n_k}f^{(j)}(X_{k,i},\theta_k^*)|$ can be bounded as $ \mathcal{O}(\dfrac{1}{\sqrt{n_k n_{k_1}}})$due to independence, it can be bounded as follows:
\[
\mathcal{O}({K^{-1/2}})\cdot \sum_{k\neq k_1} \hat{r}_{k_1, k}O\left(\left(\frac{1}{\sqrt{n_{k_1}}}+\frac{1}{\sqrt{n_{k}}}\right)\cdot \frac{1}{\sqrt{n_{k_1}}}\right)=\mathcal{O}({K^{-1/2}})\cdot \sum_{k\neq k_1} \hat{r}_{k_1, k}O\left(\frac{1}{n_{k_1}}+\frac{1}{\sqrt{n_1 n_k}}\right)
\]
\[\leq \mathcal{O}({K^{-1/2}})\cdot \sum_{k\neq k_1} \hat{r}_{k_1, k}O\left(\frac{1}{n_{k_1}}\right)\cdot\left(1+\sqrt{\max_{k_3\ne k_4} \dfrac{n_{k_3}}{n_{k_4}}}\right), \] since we've proved $\sum_{k\neq k_1} \hat{r}_{k_1, k}O\left(\frac{1}{n_{k_1}}\right)=\mathcal{O}(1)$, thus we only need $\max_{k_3\ne k_4} \dfrac{n_{k_3}}{n_{k_4}}=o({K}),$ which has been assumed in Theorem \ref{theorem:tilde_W_NK_normal_independence_ECT}. 
Above all we have proven the lemma.
\end{proof}

\begin{lemma}
For the average of gradient matrix's moment bound, under Assumptions \ref{assumption:identifiability_compactness}–\ref{assumption: Local Strong Convexity},
\[\mathbb{E}\left[\left\|\dfrac{1}{n_k}\sum\limits_{i=1}^{n_k}\nabla M_k(X_{k,i},\theta_k^*)\right\|_2^{2v}\right] = O\left( n_{k}^{-v}\right),
\]
 \[
\mathbb{E}\left[\left|\left|\left|\dfrac{1}{n_k} \sum\limits_{i=1}^{n_k} \nabla_{\theta_k}^2 M_k(X_{k,i}, \theta_k^*)-\mathbb{E} \nabla_{\theta_k}^2 M_k(X_k, \theta_k^*) \right|\right|\right|^{2v}\right] = O\left(\left(\dfrac{\log p}{n_{k}}\right)^v\right).
\]

\end{lemma}
\begin{proof}See the proof for Lemma 7 in \cite{jmlr2013}.
\end{proof}

\label{app:theorem}



\end{document}